\documentclass[sigconf]{acmart}
\usepackage{etoolbox} %

\AtBeginDocument{%
  \providecommand\BibTeX{{%
    \normalfont B\kern-0.5em{\scshape i\kern-0.25em b}\kern-0.8em\TeX}}}

\copyrightyear{2022} 
\acmYear{2022} 
\setcopyright{acmlicensed} %

\acmConference[KDD '22] {Proceedings of the 28th ACM SIGKDD Conference on Knowledge Discovery and Data Mining}{August 14--18, 2022}{Washington, DC, USA.}
\acmBooktitle{Proceedings of the 28th ACM SIGKDD Conference on Knowledge Discovery and Data Mining (KDD '22), August 14--18, 2022, Washington, DC, USA}
\acmPrice{15.00}
\acmISBN{978-1-4503-9385-0/22/08}
\acmDOI{10.1145/3534678.3539418}



\usepackage{hyperref}
\usepackage{url}

\usepackage{multicol}     %

\usepackage{amsmath}
\usepackage{graphicx}
\usepackage{subcaption}
\usepackage{listings}
\usepackage{float}
\usepackage{enumitem}
\usepackage{xspace}
\usepackage{multirow}
\usepackage{array}
\usepackage{tabularx}
\usepackage{tabu}
\usepackage{bm}

\usepackage{booktabs, makecell}
\usepackage{xcolor,colortbl}
\usepackage{cleveref}
\usepackage[english]{babel}
\usepackage{pifont} 
\usepackage{soul}
\usepackage{slashbox}
\usepackage{xfrac}
\usepackage{wrapfig}
\usepackage{mdwlist}
\usepackage{arydshln}

\usepackage{adjustbox}

\usepackage{cellspace}
\usepackage{siunitx}
\usepackage{comment}

\usepackage[linesnumbered,ruled,vlined]{algorithm2e}

\newtheorem{definition}{Definition}

\newtheorem{theorem}{Theorem}

\usepackage{tikz}

\crefformat{section}{\S#2#1#3} %
\crefformat{subsection}{\S#2#1#3}
\crefformat{subsubsection}{\S#2#1#3}
\newcolumntype{L}[1]{>{\raggedright\let\newline\\\arraybackslash\hspace{0pt}}m{#1}}
\newcolumntype{C}[1]{>{\centering\let\newline\\\arraybackslash\hspace{0pt}}m{#1}}
\newcolumntype{R}[1]{>{\raggedleft\let\newline\\\arraybackslash\hspace{0pt}}m{#1}}
\newcolumntype{H}{>{\setbox0=\hbox\bgroup}c<{\egroup}@{}}

\SetCommentSty{mycommfont}

\renewcommand{\paragraph}[1]{\noindent\textbf{#1.}}

\definecolor{darkgreen}{RGB}{0,153,0}
\definecolor{dark-gray}{gray}{0.4} %

\newcommand{\githubrepo}{\href{https://github.com/GemsLab/HeteRobust}{GitHub repository\xspace}}
\newcommand{\repo}{\href{https://github.com/GemsLab/HeteRobust}{repository\xspace}}
\newcommand{\githubrepourl}{\url{https://github.com/GemsLab/HeteRobust}}

\definecolor{purple2}{RGB}{153,0,153} %
\definecolor{green2}{RGB}{0,153,0} %
\newcommand{\method}{\textsc{H$_2$GCN}\xspace}

\DeclareRobustCommand{\scaleroman}[1]{\makebox[\widthof{II}]{#1}}

\lstdefinestyle{common}{
	basicstyle = \ttfamily,
	keywordstyle=\color{blue},       %
	keywordstyle={[2]\color{cyan}}, %
	stringstyle=\color{purple2},
	commentstyle=\color{green2},
	upquote=true,                  %
	breaklines=true, frame=trBL
}

\newlist{steps}{enumerate}{1}
\setlist[steps, 1]{label = Step \arabic*:}

\newcommand{\V}[1]{\mathbf{#1}}

\newcommand{\graph}{\mathcal{G}}
\newcommand{\vertexSet}{\mathcal{V}}
\newcommand{\edgeSet}{\mathcal{E}}

\newcommand{\matA}{\mathbf{A}}

\newcommand{\matD}{\mathbf{D}}

\newcommand{\matX}{\mathbf{X}}

\newcommand{\matW}{\mathbf{W}}

\newcommand{\matI}{\mathbf{I}}

\newcommand{\vecy}{\mathbf{y}}

\newcommand{\setY}{\mathcal{Y}}
\newcommand{\setT}{\mathcal{T}_{\vertexSet}}
\newcommand{\T}{\mathsf{T}}

\newcommand{\neighNoSelfLoop}{N}
\newcommand{\attackLoss}{\mathcal{L}_{\mathrm{atk}}}
\newcommand{\matAs}{\mathbf{A}_{\mathrm{s}}}

\newcolumntype{H}{>{\setbox0=\hbox\bgroup}c<{\egroup}@{}}

\newcommand{\nettack}{\textsc{Nettack}}
\newcommand{\gambit}{gambit\xspace}
\newcommand{\columnmatrix}[4][]{\mathbf{C}^{#1}_{#2}\left(#3, #4\right)}
\newcommand{\columnvec}[4][]{\mathbf{c}^{#1}_{#2}\left(#3, #4\right)}
\newcommand{\circulantmatrix}[2]{\columnmatrix[*]{}{#1}{#2}}
\newcommand{\blockmatrix}[2]{\mathbf{B}\left(#1, #2\right)}

\makeatletter
\def\adl@drawiv#1#2#3{%
        \hskip.5\tabcolsep
        \xleaders#3{#2.5\@tempdimb #1{1}#2.5\@tempdimb}%
                #2\z@ plus1fil minus1fil\relax
        \hskip.5\tabcolsep}
\newcommand{\cdashlinelr}[1]{%
  \noalign{\vskip\aboverulesep
           \global\let\@dashdrawstore\adl@draw
           \global\let\adl@draw\adl@drawiv}
  \cdashline{#1}
  \noalign{\global\let\adl@draw\@dashdrawstore
           \vskip\belowrulesep}}
\makeatother

\ifdefmacro{\ispreprint}{%
  \makeatletter
  \newcommand\footnoteref[1]{\protected@xdef\@thefnmark{\ref{#1}}\@footnotemark}
  \makeatother
}{}

\ifdefmacro{\ispreprint}{}{%

}

\begin{document}

\title{How does Heterophily Impact the Robustness of Graph Neural Networks? Theoretical Connections and Practical Implications}

\author{Jiong Zhu}
\affiliation{%
  \institution{University of Michigan}}
\email{jiongzhu@umich.edu}

\author{Junchen Jin}
\affiliation{%
  \institution{Northwestern University}}
\email{mark.jin@u.northwestern.edu}

\author{Donald Loveland}
\affiliation{%
  \institution{University of Michigan}}
\email{dlovelan@umich.edu}

\author{Michael T. Schaub}
\affiliation{%
  \institution{RWTH Aachen University}}
\email{schaub@cs.rwth-aachen.de}

\author{Danai Koutra}
\affiliation{%
  \institution{University of Michigan}}
\email{dkoutra@umich.edu}

\renewcommand{\shortauthors}{Jiong Zhu et al.}
\renewcommand{\shorttitle}{How does Heterophily Impact the Robustness of Graph Neural Networks? \\Theoretical Connections and Practical Implications}

\begin{abstract}

We bridge two research directions on graph neural networks (GNNs), by formalizing the relation between heterophily of node labels (i.e., connected nodes tend to have dissimilar labels) and the robustness of GNNs to adversarial attacks.
Our theoretical and empirical analyses show that for homophilous graph data, impactful structural attacks always lead to reduced homophily, while for heterophilous graph data the change in the homophily level depends on the node degrees.
These insights have practical implications for defending against attacks on real-world graphs:  
we deduce that separate aggregators for ego- and neighbor-embeddings, a design principle which has been identified to significantly improve prediction for heterophilous graph data, can also offer increased robustness to GNNs. 
Our comprehensive experiments show that GNNs merely adopting this design achieve improved \emph{empirical and certifiable} robustness compared to the best-performing unvaccinated model.
Additionally, combining this design 
with explicit defense mechanisms against adversarial attacks leads to an improved robustness with up to 18.33\% performance increase under attacks compared to the best-performing vaccinated model.

\end{abstract}

\maketitle

\section{Introduction}

Graph neural networks (GNNs) aim to translate the enormous empirical success of deep learning to data defined on non-Euclidean domains, such as manifolds or graphs~\citep{bronstein2017geometric}, 
and have become important tools to solve a variety of learning problems for graph structured and geometrically embedded data.
However, recent works
show that GNNs---much like their ``standard'' deep learning counterparts---have a high sensitivity to adversarial attacks: intentionally introduced minor changes in the graph structure can lead to significant changes in performance.
This finding, first articulated by \citet{zugner2018adversarial} and \citet{dai2018adversarial}, has triggered studies that investigated different attack scenarios~\citep{xu2019topology,wu2019adversarial,li2020adversarial,ma2020towards}.

A different aspect of GNNs that has been scrutinized recently is that most GNNs do not perform well with many heterophilous datasets. %
GNNs generally perform well under homophily (or assortativity), i.e., the tendency of nodes with similar features or class labels to connect~\citep{Pei2020Geom-GCN,zhu2020beyond}. 
Such datasets are thus called \emph{homophilous} (or \emph{assortative}). 
While homophilous datasets dominate the study of networks, homophily is not a universal principle; certain networks, such as romantic relationship networks or predator-prey networks in ecology, are mostly \emph{heterophilous} (or \emph{disassortative}).
Employing a GNN which does not account for heterophily can lead to significant performance loss in heterophilous settings~\citep{MixHop,zhu2020beyond,bo2021beyond}.
Previous works have thus proposed architectures for heterophilous data.

While previous work has focused on naturally-occurring heterophily, heterophilous interactions may also be introduced as adversarial noise: 
as many GNNs exploit homophilous correlation, they can be sensitive to changes that render the data more heterophilous. A natural follow-up question is if and how this observation manifests itself in previously proposed attacking strategies on GNNs.
In this work, we thus investigate the relation between heterophily and robustness of GNNs against adversarial attacks of graph structure, focusing on semi-supervised node classification.
More specifically, our main contributions are:
\vspace{-0.3cm}
\setlist{leftmargin=*}
\begin{itemize*}
    \item \textbf{Formalization:}
        We formalize the relation between adversarial structural attacks and the change of homophily level in the underlying graphs with theoretical (\S\ref{sec:theories-connections}) and empirical (\S\ref{sec:exp-perturb-observations}) analysis.
        Specifically, we show that on homophilous graphs, effective structural attacks lead to increased heterophily, 
        while, on heterophilous graphs, they alter the homophily level contingent on node degrees. To our knowledge, this is the first formal analysis of such kind. %
    \item \textbf{Heterophily-inspired Design:}
        We show how the relation between attacks and heterophily can inspire more robust GNNs by demonstrating %
        that a key architectural feature in handling heterophily, separate aggregators 
        for ego- and neighbor-embeddings, also improves the robustness of GNNs against attacks (\S\ref{sec:robustness-design}).
    \item \textbf{Extensive Empirical Analysis:}
        We show the effectiveness of the heterophilous design in improving empirical (\S\ref{sec:exp-benchmark-study}) and certifiable (\S\ref{sec:exp-cert-robustness}) robustness of GNNs
        with extensive experiments on real-world homophilous and heterophilous datasets. 
        Specifically, we compare GNNs with this design, which we refer to as \emph{heterophily-adjusted} GNNs, to non-adjusted models, including state-of-the-art models designed with robustness in mind. 
        We find that heterophily-adjusted GNNs are up to 11.1 times more certifiably robust and have stronger performance under attacks by up to 40.00\% compared to non-adjusted, standard models.
        Moreover, this design can be combined with existing vaccination mechanisms, yielding up to 18.33\% higher accuracy under attacks than the best non-adjusted vaccinated model.
        Our code is available at \githubrepourl.
\end{itemize*}

\section{Notation and Preliminaries}
\label{sec:notation}

Let $\graph=(\vertexSet,\edgeSet,\matX)$ be a simple graph with node set $\vertexSet$, edge set $\edgeSet$ and node attributes $\matX$.
The one-hop neighborhood $N(v) = \{u: (u,v) \in \edgeSet \}$ of a node $v\in \vertexSet$ is the set of all nodes directly adjacent to $v$;
the $k$-hop neighborhood of $v\in \vertexSet$ is the set of nodes reachable by a shortest path of length $k$.
We represent the graph $\graph$ algebraically by an adjacency matrix $\matA \in \{0,1\}^{|\vertexSet|\times |\vertexSet|}$ and node feature matrix $\matX \in \mathbb{R}^{|\vertexSet| \times F}$. 
We use $\matAs = \matA + \matI$ to denote the adjacency matrix with self-loops added, and denote the corresponding row-stochastic matrices as $\bar{\matA} = \matD^{-1}\matA$ and $\bar{\matA}_{\mathrm{s}} = \matD_{\mathrm{s}}^{-1}\matAs$, respectively, where $\matD$ is a diagonal matrix with $\matD_{ii} = \sum_{j} \matA_{ij}$ ($\matD_{\mathrm{s}}$ is defined analogously).
We further assume that there exists a vector $\vecy$, which contains a unique class label $y_v$ for each node $v$.
Given a training set $\setT = \{(v_1,y_1), (v_2, y_2), ...\}$ of labeled nodes, the goal of semi-supervised node classification is to learn a mapping $\ell: \vertexSet \rightarrow \setY$ from the nodes to the set $\setY$ of class labels.

\paragraph{Graph neural networks (GNNs)} 
Most current GNNs operate according to a message passing paradigm where a representation vector $\V{r}_v$ is assigned to each node $v \in \vertexSet$ and continually updated by $K$ layers of learnable transformations.
These layers first aggregate representations over neighboring nodes $N(v)$ and then update the current representation via an encoder \texttt{ENC}. For prevailing GNN models like GCN~\citep{kipf2016semi} and GAT~\citep{velickovic2018graph}, each layer can be formalized as
    {\small $\V{r}^{(k)}_v = \texttt{ENC}\left(\texttt{AGGR}\left(\left\{\V{r}^{(k-1)}_u: u \in \neighNoSelfLoop(v) \cup \{v\}\right\}\right)\right)$},
where $\texttt{AGGR}$ is the mean function weighted by node degrees (GCN) or an attention mechanism (GAT), and $\texttt{ENC}$ is a learnable (nonlinear) mapping.

\paragraph{Adversarial attacks on graphs} 
Given a  graph $\graph=(\vertexSet,\edgeSet,\matX)$ and a GNN $f$ that processes $\graph$, an adversarial attacker tries to create a perturbed graph $\graph'=(\vertexSet,\edgeSet',\matX)$ with a modified edge-set $\edgeSet'$ such that the performance of the GNN $f$ is maximally degraded.
The information available to the attacker can vary under different scenarios~\citep{jin2020adversarial,sun2020adversarial}. 
Here, we follow the gray-box formalization by \cite{zugner2018adversarial}, where the attacker knows the training set $\setT$, but not the trained GNN $f$. 
The attacker thus considers a surrogate GNN and picks perturbations that maximize an attack loss $\attackLoss$~\citep{zugner2018adversarial,zugner2019adversarial}, assuming that attacks to the surrogate model are transferable to the attacked GNN.
For node classification, the attack loss $\attackLoss$ quantifies how the predictions $\mathbf{z}_v \in [0,1]^{|\setY|}$ made by the GNN $f$ differ from the true labels $\vecy$. 
For a targeted attack of node $v$ with class label $y_v \in \setY$, we adopt the negative classification margin (\textbf{CM-type})~\citep{zugner2018adversarial,xu2019topology}:
    $\attackLoss = - \Delta_c = - (\mathbf{z}_{v, y_v} - \max_{y \neq y_v} \mathbf{z}_{v, y}).$
The attacker usually has additional constraints, such as a limit on the size of the perturbations allowed~\citep{zugner2018adversarial,zugner2019adversarial}.

\paragraph{Taxonomy of attacks}
We follow the taxonomy of attacks introduced in~\citep{jin2020adversarial,sun2020adversarial}. 
For node classification, the attacker may aim to change the classification of a specific node $v\in \vertexSet$ (\textbf{targeted attack}), or to decrease the overall classification accuracy (\textbf{untargeted attack}). 
Attacks can also happen at different stages of the training process: we refer to attacks introduced before training as (pre-training) \textbf{poison attacks}, and attacks introduced after the training process (and before potential retraining on perturbed data) as (post-training) \textbf{evasion attacks}. While our theoretical analysis (\S\ref{sec:theories}) mainly considers targeted evasion attacks, we consider other attacks in our empirical evaluation (\S\ref{sec:exp}). 

\paragraph{Characterizing homophily and heterophily in graphs} 
{Using class labels}, we characterize the types of connections in a graph contributing to its overall level of homophily/heterophily as follows:
\begin{definition}[Homo/Heterophilous path and edge]
\vspace{-0.2cm}
\label{dfn:homophilous-heterophilous-connections}
A \emph{$k$-hop homophilous path} from node $w$ to $u$ is a length-$k$ path  
between endpoint nodes with the same class label $y_w = y_u$. 
Otherwise, the %
path is called \emph{heterophilous}. 
A homophilous or heterophilous edge is a special case with $k=1$.
\end{definition} 

Following \citep{zhu2020beyond,lim2021new}, we define the homophily ratio $h$ as:
\begin{definition}[Homophily ratio]
\vspace{-0.2cm}
\label{dfn:homophily-ratio}
The \emph{homophily ratio} 
is the fraction of homophilous edges among all the edges in a graph: $h = |\{(u,v) \in \edgeSet | y_u = y_v\}|/|\edgeSet|$. 
\end{definition}

\vspace{-0.1cm}
When the edges in a graph are wired randomly, independent to the node labels, the expectation for $h$ is $h_r =  {1}/{|\setY|}$ for balanced classes~\citep{lim2021new}. 
For simplicity, we informally refer to graphs with homophily ratio $h \gg {1}/{|\setY|}$ as \textbf{homophilous graphs} (which have been the focus in most prior works), graphs with homophily ratio $h \ll {1}/{|\setY|}$ as \textbf{heterophilous graphs}, and graphs with homophily ratio $h \approx {1}/{|\setY|}$ as \textbf{weakly heterophilous graphs}.

\section{Relation between Graph Heterophily \&  Model Robustness}
\label{sec:theories}
In this section, we first show theoretical results on the relation between adversarial structural attacks and the change in the homophily level of the underlying graphs. 
Though empirical analyses from previous works have suggested this relation on homophilous graphs~\citep{wu2019adversarial,jin2020adversarial}, to our knowledge, we are the first to formalize it with theoretical analysis and address the case of heterophilous graphs. 
As an implication of the relation, we then discuss how a key design that improves predictive performance of GNNs under heterophily can also help boost their robustness. 
\vspace{-0.2cm}
\subsection{How Do Structural Attacks Change Homophily in Graphs?} 
\label{sec:theories-connections}

\paragraph{Homophilous graphs: structural attacks are mostly hetero\-philous attacks} 
Our first result shows that, for homophilous data, effective structural attacks on GNNs (as measured by loss $\attackLoss$) always result in a reduced level of homophily where either new heterophilous connections are added or existing homophilous connections are removed. 
It also states that direct perturbations on 1-hop neighbors of the target nodes are more effective than indirect perturbations (influencer attacks~\citep{zugner2018adversarial}) on multi-hop neighbors. 
For simplicity, akin to previous works \citep{zugner2018adversarial, zugner2019adversarial} we establish our results for targeted evasion (post-training) attacks in a stylized learning setup with a linear GNN. 
However, our findings generalize to more general setups on real-world datasets as we show in our experiments (\S\ref{sec:exp-perturb-observations}).
{In the theorems below, we use the notion of \emph{\gambit node}: node $u$ is called a gambit if  a perturbation that targets node $v \in \vertexSet$  adjusts the connectivity of node $u \in \vertexSet$.}

\vspace{-0.1cm}
\begin{theorem}
    \label{thm:1} Let $\graph=(\vertexSet,\edgeSet,\matX)$ be a 
    self-loop-free graph with adjacency matrix $\matA$ and node features {\small $\V{x}_v = p\cdot \mathrm{onehot}(y_v) + \tfrac{1-p}{|\setY|}\cdot\mathbf{1}$} for each node $v$, where $\mathbf{1}$ is an all-1 vector, and $p$ is a parameter that regulates the signal to noise ratio. 
    Assume that a fraction $h$ of each node's neighbors belong to the same class, while a fraction {\small $\tfrac{1-h}{|\setY|-1}$} belongs uniformly to any other class.
    Consider a 2-layer linear GNN 
    {\small $f_s^{(2)}(\matA,\matX) = \bar{\matA}^2_\mathrm{s}\matX\matW$} trained on a training set $\setT \subseteq \mathcal{D}_{\vertexSet}$, %
    {with at least one node from each class $y\in\setY$, and degree $d$ for all nodes with a distance less than 2 to any $v\in \mathcal{D}_{\vertexSet}$}. %
    For a unit structural perturbation that %
    {involves a target node $v \in  \mathcal{D}_{\vertexSet}$, and a %
    {correctly classified} \gambit node with degree $d_a$}, the following statements hold if {\small $h \geq \tfrac{1}{|\setY|}$:}
    \begin{enumerate*}%
        \item the attack loss $\attackLoss$ (\S\ref{sec:notation}) of the target $v$ increases only for actions \emph{increasing heterophily}, i.e., when removing a homophilous edge or path, or adding a heterophilous edge or path to node~$v$;
        \item direct perturbations on edges (or 1-hop paths) incident to the target node $v$ lead to greater increase in $\attackLoss$ than indirect perturbations on multi-hop paths to target node~$v$.
    \end{enumerate*}
\end{theorem}

\vspace{-0.2cm}
We give the proof in App.~\ref{app:proof-thm1}. 
Intuitively, the relative inability of existing GNNs to make full use of heterophilous data~\citep{Pei2020Geom-GCN,zhu2020beyond} can be exploited by inserting heterophilous connections in graphs where homophilous ones are expected. Though the theorem shows that effective attacks on homophilous graphs \emph{necessarily} reduce the homophily level, the converse is not true: not all perturbations which reduce the homophily level are effective attacks~\citep{ma2021homophily}.

\paragraph{Heterophilous graphs: structural attacks can be homo\-phil\-ous or hetero\-philous, depending on node degrees} 
When a graph displays heterophily, our analysis shows a more complicated picture on how the level of homophily in the graph is changed
by effective structural attacks: in heterophilous case, the direction of change is dependent on the degrees of %
{both the target node $v$ and the gambit node $u$ of the attack}. 
Specifically, if %
{the degree of \emph{either node}} is low, attacks increasing the heterophily are still effective; however, %
{if the degrees $d$ and $d_a$ of \emph{both nodes} are high}, attacks \emph{decreasing} the heterophily will be effective. Similar to the homophilous case, we formalize our results below for targeted evasion attacks in a stylized learning setup.

\vspace{-0.2cm}
\begin{theorem}
    \label{thm:heterophily}
    Under the setup of Thm.~\ref{thm:1}, 
    for a unit perturbation that %
    {involves a target node $v$ with degree $d$, and a %
    {correctly classified} \gambit node with degree $d_a$}, the following statements hold:
    \begin{enumerate*}%
            \item \emph{\textbf{(Low-degree target node)}} if $0 < d \leq |\setY| - 2$, for any $d_a \geq 0$ and $h\in [0, 1]$, 
            the attack loss $\attackLoss$ (\S\ref{sec:notation}) of $v$ increases only under actions {\emph{increasing heterophily}} in the graph;
            \item \emph{\textbf{(High-degree target node)}} if $d> |\setY| - 2$, conditioning on the degree $d_a$ of the \gambit node:
            \begin{enumerate}
                \item \emph{\textbf{(Low-degree \gambit node)}} if $d_a < \frac{(d+2) (|\setY|-1)}{d-|\setY|+2}$, for any $h\in [0, 1]$, the attack loss $\attackLoss$ (\S\ref{sec:notation}) of $v$ increases only under actions \emph{increasing heterophily} in the graph;
                \item \emph{\textbf{(High-degree \gambit node)}} if $d_a \geq \tfrac{(d+2) (|\setY|-1)}{d-|\setY|+2}$, for $0 \leq h < \tfrac{d_a (d-|\setY|+2)-(d+2) (|\setY|-1)}{(d+1) |\setY| d_a} < \frac{1}{|\setY|}$, $\attackLoss$ (\S\ref{sec:notation}) of $v$ increases only under actions \emph{reducing heterophily}.
            \end{enumerate}
    \end{enumerate*}
    In the statements above, the actions \emph{increasing heterophily} include removing a homophilous edge or adding a heterophilous edge to node $v$, and the actions \emph{reducing heterophily} include adding a homophilous edge or removing a heterophilous edge to node $v$.
\end{theorem}

\vspace{-0.2cm}
{The above theorems cover the situation when the gambit nodes are initially classified correctly (where attacks introducing heterophily can be unambiguously defined using the ground-truth class labels of the nodes involved).
However, in \S\ref{sec:exp-perturb-observations}, we show on real-world datasets that a relaxed interpretation of the theorems, where heterophily is instead defined by the \emph{predicted} class labels of GNNs, 
can explain the behavior of the attacks regardless of the initial correctness of the gambits.}

\vspace{-0.2cm}
\subsection{Boosting Robustness with A Simple Heterophilous Design} 
\label{sec:robustness-design}
A natural follow-up question is whether GNNs with better performance under heterophily are also more robust against structural attacks. 
We deduce that a key design for improving GNN performance for heterophilous data---separate aggregators for ego- and neighbor-embeddings---can also boost the robustness of GNNs by enabling them to better cope with {adversarial changes in heterophily}.

\paragraph{Separate aggregators for ego- and neighbor-embeddings} 
This design uses separate GNN aggregators for ego-embedding $\V{r}_v$ and neighbor-embeddings $\{\V{r}_u: u \in \neighNoSelfLoop(v)\}$. 
Formally, the representation learned for node $v$ in the $k$-th layer is:
{\small
\begin{equation}
    \V{r}^{(k)}_v = \texttt{ENC}
    \left( {\texttt{AGGR1}}(\V{r}^{(k-1)}_v, \V{r}^{(k-2)}_v, ..., \V{r}^{(0)}_v), \; 
    {\texttt{AGGR2}}(\{\V{r}^{(k-1)}_u: u \in {\neighNoSelfLoop(v)}\})
    \right),
    \label{eq:design1}
\end{equation}
}%
where \texttt{AGGR1} and \texttt{AGGR2} are \emph{separate} aggregators, such as averaging functions (GCN), attention mechanisms (GAT), or other pooling mechanisms~\citep{hamilton2017inductive}.
This design has been utilized in existing GNN models 
{(we show examples later in this section)}, 
and has been shown to significantly boost the representation power of GNNs under natural heterophily~\citep{zhu2020beyond}. 
The ego-aggregator \texttt{AGGR1} may also introduce skip connections~\citep{XuLTSKJ18-jkn} to the ego-embeddings aggregated in previous layers as shown in Eq.~\eqref{eq:design1}, {which is another design that further improves the representation power under heterophily~\citep{zhu2020beyond}.}

\paragraph{Intuition}
The key design changes, as compared to the GCN formulation in \S\ref{sec:notation}, allow for the ego-embedding $\V{r}_v$ to be aggregated and weighted \emph{separately} from the neighbor-embeddings $\{\V{r}_u: u \in \neighNoSelfLoop(v)\}$, as well as for the use of skip connections to ego-embeddings of previous layers.
Intuitively, ego-embeddings of feature vectors at the first layer are independent of the graph structure and thus unaffected by adversarial structural perturbations. 
Hence, a separate aggregator and skip connections can provide better access to unperturbed information and mitigate the effects of the attacks.

\paragraph{Theoretical analysis} 
We formalize the above intuition 
that shows how separate aggregators for ego- and neighbor-embeddings enable GNN layers to reduce the attack loss.

\begin{theorem}
    \label{thm:2}
    Under the setup of Thm.~\ref{thm:1}, consider two alternative layers from which a two-layer linear GNN is built: \textbf{(1)} a layer defined as $f_s(\matA,\matX)=\bar{\matA}_\mathrm{s}\matX\matW$; and %
    \textbf{(2)} a layer formulated as $f(\matA,\matX;\alpha) = \left((1-\alpha) \bar{\matA} + \alpha \matI \right) \matX \matW$, which mixes the ego- and neighbor-embedding linearly under a predefined weight $\alpha \in [0, 1]$. 
    Then, for $h > {1}/{|\setY|}$, $\alpha > {1}/{(1+d_a)}$, and a unit perturbation increasing $\attackLoss$ as in Thm.~\ref{thm:1}, outputs of layer $f$ lead to a strictly smaller increase in $\attackLoss$ than $f_s$.
    \vspace{-0.1cm}
\end{theorem}

We provide the proof in App.~\ref{app:proof-thm2};  note that for $\alpha = {1}/{(1+d_a)}$, the two layers are the same: $f(\matA,\matX;\alpha) = f_s(\matA,\matX)$. Theorem~\ref{thm:2} shows that an increase to the weights of ego-embedding 
{(manually or through training)}
improves the robustness of the GNN $f$ for a homophily ratio {$h > {1}/{|\setY|}$}. 
Though aggregators and encoders are stylized 
{in this \emph{simple} instantiation of the design}
in the theorem, the empirical analysis in \S\ref{sec:exp-benchmark-study} confirms that GNNs with more advanced aggregators and encoders,
{which we will discuss next},
also benefit from separate aggregators. Specifically, we find that such GNNs outperform methods without this design by up to 40.00\% and 48.88\% on homophilous and heterophilous graphs, respectively, while performing comparably on clean datasets.

\paragraph{Instantiations of the design on GNNs} %
\label{sec:robustness-design-instantiations}
We demonstrate how the heterophilous design outlined in Eq.~\eqref{eq:design1} 
is instantiated in various GNN models, which are used in our empirical evaluation in \S\ref{sec:exp}. 
In particular, we highlight how these GNN architectures allow separate aggregations of the ego- and neighbor-embeddings. 
\begin{itemize}
    \item In \textbf{H$_2$GCN}~\cite{zhu2020beyond}, a final representation is computed for each node $v\in \vertexSet$ through $\mathbf{r}_{v}^{(\mathrm{final})} =  \texttt{CONCAT}(\mathbf{r}_{v}^{(0)}, \mathbf{r}_{v}^{(1)}, ..., \mathbf{r}_{v}^{(K)})$, where $\mathbf{r}_{v}^{(0)}$ is the non-linear ego-embedding of node features and $\mathbf{r}_{v}^{(k)}$ are the intermediate representations aggregated in the $k$-th layer, where $k \in (1, ..., K)$. By interpreting the update rule's $\texttt{CONCAT}$ as the $\texttt{ENC}$ operation, $\texttt{AGGR1}$ as the skip connection to the ego-embedding of node features, and the concatenation of the intermediate representations as $\texttt{AGGR2}$, the ego- and neighbor-embeddings are separately aggregated as stated in the design.
    
    \item \textbf{GraphSAGE} (with mean aggregator)~\citep{hamilton2017inductive} utilizes a concatenation-based encoding scheme through their update of 
    \vspace{-0.2cm}
    \begin{equation*}
        \mathbf{r}_{v}^{(k)}= \sigma\left(
            \texttt{CONCAT}\left(
                \mathbf{r}_{v}^{(k-1)}, \;
                \texttt{MEAN}\left(
                    \{\mathbf{r}_{u}^{(k-1)}, \forall u\in N(i)\}\right)
            \right) \cdot \mathbf{W}
        \right),
    \vspace{-0.2cm}
    \end{equation*}
    where $\texttt{ENC}(\mathbf{x}_1, \mathbf{x}_2) = \sigma(\texttt{CONCAT}(\mathbf{x}_1, \mathbf{x}_2)\cdot\mathbf{W})$, $\texttt{AGGR1}(\cdot) = \mathbf{r}_{u}^{(k-1)}$, and $\texttt{AGGR2}$ is the mean function. 

    \item \textbf{GPR-GNN}~\citep{chien2021adaptive} embeds each node feature vector separately with a fully connected layer to compute $\mathbf{R}_{v:}^{(0)}$ (or $\mathbf{H}_{v:}^{(0)}$ as in the original paper), similar to \method, and then updates each node's hidden representations through a weighted sum of all $k$-th hop layers around the ego-node, where $k \in (0, 1, ... , K)$. By interpreting the summation as the $\texttt{ENC}$ operation, $\texttt{AGGR1}(\cdot) = \bm{\gamma}_{0}\mathbf{R}^{(0)}$, and $\texttt{AGGR2}(\cdot) = \sum_{k=1}^{K} \bm{\gamma}_{k}{\tilde{\mathbf{A}}^k}_{\mathrm{sym}}\mathbf{R}^{(k-1)}$, where $\bm{\gamma}$ denotes the weights associated with each $k$-hop ego network, the aggregation of the ego- and neighbor-embeddings is decoupled.
    
    \item \textbf{FAGCN}~\citep{bo2021beyond} follows a similar update function to GPR-GNN with 
    \vspace{-0.2cm}
    \begin{equation*}
        \mathbf{r}_{i}^{(l)} = \varepsilon \mathbf{r}_{i}^{(0)} + \sum_{j\in N(i)}  \frac{\alpha_{ij}^{G}}{\sqrt{d_{i}d_{j}}} \mathbf{r}_{j}^{(l-1)}
        \vspace{-0.3cm}
    \end{equation*}
    where $\mathbf{r}_{i}^{(0)}$ (or $\mathbf{h}_{i}^{(0)}$ in the original paper) represents the non-linear ego-embedding and $\alpha_{ij}^{G}$ is a constant measuring the ratio of low and high frequency components. The heterophilous design can similarly be recovered by interpreting the sum as the $\texttt{ENC}$ operation, $\texttt{AGGR1}(\cdot) = \varepsilon \mathbf{r}_{i}^{(0)}$ as a weighted skip connection to the ego-embedding of features, and the weighted sum of embeddings within the neighborhood $N(i)$ of node $i \in \vertexSet$ as $\texttt{AGGR2}(\cdot)$.
    
    \item \textbf{CPGNN}~\citep{zhu2020graph} formulates the update function of belief vectors $\mathbf{R}^{(k)}$ after the $k$-th propagation layer as
    $
        \mathbf{R}^{(k)} = \mathbf{R}^{(0)} + \mathbf{A} \mathbf{R}^{(k-1)} \bar{\mathbf{H}}
    $, where $\mathbf{R}^{(0)}$ ($\bar{\mathbf{B}}^{(0)}$ in the original paper) consists of prior belief vectors for each node (as the ego-embeddings $\mathbf{r}_{i}^{(0)}$ in Eq.~\eqref{eq:design1}), and $\bar{\mathbf{H}}$ is the learnable compatibility matrix. The heterophilous design is recovered by letting $\texttt{AGGR1}(\cdot) = {\mathbf{R}}^{(0)}$ as a skip connection, $\texttt{AGGR2}(\cdot) = \mathbf{A} {\mathbf{R}}^{(k-1)} \bar{\mathbf{H}}$, and the $\texttt{ENC}$ operation as the summation.

    \item \textbf{APPNP}~\cite{klicpera2018predict} first generates predictions $\mathbf{R}_{v:}^{(0)}$ (or $\mathbf{H}_{v:}^{(0)}$ as in the original paper) of each node $v$ based on its own feature, then updates the predictions through power iterations of Personalized PageRank. More specifically, the $k$-th iteration step is formulated as $\mathbf{R}^{(k)} = (1 - \alpha) {\tilde{\mathbf{A}}}_{\mathrm{sym}}\mathbf{R}^{(k-1)} + \alpha \mathbf{R}^{(0)}$. The heterophilous design can be recovered by letting $\texttt{AGGR1}(\cdot) = \mathbf{R}^{(0)}$ as a skip connection to the initial prediction, $\texttt{AGGR2}(\cdot) = {\tilde{\mathbf{A}}}_{\mathrm{sym}}\mathbf{R}^{(k-1)}$, and the summation weighted by $\alpha$ as the $\texttt{ENC}$ operation.
\end{itemize}

\section{Related Work}

\paragraph{Adversarial attacks and defense strategies for graphs} 
Since \nettack~\citep{zugner2018adversarial} and RL-S2V~\citep{dai2018adversarial} first demonstrated the vulnerabilities of GNNs against adversarial perturbations, a variety of attack strategies under different scenarios have been proposed,
including adversarial attacks on the graph structure~\citep{dai2018adversarial,xu2019topology,bojchevski2019adversarial,li2020adversarial,chang2020restricted}, node features \citep{takahashi2019indirect,ma2020towards}, or combinations of both~\citep{zugner2018adversarial,zugner2019adversarial,wu2019adversarial}. 
On the defense side, 
various techniques for 
improving the GNN robustness against adversarial attacks 
have been proposed, including: 
adversarial training~\citep{xu2019topology,zugner2019adversarial,bojchevski2019certifiable};  
RGCN~\citep{zhu2019robust}, which adopts Gaussian-based embeddings and a variance-based attention mechanism;
low-rank approximation of graph adjacency~\citep{entezari2020all} against Nettack~\citep{zugner2018adversarial};
Pro-GNN~\citep{jin2020graph}, which estimates the unperturbed graph structure in training with the assumptions of low-rank, sparsity, and homophily of node features; GCN-Jaccard~\citep{wu2019adversarial} and GNNGuard~\citep{zhang2020gnnguard}, which assume homophily of features (or structural embeddings) and train GNN models on a pruned graph with only strong homophilous links;
and Soft Medoid~\citep{geisler2020reliable}, an aggregation function with improved robustness. Other recent works have looked into the certification of nodes that are guaranteed to be robust against certain structural and feature perturbations~\citep{zugner2019certifiable,bojchevski2019certifiable,zugner2020certifiable}, including approaches based on model-agnostic randomized smoothing~\citep{cohen2019certified,lee2019tight,bojchevski2020efficient}. 
Interested readers can refer to the recent surveys~\citep{jin2020adversarial,sun2020adversarial} for a comprehensive review. 

\paragraph{GNNs \& Heterophily}
Recent works~\citep{Pei2020Geom-GCN,liu2020non,zhu2020beyond,ma2021homophily} have shown that heterophilous datasets can lead to significant performance loss for popular GNN architectures 
(e.g., GCN~\citep{kipf2016semi},  GAT~\citep{velickovic2018graph}).
This issue is also known in classical semi-supervised learning~\citep{peel2017graph}.
To address this issue, several GNN designs for handling heterophilous connections have been proposed~\citep{MixHop,Pei2020Geom-GCN,zhu2020beyond,dong2021graph,li2021beyond,zhu2020graph,bo2021beyond}.
\citet{yan2021two} recently discussed the connection between heterophily and oversmoothing for GNNs, and  designs to address both issues;
{\cite{donald2022fairness} studied how locally-occuring heterophily affects fairness of GNNs.}
However, the formal connection between heterophily and robustness of GNNs
has received little attention.
Here we focus on a simple yet powerful design that significantly improves performance under heterophily~\citep{zhu2020beyond}, and can be readily incorporated into GNNs. 

\section{Empirical Evaluation}
\label{sec:exp}

Our analysis seeks to answer the following questions:
(Q1) Does our theoretical analysis on the relations between adversarial attacks and changes in heterophily level generalize to real-world datasets? 
(Q2) Do heterophily-adjusted GNNs, i.e., models with separate aggregators for ego- and neighbor-embeddings,
show improved robustness against state-of-the-art attacks? 
(Q3) Does the identified design improve the \emph{certifiable} robustness of GNNs? 

First, we describe the experimental setup and datasets that we use to answer the above questions. 

\label{sec:exp-setups}

\vspace{0.1cm}
\paragraph{Attack Setup} We consider both targeted and untargeted attacks (\S\ref{sec:notation}), generated by \nettack~\citep{zugner2018adversarial} and Metattack~\citep{zugner2019adversarial}, respectively.
For each attack method, we consider poison (pre-training) and evasion (post-training) attacks, yielding 4 attack scenarios in total.
We focus on robustness against structural perturbations and keep the node features unchanged. 
We randomly generate 3 sets of perturbations per attack method and dataset, and consistently evaluate each GNN model on them.
For \nettack, we randomly select 60 nodes from the graph as the target nodes for each set of perturbations, instead of the GCN-based target selection approach as in \citep{zugner2018adversarial}:
the approach in \citep{zugner2018adversarial} only selects nodes that are correctly classified by GCN~\citep{kipf2016semi} on clean data;
this introduces unfair advantage towards GCN, especially on heterophilous datasets where GCN can exhibit significantly inferior accuracy to models like GraphSAGE~\citep{zhu2020beyond}. 
For the experiments in \S\ref{sec:exp-perturb-observations}, we use a budget of 1 perturbation per target node to match the setup of our theorems; for the benchmark study in \S\ref{sec:exp-benchmark-study}, we use an attack budget equal to a node's degree and allow direct attacks on target nodes. 
For Metattack, we budget the attack as 20\% of the number of edges in each dataset, and we use the Meta-Self variant as it shows the most destructiveness~\citep{zugner2019adversarial}.

\paragraph{GNN Models} 
To show the effectiveness of our identified design, we evaluate four groups of models against adversarial attacks: 
\textbf{(1)}~Baseline models without any vaccination, including some of the most popular methods: GCN~\citep{kipf2016semi},  GAT~\citep{velickovic2018graph},
and the graph-agnostic multilayer perceptron (MLP) which relies only on node features; 
\textbf{(2)}~State-of-the-art ``vaccinated'' baselines designed with robustness in mind: 
ProGNN~\citep{jin2020graph}, GNNGuard~\citep{zhang2020gnnguard}, GCN-SVD~\citep{entezari2020all} and GCN-SMGDC, which adopts the Soft Medoid aggregator~\cite{geisler2020reliable} and GDC~\cite{klicpera_diffusion_2019} on GCN~\cite{kipf2016semi} architecture;  
\textbf{(3)}~Models with the heterophilous design only: GraphSAGE~\citep{hamilton2017inductive}, \method~\citep{zhu2020beyond}, CPGNN~\citep{zhu2020graph}, GPR-GNN~\citep{chien2021adaptive} FAGCN~\citep{bo2021beyond} and APPNP~\cite{klicpera2018predict}; 
we discussed how these models instantiate this design in \S\ref{sec:robustness-design-instantiations}; 
\textbf{(4)}~{
Models with both the %
heterophilous design and explicit robustness-enhancing mechanisms, where we adopt two existing mechanisms: (i)~SVD-based low-rank approxmiation~\citep{entezari2020all} (H$_2$GCN-SVD and GraphSAGE-SVD), 
and (ii)~Soft Medoid aggregator~\cite{geisler2020reliable} with GDC~\cite{klicpera_diffusion_2019} (\method-SMGDC and GraphSAGE-SMGDC).
We combine both these mechanisms with heterophily-adjusted GNNs instead of non-adjusted models (e.g., GCN)---detailed formulations are given on our \repo{}. 
}
We set the number of layers as 2 and the size of hidden units per layer as 64 for all models to ensure a fair comparison between different architectures and designs.
We provide more implementation details and hyperparameter settings on our \repo{} (App. \S\ref{app:exp-details}).

\begin{table}[t]
    \caption{Dataset statistics.}
    \label{table:dataset-stats}
    \vspace{-0.4cm}
    \resizebox{\columnwidth}{!}
    {%
    \begin{tabular}{Hl c  c  c  c  c  c}
        \toprule
        & & \multicolumn{3}{c}{\bf Homophilous} && \multicolumn{2}{c}{\bf Heterophilous} \\
        \cmidrule{3-5} \cmidrule{7-8}  
        & & \multicolumn{1}{c}{\texttt{\bf Cora}} & \multicolumn{1}{c}{\texttt{\bf Pubmed}} & \multicolumn{1}{c}{\texttt{\bf Citeseer}}  && \multicolumn{1}{c}{\texttt{\bf FB100}} & \multicolumn{1}{c}{\texttt{\bf Snap}} \\
        \midrule
        & \textbf{\#Nodes} $|\vertexSet|$ & \multicolumn{1}{c}{2,485} & \multicolumn{1}{c}{19,717} & \multicolumn{1}{c}{2,110} && \multicolumn{1}{c}{2,032} & \multicolumn{1}{c}{4,562}\\
        & \textbf{\#Edges}  $|\edgeSet|$ & \multicolumn{1}{c}{5,069} & \multicolumn{1}{c}{44,324} & \multicolumn{1}{c}{3,668} && \multicolumn{1}{c}{78,733} & \multicolumn{1}{c}{12,103}\\
        & \textbf{\#Classes} $|\setY|$ & \multicolumn{1}{c}{7} & \multicolumn{1}{c}{3} & \multicolumn{1}{c}{6} && \multicolumn{1}{c}{2} & \multicolumn{1}{c}{5}\\
        & \textbf{\#Features}  $F$ & \multicolumn{1}{c}{1,433} & \multicolumn{1}{c}{500} & \multicolumn{1}{c}{3,703} && \multicolumn{1}{c}{1,193} & \multicolumn{1}{c}{269}\\
        & \textbf{Homophily} $h$ & \multicolumn{1}{c}{0.804} & \multicolumn{1}{c}{0.802} & \multicolumn{1}{c}{0.736} && \multicolumn{1}{c}{0.531} & \multicolumn{1}{c}{0.134}  \\
    \bottomrule \end{tabular}}
    \vspace{-0.6cm}
\end{table}

\paragraph{Datasets \& Evaluation Setup} We consider {three widely-used citation networks~\citep{citeseer_dataset,namata2012query}} with strong homophily---Cora~\citep{cora_dataset}, {Pubmed, and} Citeseer---along with one weakly and one strongly heterophilous graph, introduced by \citet{lim2021new}: FB100~\citep{FB100Source} and Snap Patents~\citep{SnapSource1, snapnets}. 
We report summary statistics in Table~\ref{table:dataset-stats}, and provide more details %
on our \repo.
For computational tractability, we subsample the Snap Patents data via snowball sampling~\citep{Leo1961Snowball}, where we keep 20\% of the neighbors for each traversed node; we give detailed algorithm on our \repo.
{The sizes of the datasets that we used in our experiments are similar to those in previous works on GNN robustness~\cite{geisler2020reliable, jin2020graph}.}
We follow the evaluation procedure of \cite{zugner2018adversarial, jin2020graph} to split the nodes of each dataset into training (10\%), validation (10\%) and test (80\%) data, and determine the model parameters on training and validation splits.
We report the average performance and standard deviation on the 3 sets of generated perturbations. 
For targeted attacks with \nettack, we report the classification accuracy on the target nodes; for untargeted attacks with Metattack, we report it over the whole test data.

\paragraph{Robustness Certificates} We adopt randomized smoothing for GNNs~\citep{bojchevski2020efficient} to evaluate the certifiable robustness, with parameter choices detailed in our \githubrepo.
We only consider structural perturbations in the randomization scheme. 
Following \citet{geisler2020reliable}, we measure the certifiable robustness of GNN models with the accumulated certifications (AC) and the average maximum certifiable radii for edge additions ($\bar{r}_a$) and deletions ($\bar{r}_d$) over all correctly predicted nodes. 
More specifically, AC is defined as
$-R(0, 0) + \sum_{r_a, r_d \geq 0} R(r_a, r_d)$,
where $R(r_a, r_d)$ is the \emph{certifiably correct ratio}, i.e., the ratio of the nodes in the test splits that are \emph{both} predicted correctly by the smoothed classifier \emph{and} certifiably robust at radius $(r_a, r_d)$.
In addition, we report the accuracy of each model with randomized smoothing enabled on the test splits of the clean datasets, which is equal to $R(0, 0)$.
We report the average and standard deviation of each statistic over the 3 different training, validation and test splits. %

\paragraph{Hardware Specifications} 
We use a workstation with a 12-core AMD Ryzen 9 3900X CPU, 64GB RAM, and a Quadro P6000 GPU with 24 GB GPU Memory. 

\paragraph{Code and Additional Details} Code and additional details on the setups and results are available on GitHub repository: \githubrepourl.

\vspace{-0.2cm}
\subsection{
{(Q1) Structural Attacks are Mostly Heterophilous: Empirical Validation}}
\label{sec:exp-perturb-observations}

\begin{figure}[t]
    \centering
    \includegraphics[width=0.9\columnwidth, keepaspectratio, trim={0 0 0 0.6cm}]{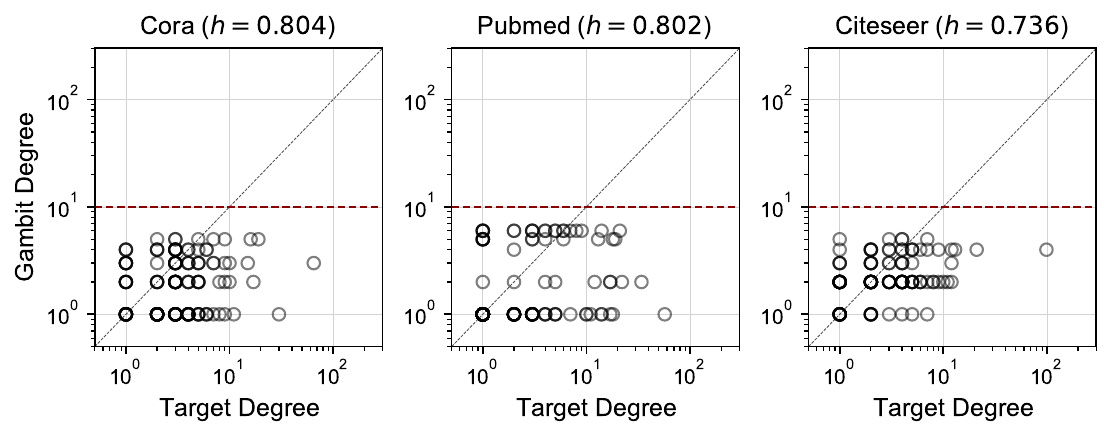}\\
    \vspace{0.2cm}
    \includegraphics[width=0.6\columnwidth, keepaspectratio, trim={0 0 0 0.6cm}]{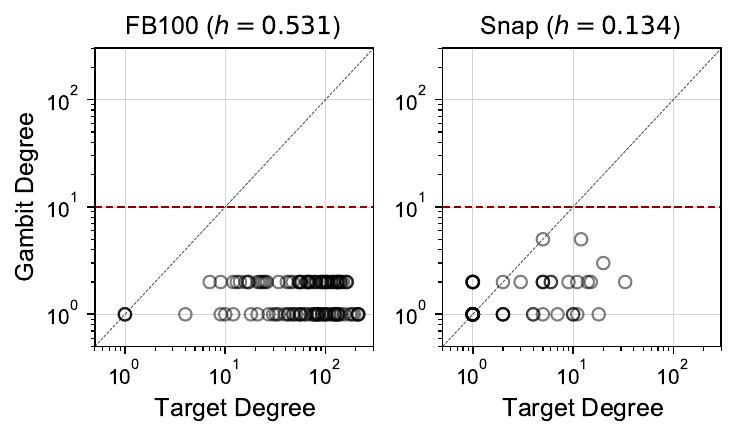}
    \vspace{-0.3cm}
    \caption{Scatter plots of the degrees of the target nodes (x-axis) and gambit nodes (y-axis) involved in the targeted attacks (\S\ref{sec:exp-perturb-observations}). Attacks tend to leverage gambit nodes with low degrees, 
    which makes attacks increasing heterophily effective for heterophilous graphs following Thm.~\ref{thm:heterophily}. }
    \label{fig:nettack-degree-scatter}
    \vspace{-0.5cm}
\end{figure}

To show that our theoretical analysis in \S\ref{sec:theories-connections} generalizes to more complex settings beyond the assumptions we made in the theorems, we look into effective targeted attacks made by \nettack{} on real-world homophilous and heterophilous datasets, and present statistics of the attacks in Table~\ref{tab:perturbation-stats}, with a focus on the ratios of heterophilous attacks.
We use a budget of 1 perturbation per target node in this experiment, and the statistics are reported among all effective perturbations
targeting nodes that are correctly classified on clean datasets by the surrogate GNN of \nettack{} (i.e., GCN) as described in \S\ref{sec:exp-setups}.
To validate the dependency between the degrees of the target/gambit nodes and the changes of heterophily predicted by Thm.~\ref{thm:heterophily}, we also show the scatter plots of node degrees in Fig.~\ref{fig:nettack-degree-scatter}.

\begin{table}[t]
    \centering
    \caption{Effective targeted attacks by \nettack~(\S\ref{sec:exp-perturb-observations}): ratios of edge additions, deletions and heterophilous attacks (i.e., attacks increasing heterophily). 
    We consider two heterophily definitions, one based on ground-truth class labels (Label), and the other on predicted class labels by GCN on clean datasets (Pred.).
    All attacks are direct perturbations on edges incident to the targets.
    Degrees of target and gambit nodes in the attacks are shown in Fig.~\ref{fig:nettack-degree-scatter}. 
    All attacks introduce heterophilous edges that connect nodes with different \emph{predicted} labels, following the takeaways of Thm.~\ref{thm:1} and \ref{thm:heterophily}. 
    }
    \label{tab:perturbation-stats}
    \ifdefmacro{\ispreprint}{}{\vspace{-0.15cm}}
    \resizebox{\columnwidth}{!}{
        \begin{tabular}{l l l rr c rr}
        \toprule
        
        & \multirow{2.5}{*}{\textbf{Dataset}} 
        & \multirow{2.5}{*}{\textbf{\shortstack[l]{Sample\\Sizes}}}
        & \multicolumn{2}{c}{\textbf{Attack Type}} & 
        & \multicolumn{2}{c}{\textbf{Hete. Attacks}} 
        \\
        \cmidrule{4-5} \cmidrule{7-8}
        & & & Add. & Del. && Label & Pred. \\
        \midrule
        \multirow{4}{*}{\rotatebox[origin=r]{90}{\nettack}} 
        & Cora & 150 & 99.33\% & 0.67\% &  & 100.00\% & 100.00\%\\
        & Pubmed & 153 & 100.00\% & 0.00\% &  & 100.00\% & 100.00\%\\
        & Citeseer & 121 & 100.00\% & 0.00\% &  & 100.00\% & 100.00\%\\
        
        \noalign{\vskip 0.25ex}
        \cdashline{2-8}[0.8pt/2pt]
        \noalign{\vskip 0.25ex}
        
        & FB100 & 112 & 100.00\% & 0.00\% &  & 50.00\% & 100.00\%\\
        & Snap & 51 & 100.00\% & 0.00\% &  & 64.71\% & 100.00\%\\

        \bottomrule
        \end{tabular}
    }
    \vspace{-0.01cm}
\end{table}

\paragraph{Homophilous Networks} For the strongly homophilous {Cora, Pubmed and Citeseer graphs, all changes introduced by effective attacks in the graph structure} follow the conclusion of Thm.~\ref{thm:1}: they reduce homophily (increase heterophily) by adding heterophilous edges or removing homophilous edges. 
These results show that despite the simplified analysis, the takeaway of Thm.~\ref{thm:1} can be generalized to real-world datasets. 
In addition, the attacks mostly introduce, rather than prune, edges, suggesting that attacks adding outlier edges to the graph are more powerful than attacks removing informative existing edges. 
These observations in our experiments are consistent with the observations from previous works~\citep{jin2020adversarial,geisler2020reliable}.

\paragraph{Heterophilous Networks} For heterophilous graphs FB100 ($h \approx {1}/{|\setY|}$) and Snap ($h < {1}/{|\setY|}$), 
Fig.~\ref{fig:nettack-degree-scatter} %
shows that almost all attacks leverage gambit nodes with low degrees (1 or 2); no node with degree higher than 5 is leveraged.
All attacks leveraging correctly classified gambit nodes are connecting node $u \in \vertexSet$ with a different ground-truth class label $y_u \neq y_v$ to the target nodes $v \in \vertexSet$; attacks leveraging incorrectly classified gambit nodes are always connecting node $u$ with a different \emph{predicted} class label $\hat{y}_u \neq \hat{y}_v = y_v $ to the target node $v$, even though some gambit nodes have the same \emph{ground-truth} class label $y_u = y_v \neq \hat{y}_u$ as the target nodes. 
These results validate the conclusion of Thm.~\ref{thm:heterophily} on correctly classified gambit nodes, and demonstrate its generalizability under the heterophily definition based on predicted class labels. Note that the predicted class labels $\hat{y}_u$ for each node $u \in \vertexSet$ are based on GCN, which is the surrogate GNN used by \nettack{}.

\vspace{-0.4cm}
\subsection{(Q2) Benchmark Study of GNN Models: Heterophilous Design Leads to Improved Empirical Robustness}
\label{sec:exp-benchmark-study}

\begin{figure}[t]
    \vspace{0.05cm}
    \begin{center}
    \includegraphics[width=\linewidth,keepaspectratio]{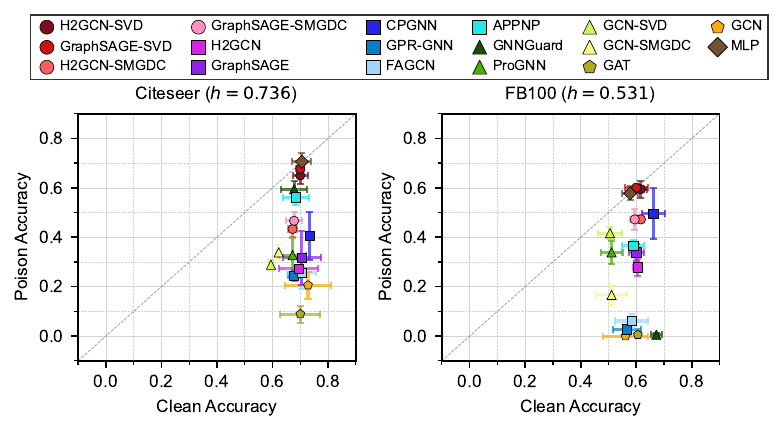}
    \end{center}
    \vspace{-0.5cm}
    \caption{(Best viewed in color.) Classification accuracy on clean data and against poison attacks for target nodes attacked by \textsc{Nettack}. Error bars show standard deviation across different sets of experiments. Detailed results are listed in 
    Table \ref{table:real-results-poison-only}. 
    As expected, MLP is not influenced by the adversarial structural attacks.}
    
    \label{fig:nettack_cora_FB100}
    \ifdefmacro{\ispreprint}{}{\vspace{-0.5cm}}
\end{figure}

To answer (Q2) on whether heterophily-adjusted GNN models show improved performance against state-of-the-art attacks, we conduct a comprehensive benchmark study. 
We consider all four categories of GNN models mentioned in \S\ref{sec:exp-setups}, and evaluate their robustness against both targeted and untargeted attacks. 
We report the hyperparameters for each method 
on our \repo{} (App. \S\ref{app:exp-details-params}). 
Table~\ref{table:real-results-poison-only} shows the performance of each method under poison (pre-training) attacks and on clean (unperturbed) data, and
Fig.~\ref{fig:nettack_cora_FB100} visualizes the corresponding performance changes relative to the clean datasets. 
For conciseness, we report {additional results on Pubmed in Table~\ref{table:results-pubmed}}, and under evasion (post-training) attacks on our \githubrepo~(Table~\ref{table:real-results-detailed-netattack} and \ref{table:real-results-detailed-metattack-20p}), where
we also discuss how our simple heterophilous design leads to only minor computational overhead compared to existing vaccination mechanisms (App. \S\ref{app:real-runtime-complexity}). 

\begin{table*}[t]

    \centering
    \caption{Benchmark study: %
    {mean accuracy $\pm$ stdev against poison attacks, with accuracy on clean datasets in gray for reference}. Accuracy is reported on target nodes for \nettack, and on full test splits for Metattack. Best GNN performance against attacks is highlighted in blue per dataset, and in gray per model group. MLP is immune to structural attacks and not considered as a GNN model. 
    Accuracy against evasion attacks are listed on our \githubrepo{} (App. \S\ref{app:real-benchmark-results}), 
    and the setups in \S\ref{sec:exp-setups}. 
    Additional results on Pubmed are listed in App. Table~\ref{table:results-pubmed}.
    GNNs merely adopting this design achieve up to 40.00\% improvement in accuracy against \nettack{} compared to the best-performing unvaccinated model (GCN).
    Additionally, methods combining this design 
    alongside explicit defense mechanisms (e.g., GraphSAGE-SVD) achieve further robustness improvement to the corresponding base mechanism without the design (e.g., GCN-SVD), and outperform the best vaccinated baseline by up to 18.33\%.
    }
    \label{table:real-results-poison-only}
    \ifdefmacro{\ispreprint}{}{\vspace{-0.15cm}}
    \resizebox{\linewidth}{!}{
    \begin{tabular}{ll cc c>{\color{dark-gray}}c c c>{\color{dark-gray}}c c c>{\color{dark-gray}}c c c>{\color{dark-gray}}c c c>{\color{dark-gray}}c c c>{\color{dark-gray}}c c c>{\color{dark-gray}}c c c>{\color{dark-gray}}c}
        \toprule
        &  &  &
        
        & \multicolumn{11}{c}{\textsc{ Nettack}}  
        && \multicolumn{11}{c}{\texttt{\bf Metattack}} \\ \cmidrule{5-15} \cmidrule{17-27}
        
        & \multirow{3}{*}{\rotatebox[origin=r]{90}{\textbf{Hetero.}}} & \multirow{3}{*}{\rotatebox[origin=r]{90}{\textbf{Vaccin.}}} &
        
        & \multicolumn{2}{c}{\texttt{\bf Cora}} && \multicolumn{2}{c}{\texttt{\bf Citeseer}} 
        && \multicolumn{2}{c}{\texttt{\bf FB100}} && \multicolumn{2}{c}{\texttt{\bf Snap}} 
        && \multicolumn{2}{c}{\texttt{\bf Cora}} && \multicolumn{2}{c}{\texttt{\bf Citeseer}} 
        && \multicolumn{2}{c}{\texttt{\bf FB100}} && \multicolumn{2}{c}{\texttt{\bf Snap}}\\
        
        & & & & \multicolumn{2}{c}{$h$=0.804} && \multicolumn{2}{c}{$h$=0.736} && \multicolumn{2}{c}{$h$=0.531} && \multicolumn{2}{c}{$h$=0.134} && \multicolumn{2}{c}{$h$=0.804} && \multicolumn{2}{c}{$h$=0.736} && \multicolumn{2}{c}{$h$=0.531} && \multicolumn{2}{c}{$h$=0.134}  \\
        
        \cmidrule{5-6}  \cmidrule{8-9}  \cmidrule{11-12} \cmidrule{14-15} \cmidrule{17-18} \cmidrule{20-21} \cmidrule{23-24} \cmidrule{26-27}
        
        & & & & \multicolumn{1}{c}{Poison} & \multicolumn{1}{c}{Clean} && 
                \multicolumn{1}{c}{Poison} & \multicolumn{1}{c}{Clean} &&
                \multicolumn{1}{c}{Poison} & \multicolumn{1}{c}{Clean} &&
                \multicolumn{1}{c}{Poison} & \multicolumn{1}{c}{Clean} &&
                \multicolumn{1}{c}{Poison} & \multicolumn{1}{c}{Clean} && 
                \multicolumn{1}{c}{Poison} & \multicolumn{1}{c}{Clean} &&
                \multicolumn{1}{c}{Poison} & \multicolumn{1}{c}{Clean} &&
                \multicolumn{1}{c}{Poison} & \multicolumn{1}{c}{Clean} \\

        \cmidrule{1-3} \cmidrule{5-6}  \cmidrule{8-9}  \cmidrule{11-12} \cmidrule{14-15} \cmidrule{17-18} \cmidrule{20-21} \cmidrule{23-24} \cmidrule{26-27}
        \textbf{H$_2$GCN-SVD} & \checkmark & \checkmark  &    &   $\underset{\scriptscriptstyle{\pm 2.72}}{70.00}$ &  $\underset{\scriptscriptstyle{\pm 3.42}}{74.44}$ & &  $\underset{\scriptscriptstyle{\pm 3.60}}{65.00}$ &            $\underset{\scriptscriptstyle{\pm 2.72}}{70.00}$ &
        
        &  $\underset{\scriptscriptstyle{\pm 3.42}}{59.44}$ & $\underset{\scriptscriptstyle{\pm 2.36}}{61.67}$& & \cellcolor{blue!20} $\underset{\scriptscriptstyle{\pm 3.42}}{28.89}$ &                $\underset{\scriptscriptstyle{\pm 2.08}}{30.56}$ 
        
        && $\underset{\scriptscriptstyle{\pm 0.47}}{67.87}$ &  $\underset{\scriptscriptstyle{\pm 0.37}}{76.89}$& &        \cellcolor{blue!20}$\underset{\scriptscriptstyle{\pm 0.46}}{70.42}$ &      $\underset{\scriptscriptstyle{\pm 1.03}}{73.42}$& &     \cellcolor{gray!20}$\underset{\scriptscriptstyle{\pm 0.08}}{56.72}$ &   $\underset{\scriptscriptstyle{\pm 0.77}}{56.81}$& &                        $\underset{\scriptscriptstyle{\pm 0.14}}{25.60}$ &                       $\underset{\scriptscriptstyle{\pm 0.26}}{27.63}$
        
        \\

        \textbf{GraphSAGE-SVD} & \checkmark & \checkmark & & \cellcolor{blue!20}$\underset{\scriptscriptstyle{\pm 2.36}}{71.67}$ &  $\underset{\scriptscriptstyle{\pm 4.78}}{77.22}$& &    \cellcolor{blue!20}$\underset{\scriptscriptstyle{\pm 3.42}}{67.78}$ &  $\underset{\scriptscriptstyle{\pm 1.36}}{70.00}$& &   \cellcolor{blue!20}$\underset{\scriptscriptstyle{\pm 1.36}}{60.00}$ &  $\underset{\scriptscriptstyle{\pm 4.08}}{60.00}$& &                    $\underset{\scriptscriptstyle{\pm 6.80}}{26.67}$ &                   $\underset{\scriptscriptstyle{\pm 5.50}}{27.22}$ 
        
        && \cellcolor{gray!20}$\underset{\scriptscriptstyle{\pm 1.32}}{68.86}$ &  $\underset{\scriptscriptstyle{\pm 0.29}}{77.52}$& &        $\underset{\scriptscriptstyle{\pm 0.52}}{69.10}$ &      $\underset{\scriptscriptstyle{\pm 0.17}}{72.16}$& &     $\underset{\scriptscriptstyle{\pm 0.33}}{55.76}$ &   $\underset{\scriptscriptstyle{\pm 0.86}}{57.38}$& &                        \cellcolor{gray!20}$\underset{\scriptscriptstyle{\pm 0.30}}{26.58}$ &                       $\underset{\scriptscriptstyle{\pm 0.70}}{26.72}$
        
        \\
        
        \textbf{H$_2$GCN-SMGDC} & \checkmark & \checkmark & & $\underset{\scriptscriptstyle{\pm 4.37}}{59.44}$ &  $\underset{\scriptscriptstyle{\pm 4.78}}{77.22}$& &    $\underset{\scriptscriptstyle{\pm 3.60}}{43.33}$ &  $\underset{\scriptscriptstyle{\pm 1.57}}{67.22}$& &   $\underset{\scriptscriptstyle{\pm 1.57}}{47.22}$ &  $\underset{\scriptscriptstyle{\pm 0.00}}{61.67}$& &                    $\underset{\scriptscriptstyle{\pm 1.57}}{22.22}$ &                   $\underset{\scriptscriptstyle{\pm 0.79}}{30.56}$ 
        
        && $\underset{\scriptscriptstyle{\pm 1.65}}{66.50}$ &  $\underset{\scriptscriptstyle{\pm 0.33}}{80.60}$& &        $\underset{\scriptscriptstyle{\pm 1.24}}{69.04}$ &      $\underset{\scriptscriptstyle{\pm 0.92}}{74.31}$& &     $\underset{\scriptscriptstyle{\pm 1.51}}{54.63}$ &   $\underset{\scriptscriptstyle{\pm 0.10}}{56.52}$& &                        $\underset{\scriptscriptstyle{\pm 1.09}}{24.41}$ &                       $\underset{\scriptscriptstyle{\pm 0.62}}{27.50}$
        
        \\
        
        \textbf{GraphSAGE-SMGDC} & \checkmark & \checkmark & & $\underset{\scriptscriptstyle{\pm 8.28}}{56.67}$ &  $\underset{\scriptscriptstyle{\pm 5.44}}{78.33}$& &    $\underset{\scriptscriptstyle{\pm 3.60}}{46.67}$ &  $\underset{\scriptscriptstyle{\pm 2.83}}{67.78}$& &   $\underset{\scriptscriptstyle{\pm 4.16}}{47.22}$ &  $\underset{\scriptscriptstyle{\pm 1.57}}{59.44}$& &                    $\underset{\scriptscriptstyle{\pm 3.14}}{20.56}$ &                   $\underset{\scriptscriptstyle{\pm 4.16}}{29.44}$ 
        
        && $\underset{\scriptscriptstyle{\pm 2.07}}{66.95}$ &  $\underset{\scriptscriptstyle{\pm 0.26}}{79.39}$& &        $\underset{\scriptscriptstyle{\pm 0.97}}{68.68}$ &      $\underset{\scriptscriptstyle{\pm 0.38}}{74.31}$& &     $\underset{\scriptscriptstyle{\pm 0.29}}{55.39}$ &   $\underset{\scriptscriptstyle{\pm 0.19}}{55.19}$& &                        $\underset{\scriptscriptstyle{\pm 0.76}}{25.21}$ &                       $\underset{\scriptscriptstyle{\pm 0.29}}{26.38}$
        
        \\

        \noalign{\vskip 0.25ex}
        \cdashline{1-3}[0.8pt/2pt]
        \cdashline{5-15}[0.8pt/2pt]
        \cdashline{17-27}[0.8pt/2pt]
        \noalign{\vskip 0.25ex}
        
        \textbf{H$_2$GCN} & \checkmark &         & &  $\underset{\scriptscriptstyle{\pm 5.50}}{38.89}$ &  $\underset{\scriptscriptstyle{\pm 8.31}}{82.78}$& &    $\underset{\scriptscriptstyle{\pm 1.57}}{27.22}$ &  $\underset{\scriptscriptstyle{\pm 6.98}}{69.44}$& &   $\underset{\scriptscriptstyle{\pm 3.42}}{27.78}$ &  $\underset{\scriptscriptstyle{\pm 1.57}}{60.56}$& &                    $\underset{\scriptscriptstyle{\pm 2.83}}{12.78}$ &                   $\underset{\scriptscriptstyle{\pm 2.72}}{30.00}$ && 
        
        $\underset{\scriptscriptstyle{\pm 6.61}}{57.75}$ &  $\underset{\scriptscriptstyle{\pm 0.97}}{83.94}$& &        $\underset{\scriptscriptstyle{\pm 0.82}}{54.34}$ &      $\underset{\scriptscriptstyle{\pm 0.90}}{75.34}$& &     $\underset{\scriptscriptstyle{\pm 0.76}}{54.84}$ &   $\underset{\scriptscriptstyle{\pm 0.13}}{56.95}$& &                        $\underset{\scriptscriptstyle{\pm 0.59}}{25.34}$ &                       $\underset{\scriptscriptstyle{\pm 0.05}}{27.49}$
        
        \\
        
        \textbf{GraphSAGE} & \checkmark &     & & $\underset{\scriptscriptstyle{\pm 2.72}}{36.67}$ &  $\underset{\scriptscriptstyle{\pm 9.56}}{82.22}$& &   $\underset{\scriptscriptstyle{\pm 10.89}}{31.67}$ &  $\underset{\scriptscriptstyle{\pm 6.85}}{70.56}$& &   $\underset{\scriptscriptstyle{\pm 3.42}}{33.89}$ &  $\underset{\scriptscriptstyle{\pm 2.72}}{60.00}$& &                    $\underset{\scriptscriptstyle{\pm 7.07}}{16.67}$ &                   $\underset{\scriptscriptstyle{\pm 4.16}}{24.44}$ 
        
        && $\underset{\scriptscriptstyle{\pm 2.56}}{54.68}$ &  $\underset{\scriptscriptstyle{\pm 0.63}}{82.21}$& &        $\underset{\scriptscriptstyle{\pm 1.74}}{59.74}$ &      $\underset{\scriptscriptstyle{\pm 0.93}}{74.64}$& &     $\underset{\scriptscriptstyle{\pm 0.83}}{54.72}$ &   $\underset{\scriptscriptstyle{\pm 1.40}}{56.60}$& &                        $\underset{\scriptscriptstyle{\pm 0.76}}{24.14}$ &                       $\underset{\scriptscriptstyle{\pm 0.84}}{27.18}$
        \\
        
        \textbf{CPGNN} & \checkmark &         && $\underset{\scriptscriptstyle{\pm 6.14}}{47.22}$ &  $\underset{\scriptscriptstyle{\pm 8.28}}{81.67}$& &    $\underset{\scriptscriptstyle{\pm 9.65}}{40.56}$ &  $\underset{\scriptscriptstyle{\pm 1.36}}{73.33}$& &  \cellcolor{gray!20}$\underset{\scriptscriptstyle{\pm 10.30}}{49.44}$ &  $\underset{\scriptscriptstyle{\pm 4.16}}{66.11}$& &                    $\underset{\scriptscriptstyle{\pm 2.72}}{21.67}$ &                   $\underset{\scriptscriptstyle{\pm 5.50}}{28.89}$ 
        
        && \cellcolor{blue!20}$\underset{\scriptscriptstyle{\pm 1.23}}{74.55}$ &  $\underset{\scriptscriptstyle{\pm 0.51}}{80.67}$& &        \cellcolor{gray!20}$\underset{\scriptscriptstyle{\pm 1.93}}{68.07}$ &      $\underset{\scriptscriptstyle{\pm 0.62}}{74.92}$& &     \cellcolor{blue!20}$\underset{\scriptscriptstyle{\pm 1.50}}{61.58}$ &   $\underset{\scriptscriptstyle{\pm 7.09}}{60.17}$& &                        $\underset{\scriptscriptstyle{\pm 0.41}}{26.76}$ &                       $\underset{\scriptscriptstyle{\pm 0.63}}{27.13}$
        
        \\
        
        \textbf{GPR-GNN} & \checkmark & & & $\underset{\scriptscriptstyle{\pm 2.72}}{21.67}$ &  $\underset{\scriptscriptstyle{\pm 7.49}}{82.22}$& &    $\underset{\scriptscriptstyle{\pm 2.08}}{24.44}$ &  $\underset{\scriptscriptstyle{\pm 2.08}}{67.78}$& &    $\underset{\scriptscriptstyle{\pm 0.79}}{2.78}$ &  $\underset{\scriptscriptstyle{\pm 4.91}}{56.67}$& &                     $\underset{\scriptscriptstyle{\pm 2.08}}{4.44}$ &                   $\underset{\scriptscriptstyle{\pm 3.42}}{27.78}$ 
        
        && $\underset{\scriptscriptstyle{\pm 5.23}}{48.29}$ &  $\underset{\scriptscriptstyle{\pm 1.75}}{81.84}$& &        $\underset{\scriptscriptstyle{\pm 2.77}}{35.25}$ &      $\underset{\scriptscriptstyle{\pm 0.46}}{70.71}$& &     $\underset{\scriptscriptstyle{\pm 0.60}}{59.94}$ &   $\underset{\scriptscriptstyle{\pm 0.83}}{62.40}$& &                        $\underset{\scriptscriptstyle{\pm 1.29}}{21.06}$ &                       $\underset{\scriptscriptstyle{\pm 0.31}}{26.08}$
        
        \\
        
        \textbf{FAGCN} & \checkmark & & &
        $\underset{\scriptscriptstyle{\pm 6.14}}{26.11}$ &  $\underset{\scriptscriptstyle{\pm 8.16}}{83.33}$& &    $\underset{\scriptscriptstyle{\pm 6.43}}{25.56}$ &  $\underset{\scriptscriptstyle{\pm 5.15}}{70.56}$& &    $\underset{\scriptscriptstyle{\pm 2.83}}{6.11}$ &  $\underset{\scriptscriptstyle{\pm 5.93}}{58.33}$& &                     $\underset{\scriptscriptstyle{\pm 3.60}}{8.33}$ &                   $\underset{\scriptscriptstyle{\pm 0.79}}{29.44}$ 
        
        && $\underset{\scriptscriptstyle{\pm 4.82}}{60.11}$ &  $\underset{\scriptscriptstyle{\pm 0.82}}{81.59}$& &        $\underset{\scriptscriptstyle{\pm 6.00}}{53.18}$ &      $\underset{\scriptscriptstyle{\pm 0.63}}{73.99}$& &     $\underset{\scriptscriptstyle{\pm 1.81}}{55.97}$ &   $\underset{\scriptscriptstyle{\pm 1.38}}{59.64}$& &                        $\underset{\scriptscriptstyle{\pm 0.62}}{24.04}$ &                       $\underset{\scriptscriptstyle{\pm 0.23}}{27.15}$
        
        \\
        
        \textbf{APPNP} & \checkmark & & & \cellcolor{gray!20}$\underset{\scriptscriptstyle{\pm 3.60}}{58.33}$ &  $\underset{\scriptscriptstyle{\pm 5.50}}{72.22}$& &    \cellcolor{gray!20}$\underset{\scriptscriptstyle{\pm 3.14}}{56.11}$ &  $\underset{\scriptscriptstyle{\pm 4.71}}{68.33}$& &   $\underset{\scriptscriptstyle{\pm 2.36}}{36.67}$ &  $\underset{\scriptscriptstyle{\pm 3.93}}{58.89}$& &                    \cellcolor{gray!20}$\underset{\scriptscriptstyle{\pm 1.36}}{25.00}$ &                   $\underset{\scriptscriptstyle{\pm 2.36}}{28.33}$ 
        
        && $\underset{\scriptscriptstyle{\pm 0.91}}{62.56}$ &  $\underset{\scriptscriptstyle{\pm 0.38}}{72.87}$& &        $\underset{\scriptscriptstyle{\pm 1.73}}{49.70}$ &      $\underset{\scriptscriptstyle{\pm 0.23}}{69.59}$& &     $\underset{\scriptscriptstyle{\pm 0.35}}{57.81}$ &   $\underset{\scriptscriptstyle{\pm 0.59}}{57.89}$& &                        \cellcolor{blue!20}$\underset{\scriptscriptstyle{\pm 0.29}}{27.76}$ &                       $\underset{\scriptscriptstyle{\pm 0.11}}{27.41}$
        
        \\
        
        \noalign{\vskip 0.25ex}
        \cdashline{1-3}[0.8pt/2pt]
        \cdashline{5-15}[0.8pt/2pt]
        \cdashline{17-27}[0.8pt/2pt]
        \noalign{\vskip 0.25ex}
        
        \textbf{GNNGuard} & &\checkmark        & & \cellcolor{gray!20}$\underset{\scriptscriptstyle{\pm 1.36}}{58.33}$ &  $\underset{\scriptscriptstyle{\pm 6.29}}{77.22}$& &    \cellcolor{gray!20}$\underset{\scriptscriptstyle{\pm 3.14}}{59.44}$ &  $\underset{\scriptscriptstyle{\pm 4.78}}{67.78}$& &    $\underset{\scriptscriptstyle{\pm 0.79}}{0.56}$ &  $\underset{\scriptscriptstyle{\pm 2.08}}{67.22}$& &                     $\underset{\scriptscriptstyle{\pm 1.57}}{9.44}$ &                   $\underset{\scriptscriptstyle{\pm 3.60}}{28.33}$ 
        
        &&\cellcolor{gray!20}$\underset{\scriptscriptstyle{\pm 0.55}}{74.20}$ &  $\underset{\scriptscriptstyle{\pm 0.55}}{80.15}$& &        \cellcolor{gray!20}$\underset{\scriptscriptstyle{\pm 0.74}}{68.13}$ &      $\underset{\scriptscriptstyle{\pm 0.28}}{72.61}$& &     \cellcolor{gray!20}$\underset{\scriptscriptstyle{\pm 0.48}}{60.89}$ &   $\underset{\scriptscriptstyle{\pm 0.60}}{65.66}$& &                        $\underset{\scriptscriptstyle{\pm 0.67}}{23.78}$ &                       $\underset{\scriptscriptstyle{\pm 0.98}}{26.51}$ 
        \\
        
        \textbf{ProGNN} & &\checkmark          &&  $\underset{\scriptscriptstyle{\pm 7.97}}{48.89}$ &  $\underset{\scriptscriptstyle{\pm 3.42}}{79.44}$& &    $\underset{\scriptscriptstyle{\pm 7.49}}{32.78}$ &  $\underset{\scriptscriptstyle{\pm 4.78}}{67.22}$& &   $\underset{\scriptscriptstyle{\pm 4.78}}{33.89}$ &  $\underset{\scriptscriptstyle{\pm 3.93}}{51.11}$& &                    $\underset{\scriptscriptstyle{\pm 9.26}}{17.78}$ &                   $\underset{\scriptscriptstyle{\pm 5.50}}{27.22}$ 
        &&
        
        $\underset{\scriptscriptstyle{\pm 6.20}}{45.10}$ &  $\underset{\scriptscriptstyle{\pm 0.43}}{81.32}$& &        $\underset{\scriptscriptstyle{\pm 1.02}}{46.58}$ &      $\underset{\scriptscriptstyle{\pm 1.12}}{71.82}$& &     $\underset{\scriptscriptstyle{\pm 1.19}}{53.40}$ &   $\underset{\scriptscriptstyle{\pm 0.03}}{49.84}$& &                        $\underset{\scriptscriptstyle{\pm 1.09}}{24.80}$ &                       $\underset{\scriptscriptstyle{\pm 0.66}}{27.49}$
        
        \\
        
        \textbf{GCN-SVD} & &\checkmark         & & $\underset{\scriptscriptstyle{\pm 4.91}}{53.33}$ &  $\underset{\scriptscriptstyle{\pm 4.16}}{75.56}$& &    $\underset{\scriptscriptstyle{\pm 2.08}}{28.89}$ &  $\underset{\scriptscriptstyle{\pm 0.79}}{59.44}$& &   \cellcolor{gray!20}$\underset{\scriptscriptstyle{\pm 2.36}}{41.67}$ &  $\underset{\scriptscriptstyle{\pm 4.37}}{50.56}$& &                    \cellcolor{gray!20}$\underset{\scriptscriptstyle{\pm 5.44}}{25.00}$ &                   $\underset{\scriptscriptstyle{\pm 6.71}}{27.78}$ &&
        
        $\underset{\scriptscriptstyle{\pm 7.59}}{47.82}$ &  $\underset{\scriptscriptstyle{\pm 0.31}}{76.61}$& &        $\underset{\scriptscriptstyle{\pm 1.78}}{51.20}$ &      $\underset{\scriptscriptstyle{\pm 0.16}}{66.90}$& &     $\underset{\scriptscriptstyle{\pm 2.06}}{55.00}$ &   $\underset{\scriptscriptstyle{\pm 0.23}}{55.47}$& &                        \cellcolor{gray!20}$\underset{\scriptscriptstyle{\pm 0.91}}{25.25}$ &                       $\underset{\scriptscriptstyle{\pm 0.25}}{26.63}$
        \\
        
        \textbf{GCN-SMGDC} & & \checkmark     && $\underset{\scriptscriptstyle{\pm 4.91}}{40.00}$ &  $\underset{\scriptscriptstyle{\pm 3.93}}{77.78}$& &    $\underset{\scriptscriptstyle{\pm 2.83}}{33.89}$ &  $\underset{\scriptscriptstyle{\pm 0.79}}{62.22}$& &   $\underset{\scriptscriptstyle{\pm 4.08}}{16.67}$ &  $\underset{\scriptscriptstyle{\pm 5.67}}{51.11}$& &                    $\underset{\scriptscriptstyle{\pm 5.15}}{20.56}$ &                   $\underset{\scriptscriptstyle{\pm 2.36}}{28.33}$      &&
        
        $\underset{\scriptscriptstyle{\pm 1.18}}{29.66}$ &  $\underset{\scriptscriptstyle{\pm 0.52}}{77.26}$& &        $\underset{\scriptscriptstyle{\pm 2.36}}{55.04}$ &      $\underset{\scriptscriptstyle{\pm 0.59}}{72.33}$& &     $\underset{\scriptscriptstyle{\pm 1.19}}{50.76}$ &   $\underset{\scriptscriptstyle{\pm 0.30}}{51.99}$& &                        $\underset{\scriptscriptstyle{\pm 1.21}}{24.71}$ &                       $\underset{\scriptscriptstyle{\pm 0.60}}{26.06}$
        
        \\

        \noalign{\vskip 0.25ex}
        \cdashline{1-3}[0.8pt/2pt]
        \cdashline{5-15}[0.8pt/2pt]
        \cdashline{17-27}[0.8pt/2pt]
        \noalign{\vskip 0.25ex}
        
        \textbf{GAT} & &             & & $\underset{\scriptscriptstyle{\pm 0.79}}{13.89}$ &  $\underset{\scriptscriptstyle{\pm 3.42}}{84.44}$& &     $\underset{\scriptscriptstyle{\pm 3.42}}{8.89}$ &  $\underset{\scriptscriptstyle{\pm 7.20}}{70.00}$& &    \cellcolor{gray!20}$\underset{\scriptscriptstyle{\pm 0.79}}{0.56}$ &  $\underset{\scriptscriptstyle{\pm 0.79}}{60.56}$& &                     \cellcolor{gray!20}$\underset{\scriptscriptstyle{\pm 4.37}}{3.89}$ &                   $\underset{\scriptscriptstyle{\pm 2.83}}{30.56}$ 
        &&
        \cellcolor{gray!20}$\underset{\scriptscriptstyle{\pm 3.60}}{41.70}$ &  $\underset{\scriptscriptstyle{\pm 0.24}}{83.72}$& &        $\underset{\scriptscriptstyle{\pm 2.17}}{48.40}$ &      $\underset{\scriptscriptstyle{\pm 1.00}}{73.40}$& &     $\underset{\scriptscriptstyle{\pm 0.66}}{50.37}$ &   $\underset{\scriptscriptstyle{\pm 0.92}}{61.69}$& &                        \cellcolor{gray!20}$\underset{\scriptscriptstyle{\pm 0.73}}{25.00}$ &                       $\underset{\scriptscriptstyle{\pm 0.03}}{27.30}$
        
        \\
        
        \textbf{GCN} & &           &  & \cellcolor{gray!20}$\underset{\scriptscriptstyle{\pm 3.60}}{18.33}$ &  $\underset{\scriptscriptstyle{\pm 5.50}}{82.78}$& &    \cellcolor{gray!20}$\underset{\scriptscriptstyle{\pm 5.50}}{20.56}$ &  $\underset{\scriptscriptstyle{\pm 8.20}}{72.78}$& &    $\underset{\scriptscriptstyle{\pm 0.00}}{0.00}$ &  $\underset{\scriptscriptstyle{\pm 7.97}}{56.11}$& &                     $\underset{\scriptscriptstyle{\pm 3.14}}{2.22}$ &                   $\underset{\scriptscriptstyle{\pm 2.08}}{30.56}$ 
        &&
        $\underset{\scriptscriptstyle{\pm 4.83}}{31.98}$ &  $\underset{\scriptscriptstyle{\pm 0.96}}{83.12}$& &        \cellcolor{gray!20}$\underset{\scriptscriptstyle{\pm 2.52}}{49.43}$ &      $\underset{\scriptscriptstyle{\pm 1.05}}{75.30}$& &     \cellcolor{gray!20}$\underset{\scriptscriptstyle{\pm 0.25}}{52.62}$ &   $\underset{\scriptscriptstyle{\pm 0.13}}{54.20}$& &                        $\underset{\scriptscriptstyle{\pm 0.63}}{24.36}$ &                       $\underset{\scriptscriptstyle{\pm 0.13}}{26.68}$
        
        \\
    
        \noalign{\vskip 0.25ex}
        \cdashline{1-3}[0.8pt/2pt]
        \cdashline{5-15}[0.8pt/2pt]
        \cdashline{17-27}[0.8pt/2pt]
        \noalign{\vskip 0.25ex}
        
        \textbf{MLP}* & &             &
       &  $\underset{\scriptscriptstyle{\pm 3.42}}{64.44}$ &  $\underset{\scriptscriptstyle{\pm 3.42}}{64.44}$& &    $\underset{\scriptscriptstyle{\pm 3.42}}{\textbf{70.56}}$ &  $\underset{\scriptscriptstyle{\pm 3.42}}{70.56}$& &   $\underset{\scriptscriptstyle{\pm 2.83}}{57.78}$ &  $\underset{\scriptscriptstyle{\pm 2.83}}{57.78}$& &                    $\underset{\scriptscriptstyle{\pm 2.72}}{\textbf{30.00}}$ &                   $\underset{\scriptscriptstyle{\pm 2.72}}{30.00}$ 
        
        & %
        
        &$\underset{\scriptscriptstyle{\pm 1.58}}{64.55}$ &  $\underset{\scriptscriptstyle{\pm 1.58}}{64.55}$& &        $\underset{\scriptscriptstyle{\pm 0.11}}{67.67}$ &      $\underset{\scriptscriptstyle{\pm 0.11}}{67.67}$& &     $\underset{\scriptscriptstyle{\pm 0.58}}{56.56}$ &   $\underset{\scriptscriptstyle{\pm 0.58}}{56.56}$& &                        $\underset{\scriptscriptstyle{\pm 1.05}}{26.25}$ &                       $\underset{\scriptscriptstyle{\pm 1.05}}{26.25}$\\
        \bottomrule 
        \end{tabular}}
    \vspace{-0.45cm}
\end{table*}

\paragraph{Targeted attacks by \nettack}
\textcircled{\raisebox{-0.9pt}{1}}~\textit{Poison attacks}. 
Under targeted poison attacks, Table~\ref{table:real-results-poison-only} (left) shows that GraphSAGE-SVD and H$_2$GCN-SVD, which combine our identified design with a low-rank vaccination approach adopted in GCN-SVD~\citep{entezari2020all},
outperform state-of-the-art vaccinated methods across all datasets by up to 
13.34\% in homophilous settings %
and 18.33\% in heterophilous settings. %
Furthermore, GraphSAGE-SMGDC and H$_2$GCN-SMGDC, which combine our design with existing vaccinations based on Soft Medoid~\cite{geisler2020reliable} and GDC~\cite{klicpera_diffusion_2019}, show better performance against attacks in all datasets compared to GCN-SMGDC, the corresponding baseline without our design, with 
up to 19.44\% improvement on homophilous settings %
and 30.55\% improvement on heterophilous settings.
In summary, these observations show that the heterophilous design improves the robustness of GNNs alongside existing vaccination mechanisms.

Methods merely employing the identified design also show significantly improved robustness, though there are differences in the amount of robustness improvement due to architectural differences. 
Specifically, these methods outperform the best unvaccinated method (GAT) on all datasets by up to 33.75\% in average, 
despite having mostly comparable performance on clean datasets; 
methods like APPNP and CPGNN also show comparable or even better robustness than state-of-the-art vaccinated GNNs. {These observations also apply to the larger Pubmed dataset in Table~\ref{table:results-pubmed}}.
{We also note that the graph-agnostic MLP, which is immune to structural attacks, outperforms all GNNs against attacks on Citeseer and Snap; this
shows the challenges in defending against targeted attacks and 
calls for more effective defense strategies upon our discoveries.}

\textcircled{\raisebox{-0.9pt}{2}}~\textit{Evasion attacks}. 
{Under evasion attacks 
(detailed results are reported in App. Table~\ref{table:real-results-detailed-netattack} on our \repo{}),
we observe \textit{similar trends as in poison attacks}: 
GraphSAGE-SVD and H$_2$GCN-SVD are up to 20.55\% more accurate than the GCN-SVD, the corresponding baseline without the heterophilous design,
and GraphSAGE-SMGDC and H$_2$GCN-SMGDC outperform GCN-SMGDC by up to 19.44\%. }
Methods featuring the identified design alone achieve up to 38.89\% gain in average performance against the best unvaccinated baseline, {which we also observe on Pubmed.}
We note that two baselines, GNNGuard and ProGNN, are designed specifically to defend against poison attacks, and are not capable of addressing evasion attacks.

\paragraph{Untargeted attacks by Metattack}
\textcircled{\raisebox{-0.9pt}{1}}~\textit{Poison attacks}.
We also test the robustness of each method against untargeted attacks. 
Table~\ref{table:real-results-poison-only} (right) shows the performance under poison attacks.  
Though our theoretical analysis in \S\ref{sec:theories} focuses on the effect of the heterophilous design under targeted attacks, we observe similar improvements in robustness against untargeted attacks in the poison setup. GNNs with the identified design show mostly improved robustness compared to unvaccinated models, while having similar performance on the clean datasets. 
Specifically, CPGNN shows exceptional robustness, outperforming the best unvaccinated model by up to 32.85\%. %
{
Moreover, models combining the identified design with low-rank approximation show up to 21.04\% improvement in accuracy compared to GCN-SVD, which uses only low-rank approximation. 
Models combining the design with Soft Medoid and GDC show up to 37.29\% improvement in accuracy compared to GCN-SMGDC. 
We also note that the most robust method for each dataset is among the ones with the identified design. 
These results again support the effectiveness of the heterophilous design in boosting the robustness of GNNs in addition to existing vaccination mechanisms.
}

\begin{table*}[t]
    \centering
    \caption{Accumulated certifications (AC), average certifiable radii ($\bar{r}_a$ and $\bar{r}_d$) and accuracy of GNNs with randomized smoothing enabled (i.e., $f(\phi(\mathbf{s}))$) on the test splits of the clean datasets, with ramdomization schemes $\phi$ allowing both addition and deletion (i.e., $p_+ = 0.001, p_- = 0.4$), and additional only (i.e., $p_+ = 0.001, p_- = 0$). For each statistic, we report the mean and stdev across 3 runs. Best results highlighted in blue per dataset, and in gray per model group. 
    We provide results with the deletion only scheme on our \repo{} (App. \S\ref{app:real-cert-results}).
    APPNP, with the identified design, improves the accumulated certification (AC) by up to 5.3x on homophilous datasets and 10.1x on heterophilous ones compared to the best performing baseline without the design.
    }
    \vspace{-0.3cm}
    \label{table:real-results-cert-both}
    \resizebox{\linewidth}{!}{
    \begin{tabular}{l lH c rrr>{\color{dark-gray}}r c rrH>{\color{dark-gray}}r c rrr>{\color{dark-gray}}r c rrH>{\color{dark-gray}}r}
        \toprule
        & \multirow{2}{*}{\rotatebox[origin=r]{90}{\textbf{Hete.}}} & \multirow{3}{*}{\rotatebox[origin=r]{90}{\textbf{Vaccin.}}} &
        
        & \multicolumn{4}{c}{\bf Addition \& Deletion} & 
        & \multicolumn{4}{c}{\bf Addition Only } &
        & \multicolumn{4}{c}{\bf Addition \& Deletion} &
        & \multicolumn{4}{c}{\bf Addition Only} 
        \\
        \cmidrule{5-8} \cmidrule{10-13} \cmidrule{15-18} \cmidrule{20-23}
        
        & & & & 
        \multicolumn{1}{c}{AC}  & \multicolumn{1}{c}{$\bar{r}_a$} & \multicolumn{1}{c}{$\bar{r}_d$} & \multicolumn{1}{c}{Acc. \%} & & 
        \multicolumn{1}{c}{AC}  & \multicolumn{1}{c}{$\bar{r}_a$} &  & \multicolumn{1}{c}{Acc. \%} & & 
        \multicolumn{1}{c}{AC}  & \multicolumn{1}{c}{$\bar{r}_a$} & \multicolumn{1}{c}{$\bar{r}_d$} & \multicolumn{1}{c}{Acc. \%} & & 
        \multicolumn{1}{c}{AC}  & \multicolumn{1}{c}{$\bar{r}_a$} &  & \multicolumn{1}{c}{Acc. \%} 
        \\

        \cmidrule{1-3} \cmidrule{5-8} \cmidrule{10-13} \cmidrule{15-18} \cmidrule{20-23}
        
        \textbf{H$_2$GCN} & \checkmark & &
        \multirow{9}{*}{\rotatebox[origin=c]{90}{\textbf{Cora}}}
        & 3.96\tiny{$\pm$0.33} & 0.46\tiny{$\pm$0.08} & 3.90\tiny{$\pm$0.30} & 79.34\tiny{$\pm$1.93} & 
        & 0.42\tiny{$\pm$0.02} & 0.53\tiny{$\pm$0.03} & - & 80.97\tiny{$\pm$1.95} & 
        \multirow{9}{*}{\rotatebox[origin=c]{90}{\textbf{Citeseer}}}
        & 2.96\tiny{$\pm$0.88} & 0.33\tiny{$\pm$0.13} & 3.27\tiny{$\pm$0.67} & 71.76\tiny{$\pm$4.05} & 
        & 0.29\tiny{$\pm$0.05} & 0.40\tiny{$\pm$0.06} & - & 72.99\tiny{$\pm$2.22} \\

        \textbf{GraphSAGE} & \checkmark & &
        & 2.16\tiny{$\pm$0.06} & 0.13\tiny{$\pm$0.00} & 2.43\tiny{$\pm$0.03} & 79.61\tiny{$\pm$1.48} & 
        & 0.28\tiny{$\pm$0.03} & 0.34\tiny{$\pm$0.04} & - & 81.07\tiny{$\pm$1.11} & 
        & 2.21\tiny{$\pm$0.15} & 0.19\tiny{$\pm$0.01} & 2.56\tiny{$\pm$0.09} & 73.48\tiny{$\pm$2.90} & 
        & 0.33\tiny{$\pm$0.01} & 0.44\tiny{$\pm$0.01} & - & 74.70\tiny{$\pm$1.37} \\

        \textbf{CPGNN} & \checkmark & &
        & 1.87\tiny{$\pm$0.27} & 0.14\tiny{$\pm$0.05} & 2.24\tiny{$\pm$0.30} & 75.37\tiny{$\pm$1.65} & 
        & 0.17\tiny{$\pm$0.02} & 0.21\tiny{$\pm$0.03} & - & 78.34\tiny{$\pm$1.26} & 
        & 2.03\tiny{$\pm$0.17} & 0.11\tiny{$\pm$0.01} & 2.52\tiny{$\pm$0.20} & 73.48\tiny{$\pm$0.61} & 
        & 0.15\tiny{$\pm$0.02} & 0.20\tiny{$\pm$0.02} & - & 74.62\tiny{$\pm$0.30} \\

        \textbf{GPR-GNN} & \checkmark & &
        & 4.42\tiny{$\pm$0.43} & 0.63\tiny{$\pm$0.06} & 4.35\tiny{$\pm$0.22} & 74.90\tiny{$\pm$2.34} & 
        & 0.43\tiny{$\pm$0.03} & 0.55\tiny{$\pm$0.03} & - & 76.96\tiny{$\pm$2.18} & 
        & 4.63\tiny{$\pm$0.27} & 0.81\tiny{$\pm$0.07} & 4.92\tiny{$\pm$0.24} & 66.33\tiny{$\pm$0.20} & 
        & 0.40\tiny{$\pm$0.01} & 0.59\tiny{$\pm$0.02} & - & 67.52\tiny{$\pm$0.49} \\

        \textbf{FAGCN} & \checkmark & &
        & 4.30\tiny{$\pm$0.07} & 0.57\tiny{$\pm$0.02} & 4.25\tiny{$\pm$0.04} & 76.49\tiny{$\pm$1.73} & 
        & 0.43\tiny{$\pm$0.01} & 0.54\tiny{$\pm$0.01} & - & 79.04\tiny{$\pm$0.68} & 
        & 4.07\tiny{$\pm$0.15} & 0.58\tiny{$\pm$0.02} & 4.23\tiny{$\pm$0.09} & 71.82\tiny{$\pm$0.73} & 
        & 0.38\tiny{$\pm$0.02} & 0.53\tiny{$\pm$0.02} & - & 72.41\tiny{$\pm$1.03} \\

        \textbf{APPNP} & \checkmark & &
        & 10.11\tiny{$\pm$0.04}\cellcolor{blue!20} & 1.86\tiny{$\pm$0.01}\cellcolor{blue!20} & 8.52\tiny{$\pm$0.06}\cellcolor{blue!20} & 71.97\tiny{$\pm$0.25} & 
        & 0.69\tiny{$\pm$0.00}\cellcolor{blue!20} & 0.95\tiny{$\pm$0.00}\cellcolor{blue!20} & - & 72.27\tiny{$\pm$0.31} & 
        & 9.87\tiny{$\pm$0.02}\cellcolor{blue!20} & 1.88\tiny{$\pm$0.00}\cellcolor{blue!20} & 8.61\tiny{$\pm$0.01}\cellcolor{blue!20} & 69.39\tiny{$\pm$0.23} & 
        & 0.66\tiny{$\pm$0.00}\cellcolor{blue!20} & 0.95\tiny{$\pm$0.00}\cellcolor{blue!20} & - & 69.41\tiny{$\pm$0.22} \\

        \noalign{\vskip 0.25ex}
        \cdashline{1-3}[0.8pt/2pt]
        \cdashline{5-8}[0.8pt/2pt]
        \cdashline{10-13}[0.8pt/2pt]
        \cdashline{15-18}[0.8pt/2pt]
        \cdashline{20-23}[0.8pt/2pt]
        \noalign{\vskip 0.25ex}

        \textbf{GAT} & & &
        & \cellcolor{gray!20}1.61\tiny{$\pm$0.10} & \cellcolor{gray!20}0.08\tiny{$\pm$0.01} & \cellcolor{gray!20}1.85\tiny{$\pm$0.06} & 79.83\tiny{$\pm$2.36} & 
        & \cellcolor{gray!20}0.19\tiny{$\pm$0.04} & \cellcolor{gray!20}0.23\tiny{$\pm$0.04} & - & 81.99\tiny{$\pm$1.94} & 
        & 1.29\tiny{$\pm$0.07} & 0.07\tiny{$\pm$0.02} & 1.60\tiny{$\pm$0.06} & 73.62\tiny{$\pm$1.06} & 
        & 0.09\tiny{$\pm$0.01} & 0.12\tiny{$\pm$0.02} & - & 74.47\tiny{$\pm$0.26} \\

        \textbf{GCN} & & &
        & 1.40\tiny{$\pm$0.02} & 0.06\tiny{$\pm$0.01} & 1.75\tiny{$\pm$0.08} & 74.36\tiny{$\pm$3.46} & 
        & 0.13\tiny{$\pm$0.00} & 0.17\tiny{$\pm$0.01} & - & 78.17\tiny{$\pm$2.89} & 
        & \cellcolor{gray!20}1.79\tiny{$\pm$0.04} & \cellcolor{gray!20}0.17\tiny{$\pm$0.02} & \cellcolor{gray!20}2.15\tiny{$\pm$0.11} & 70.38\tiny{$\pm$4.17} & 
        & \cellcolor{gray!20}0.17\tiny{$\pm$0.01} & \cellcolor{gray!20}0.24\tiny{$\pm$0.02} & - & 72.04\tiny{$\pm$3.64} \\

        \cmidrule{1-3} \cmidrule{5-8} \cmidrule{10-13} \cmidrule{15-18} \cmidrule{20-23}
        
        \textbf{H$_2$GCN} & \checkmark & &
        \multirow{9}{*}{\rotatebox[origin=c]{90}{\textbf{FB100}}}
         & 8.12\tiny{$\pm$0.10} & 1.76\tiny{$\pm$0.02} & 8.14\tiny{$\pm$0.06} & 57.38\tiny{$\pm$0.17} & 
         & 0.54\tiny{$\pm$0.00} & 0.94\tiny{$\pm$0.00} & - & 57.11\tiny{$\pm$0.10} & 
        \multirow{9}{*}{\rotatebox[origin=c]{90}{\textbf{Snap}}}
         & 1.44\tiny{$\pm$0.18} & 0.59\tiny{$\pm$0.10} & 3.79\tiny{$\pm$0.40} & 26.97\tiny{$\pm$0.10} & 
         & 0.11\tiny{$\pm$0.01} & 0.42\tiny{$\pm$0.05} & - & 26.74\tiny{$\pm$0.18} \\
        
        \textbf{GraphSAGE} & \checkmark & &
         & 6.98\tiny{$\pm$0.06} & 1.50\tiny{$\pm$0.04} & 7.32\tiny{$\pm$0.13} & 56.72\tiny{$\pm$1.56} & 
         & 0.52\tiny{$\pm$0.01} & 0.92\tiny{$\pm$0.01} & - & 56.70\tiny{$\pm$1.41} & 
         & 0.70\tiny{$\pm$0.21} & 0.19\tiny{$\pm$0.11} & 2.16\tiny{$\pm$0.54} & 26.84\tiny{$\pm$0.47} & 
         & 0.06\tiny{$\pm$0.02} & 0.24\tiny{$\pm$0.08} & - & 27.00\tiny{$\pm$0.63} \\
        
        \textbf{CPGNN} & \checkmark & &
         & 6.80\tiny{$\pm$0.19} & 1.41\tiny{$\pm$0.21} & 7.05\tiny{$\pm$0.70} & 59.00\tiny{$\pm$5.71} & 
         & 0.54\tiny{$\pm$0.04} & 0.90\tiny{$\pm$0.04} & - & 60.39\tiny{$\pm$7.26} & 
         & 1.45\tiny{$\pm$0.23} & 0.61\tiny{$\pm$0.14} & 3.89\tiny{$\pm$0.51} & 26.71\tiny{$\pm$0.25} & 
         & 0.12\tiny{$\pm$0.02} & 0.43\tiny{$\pm$0.08} & - & 27.00\tiny{$\pm$0.41} \\
        
        \textbf{GPR-GNN} & \checkmark & &
         & 5.81\tiny{$\pm$0.16} & 1.11\tiny{$\pm$0.02} & 5.95\tiny{$\pm$0.10} & 61.99\tiny{$\pm$0.44} & 
         & 0.46\tiny{$\pm$0.01} & 0.73\tiny{$\pm$0.02} & - & 62.26\tiny{$\pm$0.26} & 
         & 0.52\tiny{$\pm$0.06} & 0.11\tiny{$\pm$0.01} & 1.70\tiny{$\pm$0.14} & 26.31\tiny{$\pm$1.03} & 
         & 0.03\tiny{$\pm$0.01} & 0.11\tiny{$\pm$0.02} & - & 26.14\tiny{$\pm$0.73} \\
        
        \textbf{FAGCN} & \checkmark & &
         & 7.45\tiny{$\pm$0.21} & 1.53\tiny{$\pm$0.02} & 7.40\tiny{$\pm$0.06} & 59.76\tiny{$\pm$1.47} & 
         & 0.55\tiny{$\pm$0.00} & 0.90\tiny{$\pm$0.01} & - & 60.60\tiny{$\pm$0.36} & 
         & 1.41\tiny{$\pm$0.10} & 0.56\tiny{$\pm$0.06} & 3.81\tiny{$\pm$0.22} & 27.07\tiny{$\pm$0.16} & 
         & 0.10\tiny{$\pm$0.01} & 0.36\tiny{$\pm$0.03} & - & 27.13\tiny{$\pm$0.16} \\
        
        \textbf{APPNP} & \checkmark & &
         & 8.90\tiny{$\pm$0.03}\cellcolor{blue!20} & 1.92\tiny{$\pm$0.02}\cellcolor{blue!20} & 8.73\tiny{$\pm$0.05}\cellcolor{blue!20} & 57.87\tiny{$\pm$0.57} & 
         & 0.57\tiny{$\pm$0.00}\cellcolor{blue!20} & 0.98\tiny{$\pm$0.01}\cellcolor{blue!20} & - & 57.89\tiny{$\pm$0.59} & 
         & 3.54\tiny{$\pm$0.03}\cellcolor{blue!20} & 1.68\tiny{$\pm$0.01}\cellcolor{blue!20} & 7.95\tiny{$\pm$0.04}\cellcolor{blue!20} & 27.45\tiny{$\pm$0.14} & 
         & 0.24\tiny{$\pm$0.00}\cellcolor{blue!20} & 0.86\tiny{$\pm$0.00}\cellcolor{blue!20} & - & 27.46\tiny{$\pm$0.17} \\
        
        \noalign{\vskip 0.25ex}
        \cdashline{1-3}[0.8pt/2pt]
        \cdashline{5-8}[0.8pt/2pt]
        \cdashline{10-13}[0.8pt/2pt]
        \cdashline{15-18}[0.8pt/2pt]
        \cdashline{20-23}[0.8pt/2pt]
        \noalign{\vskip 0.25ex}
        
        \textbf{GAT} & & &
         & 4.30\tiny{$\pm$0.26} & 0.77\tiny{$\pm$0.04} & 4.72\tiny{$\pm$0.19} & 61.56\tiny{$\pm$0.78} & 
         & \cellcolor{gray!20}0.46\tiny{$\pm$0.03} & 0.74\tiny{$\pm$0.04} & - & 61.97\tiny{$\pm$1.41} & 
         & 0.28\tiny{$\pm$0.09} & 0.04\tiny{$\pm$0.01} & 0.95\tiny{$\pm$0.33} & 27.12\tiny{$\pm$0.52} & 
         & \cellcolor{gray!20}0.02\tiny{$\pm$0.00} & \cellcolor{gray!20}0.08\tiny{$\pm$0.02} & - & 27.00\tiny{$\pm$0.59} \\
        
        \textbf{GCN} & & &
         & \cellcolor{gray!20}5.19\tiny{$\pm$0.03} & \cellcolor{gray!20}1.14\tiny{$\pm$0.00} & \cellcolor{gray!20}6.05\tiny{$\pm$0.01} & 54.16\tiny{$\pm$0.08} & 
         & 0.43\tiny{$\pm$0.00} & \cellcolor{gray!20}0.79\tiny{$\pm$0.01} & - & 54.39\tiny{$\pm$0.14} & 
         & \cellcolor{gray!20}0.32\tiny{$\pm$0.08} & \cellcolor{gray!20}0.06\tiny{$\pm$0.03} & \cellcolor{gray!20}1.08\tiny{$\pm$0.24} & 26.17\tiny{$\pm$0.34} & 
         & \cellcolor{gray!20}0.02\tiny{$\pm$0.01} & \cellcolor{gray!20}0.08\tiny{$\pm$0.03} & - & 26.38\tiny{$\pm$0.49} \\        
        
    \bottomrule
    \end{tabular}
    }
    \vspace{-0.4cm}
\end{table*}

\textcircled{\raisebox{-0.9pt}{2}}~\textit{Evasion attacks}.
We present the performance under evasion attacks 
on our \githubrepo. 
\textit{Unlike the poison attacks}, the evasion setup only leads to a slight decrease in average accuracy of less than 2\% for most models. 
Moreover, there appears to be no clearly increased robustness for vaccinated models (with the identified design or other vaccination machanisms) compared to unvaccinated models. 
This can be attributed to the reduced effectiveness of evasion vs.\ poison attacks (as in \nettack), and the increased challenges of untargeted attacks.

\begin{table}[t]
    \centering
    \caption{
        Additional results on Pubmed (details in App. \S\ref{app:real-benchmark-results}).
    }
    \vspace{-0.35cm}
    \label{table:results-pubmed}
    \resizebox*{0.89\linewidth}{!}{%
    \begin{tabular}{l ccc rr>{\color{dark-gray}}r}
        \toprule
        
        & \multirow{-0.9}{*}{\rotatebox[origin=r]{90}{\small{\textbf{Hete.}}}} & \multirow{-1.5}{*}{\rotatebox[origin=r]{90}{\small{\textbf{Vaccin.}}}} &
        
        & \multicolumn{3}{c}{\texttt{\bf Pubmed}} \\
        
        \cmidrule{5-7}
        
        &&& & \multicolumn{1}{c}{Poison} & \multicolumn{1}{c}{Evasion} & \multicolumn{1}{c}{Clean} \\
        
        \cmidrule{1-3} \cmidrule{5-7}

        \textbf{H$_2$GCN-SVD}       & \checkmark & \checkmark && \cellcolor{blue!20}  86.11\tiny{$\pm 3.93$}
        &\cellcolor{blue!20} 86.11\tiny{$\pm 3.93$}&
        87.22\tiny{$\pm 4.37$}\\
        
        \textbf{GraphSAGE-SVD}   & \checkmark &\checkmark & & 81.11\tiny{$\pm 4.16$}& 81.11\tiny{$\pm 3.42$}
        &84.44\tiny{$\pm 2.08$}\\
  
        \noalign{\vskip 0.25ex}
        \cdashline{1-3}[0.8pt/2pt]\cdashline{5-7}[0.8pt/2pt]
        \noalign{\vskip 0.25ex}
        
        \textbf{H$_2$GCN}           & \checkmark & & & 44.44\tiny{$\pm 5.67$} &
        46.67\tiny{$\pm 8.16$}& 87.78\tiny{$\pm 3.14$}\\
        
        \textbf{GraphSAGE}      & \checkmark & & & 33.33\tiny{$\pm 8.92$} &34.44\tiny{$\pm 9.06$} &84.44\tiny{$\pm 3.93$}\\
        
        \textbf{CPGNN}       & \checkmark & & &  60.00\tiny{$\pm 7.20$}&  60.00\tiny{$\pm 5.93$} & 82.78\tiny{$\pm 5.67$}\\

        \textbf{GPR-GNN}         & \checkmark & & & 13.89\tiny{$\pm 4.78$} & 15.56\tiny{$\pm 6.14$}&85.56\tiny{$\pm 1.57$}\\
        
        \textbf{FAGCN}           & \checkmark & & &  27.78\tiny{$\pm 11.00$} &31.67\tiny{$\pm 13.40$}&86.67\tiny{$\pm 2.72$}\\
        
        \textbf{APPNP}           & \checkmark & & & \cellcolor{gray!20} 79.44\tiny{$\pm 2.83$} &\cellcolor{gray!20}81.67\tiny{$\pm 2.72$} & 86.67\tiny{$\pm 2.36$}\\
        
        \noalign{\vskip 0.25ex}
        \cdashline{1-3}[0.8pt/2pt]\cdashline{5-7}[0.8pt/2pt]
        \noalign{\vskip 0.25ex}
        
        \textbf{GNNGuard}        & & \checkmark & & \cellcolor{gray!20} 73.89\tiny{$\pm 6.71$}& \multicolumn{1}{c}{-} &82.78\tiny{$\pm 2.83$}\\

        \noalign{\vskip 0.25ex}
        \cdashline{1-3}[0.8pt/2pt]\cdashline{5-7}[0.8pt/2pt]
        \noalign{\vskip 0.25ex}
        
        \textbf{GAT}           &&&  &  \cellcolor{gray!20}7.22\tiny{$\pm 4.16$}&\cellcolor{gray!20} 6.67\tiny{$\pm 4.08$}&83.33\tiny{$\pm 1.36$}\\
        
        \textbf{GCN}          &&&   &5.56\tiny{$\pm 0.79$} & 5.56\tiny{$\pm 0.79$} & 85.00\tiny{$\pm 2.72$}\\
        
        \noalign{\vskip 0.25ex}
        \cdashline{1-3}[0.8pt/2pt]\cdashline{5-7}[0.8pt/2pt]
        \noalign{\vskip 0.25ex}
        
        \textbf{MLP*}        &&&  \multirow{-14}{*}{\rotatebox[origin=c]{90}{\nettack}}   &  86.11\tiny{$\pm 4.37$}& 86.11\tiny{$\pm 4.37$}&86.11\tiny{$\pm 4.37$} \\
        \bottomrule
    \end{tabular}
    }
    \vspace{-0.5cm}
  \end{table}

\subsection{(Q3) Heterophily-adjusted GNNs are Certifiably More Robust}
\label{sec:exp-cert-robustness}

It is worth noting that robustness against specific attacks such as \nettack{} and Metattack does not guarantee robustness towards other possible attacks. 
To overcome this limitation, \emph{robustness certificates} provide guarantees (in some cases probabilistically) that attacks within a certain radius cannot change a model's predictions. 
Complementary to our evaluation on empirical robustness, we further demonstrate that heterophily-adjusted GNNs featuring our identified design are certifiably more robust than methods without it, thus answering (Q3).
For GNN models and datasets,
we exclude {the larger Pubmed dataset} and models that learn to rewrite the graph structure through the training process, or require recalculation of the low-rank approximation or inverse of matrices (as used by GDC~\cite{klicpera_diffusion_2019}) for every randomized perturbation, as we find that sampling on {these setups} is computationally challenging. 
We use the same hyperparameters as the benchmark study in \S\ref{sec:exp-benchmark-study}.

Table~\ref{table:real-results-cert-both} shows multiple metrics of certifiable robustness of each GNN model under edge randomization schemes allowing both addition and deletion, and allowing addition only; we additionally report results under a scheme allowing only deletion on our \githubrepo. 
For the scheme allowing both addition and deletion, we observe that all heterophily-adjusted methods have better certifiable robustness compared to methods without the design. 
Specifically, on homophilous datasets (Cora and Citeseer), methods with the identified design achieve an up to 5.3 times relative improvement in accumulated certification.
On heterophilous datasets (FB100 and Snap), they outperform the baselines by a factor of 11.1. 
In the more challenging case with the addition only scheme, methods with the design also show up to 2.9 times relative increase in AC on the homophilous datasets 
and 11.0 times relative increase in AC on the heterophilous datasets 
compared to the baselines. 
For the deletion only scheme, we find that unvaccinated models like GCN already have decent certifiable robustness in this scenario, commensurating with our discussions in \S\ref{sec:exp-perturb-observations}
that deletions create less severe perturbations.
Overall, our results show that models featuring our identified design achieve significantly improved \emph{certifiable robustness} compared to models lacking this design.
However, like in our empirical robustness evaluation, architectural differences lead to some variability of robustness; the results also show tradeoffs between accuracy and robustness.
We also observe that the rankings under certifiable and empirical robustness are different, as in the previous results from \cite{geisler2020reliable}; {we discuss more in our \repo}.

\vspace{-0.4cm}
\section{Conclusion}

\label{sec:conclusion}
We formalized the relation between heterophily and adversarial structural attacks, and showed theoretically and empirically %
that effective attacks gravitate towards increasing heterophily in %
{both homophilous and heterophilous graphs by leveraging low-degree (gambit) nodes}. %
Using these insights, we showed that a key design addressing heterophily, namely separate aggregators for ego- and neighbor-embeddings, can lead to competitive improvement on  empirical and certifiable robustness, with only small influence on clean performance.
Finally, we compared the design with state-of-the-art vaccination mechanisms under different attack scenarios for various datasets, and illustrated 
that they are complementary and that their combination can lead to more robust GNN models.
We note that while we focus on the structural attacks, GNNs are also vulnerable to other types of attacks such as feature perturbations.
{We hope our analysis can inspire more effective defense strategies against adversarial attacks, especially designs that improve robustness by better addressing heterophily, such as heterophily in node features, or locally-occuring heterophily in homophilous graphs.}

\vspace{-0.2cm}
\section*{Acknowledgments}
This material is based upon work supported by the National Science Foundation under CAREER Grant No.~IIS 1845491 and Medium grant, Army Young Investigator Award No.~W911NF1810397, an Adobe Digital Experience research faculty award, an Amazon faculty award, a Google faculty award, and AWS Cloud Credits for Research. We gratefully acknowledge the support of NVIDIA Corporation with the donation of the Quadro P6000 GPU used for this research. 
MTS received funding from the Ministry of Culture and Science (MKW) of the German State of North Rhine-Westphalia (``NRW R\"uckkehrprogramm'') and the Excellence Strategy of the Federal Government and the Länder.
Any opinions, findings, and conclusions 
expressed in this material are those of the authors and do not necessarily reflect the views of the funding parties.

\vspace{-0.3cm}

\bibliographystyle{iclr2022_conference}
\bibliography{BIB/abbreviations,BIB/ACM-abbreviations,BIB/main,BIB/all}

\appendix
\newpage
\twocolumn[\section*{\LARGE \hspace{3cm} Supplementary Material on Reproducibility}
\vspace{0.4cm}
]

\newpage

For reproducibility, we describe main experimental setups at the beginning of \S\ref{sec:exp}, including the attack configurations, the depth and hidden unit size for GNNs, ratios for training, validation and test splits, and hardware specifications. 
We provide code and datasets on GitHub repository: \githubrepourl, where we also include additional details on the setups and results, including the implementations and detailed hyperparameters for GNNs and randomized smoothing (\S\ref{app:exp-details}), and additional results for evasion attacks (\S\ref{app:real-benchmark-results}) and certifiable robustness (\S\ref{app:real-cert-results}). {In addition, we report results for GNN models against \nettack{} on the larger Pubmed dataset in Table~\ref{table:results-pubmed}, with more details in \S\ref{app:real-benchmark-results}; GCN-SVD and SMGDC-based vaccinations run out of memory on Pubmed.}

\section{Proofs and Discussions of Theorems}
\label{app:proof}

For simplicity of mathematical expressions, we use $\columnmatrix{i}{s}{t}$ to refer a matrix where all elements in the $i$-th column are $s$, with all remaining elements (not in the $i$-th column) as $t$; we use $\columnvec{i}{s}{t}$ when the matrix is a row vector. We further denote $\circulantmatrix{s}{t}$ as a circulant matrix with all diagonal elements as $s$ and all off-diagonal elements as $t$, and $\blockmatrix{s}{t}$ as block matrix of the following form:
\vspace{-0.1cm}
\begin{equation*}
    \blockmatrix{s}{t} =
    \left[
    \begin{array}{cccc}
        \columnmatrix[\T]{1}{s}{t} & \columnmatrix[\T]{2}{s}{t} & \cdots & 
        \columnmatrix[\T]{|\setY|}{s}{t}
    \end{array} 
    \right]^\T
\end{equation*}
\vspace{-0.1cm}
\label{app:proof-thm1}

\vspace{-0.6cm}
\begin{proof}[\textbf{Proof for Thm.~\ref{thm:1}}]
\label{prf:thm1}
We give the proof in three parts: 
first, we analyze the training process of the GNN $f_s^{(2)}(\matA,\matX) = \bar{\matA}^2_\mathrm{s}\matX\matW$ on clean data and analytically derive the optimal weight matrix $\matW_*$ in a stylized learning setup; 
then, we construct a targeted evasion attack and calculate the attack loss for a unit structural attack; last, we summarize and validate the statements in the theorem. 

\item\paragraph{Stylized learning on clean data}
Given the 2-layer linearized GNN {\small $f_s^{(2)}(\matA,\matX) = \bar{\matA}^2_\mathrm{s}\matX\matW$} and 
the training set $\setT \subseteq \mathcal{D}_{\vertexSet}$, 
the goal of the training process is to optimize the weight matrix $\matW$ to minimize the cross-entropy loss function
$\mathcal{L}([\mathbf{z}]_{\setT, :}, [\V{Y}]_{\setT, :})$, 
where predictions $[\mathbf{z}]_{\setT, :} = [\bar{\matA}^2_\mathrm{s}\matX]_{\setT, :}\matW$ correspond to the predicted class label distributions for each node $v$ in the training set $\setT$, 
and $[\V{Y}]_{\setT, :}$ is the one-hot encoding of class labels provided in $\setT$.

Without loss of generality, we reorder $\setT$ accordingly such that the one-hot encoding of labels for nodes in the training set $[\V{Y}]_{\setT, :} = \blockmatrix{1}{0}$ is in increasing order of the class label $y_v$.
Now we look at the term $[\bar{\matA}^2_\mathrm{s}\matX]_{\setT, :}$ in $[\mathbf{z}]_{\setT, :} = [\bar{\matA}^2_\mathrm{s}\matX]_{\setT, :}\matW$, which are the feature vectors aggregated by the two GNN layers for nodes $v$ in the training set $\setT$. As stated in the theorem, we assume $\setT \subseteq \mathcal{D}_{\vertexSet}$, where node $u \in \mathcal{D}_{\vertexSet}$ have degree $d$; proportion $h$ of their neighbors belong to the same class, while proportion $\tfrac{1-h}{|\setY|-1}$ of them belong to any other class uniformly, and for each node $v \in \vertexSet$ the node features are given as $\V{x}_v = p\cdot \mathrm{onehot}(y_v) + \tfrac{1-p}{|\setY|}\cdot\mathbf{1}$ for each node $v \in \vertexSet$.
Then, after the first layer, we have $[\bar{\matA}_\mathrm{s}\matX]_{\setT, :} = 
\frac{1}{d+1} \blockmatrix{(h d + 1) p}{\frac{1-h}{|\setY|-1}}$, 
and after the second layer 
$[\bar{\matA}^2_\mathrm{s}\matX]_{\setT, :} = 
\frac{1}{(d+1)^2 |\setY| (|\setY| - 1)}\blockmatrix{S_1}{T_1}$, 
where $S_1 = \left((h |\setY|-1) d +|\setY| - 1\right)^2 p$, 
and $T_1 = \frac{\left((h |\setY|-1) d +|\setY| - 1\right)^2 p}{|\setY|-1}$.

For $[\V{Y}]_{\setT, :}$ and $[\bar{\matA}^2_\mathrm{s}\matX]_{\setT, :}$, 
we can find the optimal weight matrix $\matW_*$ such that 
$[\bar{\matA}^2_\mathrm{s}\matX]_{\setT, :}\matW_* = [\V{Y}]_{\setT, :}$, 
making the cross-entropy loss $\mathcal{L}([\mathbf{z}]_{\setT, :}, [\V{Y}]_{\setT, :}) = 0$. 
To find $\matW_*$, we can proceed as follows. 
First, sample one node from each class to form a smaller set $\mathcal{T}_S \subset \setT$.
Therefore, we have $[\V{Y}]_{\mathcal{T}_S, :} = \matI_{|\setY| \times |\setY|}$ and $[\bar{\matA}^2_\mathrm{s}\matX]_{\mathcal{T}_S, :} = \circulantmatrix{S_1}{T_1}$. 
Note that $[\bar{\matA}^2_\mathrm{s}\matX]_{\mathcal{T}_S, :}$ 
is a circulant matrix, and therefore its inverse exists. 
Using the Sherman-Morrison formula, we have
{\small
\begin{align}
    \label{eq:proof1-optimal-weights}
    \left([\bar{\matA}^2_\mathrm{s}\matX]_{\mathcal{T}_S, :}\right)^{-1} 
     = & \frac{(d+1)^2 (|\setY|-1)^2}{ p (d (h |\setY|-1)+|\setY|-1)^2 |\setY|} 
     \circulantmatrix{|\setY| - 1}{-1}.
\end{align}
}%
Now, let $\matW_* = \left([\bar{\matA}^2_\mathrm{s}\matX]_{\mathcal{T}_S, :}\right)^{-1} $, then
$[\mathbf{z}]_{\mathcal{T}_S, :}=[\bar{\matA}^2_\mathrm{s}\matX]_{\mathcal{T}_S, :} \matW_* = [\V{Y}]_{\mathcal{T}_S, :}=\matI_{|\setY| \times |\setY|}$. 
It is also easy to verify that 
$[\mathbf{z}]_{\setT, :} = [\V{Y}]_{\setT, :}$. We know $\matW_* = \left([\bar{\matA}^2_\mathrm{s}\matX]_{\mathcal{T}_S, :}\right)^{-1} $
is the optimal weight matrix that we can learn under $\setT$,
since $\matW_*$ satisfies $\mathcal{L}([\mathbf{z}]_{\setT, :}, [\V{Y}]_{\setT, :}) = 0$.

\item\paragraph{Attack loss under evasion attacks}
Now consider an arbitrary target node $v \in \mathcal{D}_{\vertexSet}$ with class label $y_v \in \setY$, and a unit structural perturbation leveraging gambit node $u \in \vertexSet$ with degree $d_a$ that affects the predictions $\mathbf{z}_v$ of node $v$ made by GNN $f_s^{(2)}$. 
Without loss of generality, we assume node $v$ has $y_v = 1$.
As $f_s^{(2)}$ contains 2 GNN layers with each layer aggregating feature vectors within neighborhood $N(v)$ of each node $v$, the perturbation must take place in the direct (1-hop) neighborhood $N(v)$ or 2-hop neighborhood $N_2(v)$ to affect the predictions $\mathbf{z}_v$. For the unit perturbation, the attacker can add or remove a homophilous edge or path between nodes $u$ and $v$, which we denote as $\delta_1$ ($\delta_1 = 1$ for addition and $\delta_1 = -1$ for removal); alternatively, the attacker can add or remove a heterophilous edge or path between nodes $u$ and $v$, 
which we denote as $\delta_2 = \pm 1$ analogously. 
We denote the perturbed graph adjacency matrix as $\bar{\matA}'_\mathrm{s}$, 
and $\mathbf{z}'_v = [\bar{\matA}'^{2}_\mathrm{s}\matX]_{v, :}\matW_*$

\textcircled{\raisebox{-0.9pt}{1}}
\emph{Unit perturbation in direct neighborhood $N(v)$}. We first consider a unit perturbation in the direct (1-hop) neighborhood $N(v)$ of node $v$. For simplicity of derivation, we assume that the perturbation does not change the row-stochastic normalization of $\bar{\matA}_\mathrm{s}$, and only affects the aggregated feature vectors of the target node $v$. 

In the case of $\delta_1 = \pm 1$ and $\delta_2 = 0$, we have 
$[\bar{\matA}'_\mathrm{s}\matX]_{v, :} -  [\bar{\matA}_\mathrm{s}\matX]_{v, :} 
= \frac{\delta _1}{d_a+1} \columnvec{1}{\left(\frac{1-p}{|\setY|}+p\right)}{\left(\frac{1-p}{|\setY|}\right)}$,
and 
\vspace{-0.3cm}
$$[\bar{\matA}'^{2}_\mathrm{s}\matX]_{v, :} -  [\bar{\matA}^2_\mathrm{s}\matX]_{v, :} = 
\frac{\delta _1}{(d_a+1)^2(d + 1) |\setY|} \columnvec{1}{S_2}{\frac{T_2}{|\setY|-1}},$$
where $\small S_2 =  d_a (p (d (h |\setY|-1)+h |\setY|+|\setY|-2)+d+2)+(d+2) (|\setY|-1) p+d+2$ 
and $\small T_2 = -d_a (p (d (h |\setY|-1)+h |\setY|+|\setY|-2)+(-d-2) (|\setY|-1))-(d+2) (|\setY|-1) (p-1)$. 
By Multiplying $[\bar{\matA}'^{2}_\mathrm{s}\matX]_{v, :}$ by  $\matW_*$, we can get the predictions $\mathbf{z}'_v$ after perturbations; we omit the analytical expression of $\mathbf{z}'_v$ here due to its complexity.

On the perturbed graph, the CM-type attack loss is calculated as $\attackLoss^{\mathrm{CM}}(\mathbf{z}'_{v}) = - (\mathbf{z}'_{v, y_v} - \max_{y \neq y_v} \mathbf{z}'_{v, y})$.
Since $\attackLoss^{\mathrm{CM}}(\mathbf{z}_{v}) = -1$ on clean data, the change in attack loss before and after attack is 
{\small
\begin{align}
    \Delta\attackLoss^{\mathrm{CM}} 
    & = \attackLoss^{\mathrm{CM}}(\mathbf{z}'_{v}) - \attackLoss^{\mathrm{CM}}(\mathbf{z}_{v}) 
    = \frac{- \delta _1 (d+1) (|\setY|-1) S_3}{(d_a+1)^2 (d (h |\setY|-1)+|\setY|-1)^2},
    \label{eq:app-proof-1-direct-delta-1-cm-loss}
\end{align}
}%
where $S_3 =  \left(d_a (d (h |\setY|-1)+h |\setY|+|\setY|-2)+(d+2) (|\setY|-1)\right)$.
Solving the system of inequalities for $\delta_1$ under constraints $\Delta\attackLoss^{\mathrm{CM}} > 0$, $h \in [0, 1]$, $|\setY| \geq 2$ and $d, d_a, |\setY| \in \mathbb{Z}^{+}$, 
we get the range of $\delta_1$ as
\begin{equation}
    \label{eq:app-proof-1-direct-delta-1-cm-loss-solution}
    \begin{cases}
        \delta_1 < 0, \text{ when } 0 < d \leq |\setY| - 2 \\
        \delta_1 < 0, \text{ when } d > |\setY| - 2 \text{ and } d_a < d_1 \\
        \delta_1 < 0, \text{ when } d > |\setY| - 2 \text{ and } d_a \geq d_1 \text{ and } 1 \geq h > h_1 \\
        \delta_1 > 0, \text{ when } d > |\setY| - 2 \text{ and } d_a \geq d_1 \text{ and } 0 \leq h < h_1 \\
    \end{cases},
\end{equation}
where $d_1 = \frac{(d+2) (|\setY|-1)}{d-|\setY|+2}$ and $h_1 = \tfrac{d_a (d-|\setY|+2)-(d+2) (|\setY|-1)}{(d+1) |\setY| d_a}$.
Note that the above solution is not applicable when $h = \frac{d-|\setY|+1}{d |\setY|}$, in which case $d (h |\setY|-1)+|\setY|-1=0$ and the solution of optimal weight matrix $\matW_* = \left([\bar{\matA}^2_\mathrm{s}\matX]_{\mathcal{T}_S, :}\right)^{-1} $ is undefined.

\vspace{0.5\baselineskip}

In the case of $\delta_1 = 0$ and $\delta_2 = \pm 1$, 
following a similar derivation to that in the previous case, we can compute the change in the CM-type attack loss before and after attack as
\begin{align}
    \Delta\attackLoss^{\mathrm{CM}} 
    = \frac{\delta _2 (d+1) (|\setY|-1) S_3}{\left(d_a+1\right){}^2 (d (h |\setY|-1)+|\setY|-1)^2}.
    \label{eq:app-proof-1-direct-delta-2-cm-loss}
\end{align}
Solving the same system of inequalities as 
in the previous case for $\delta_2$, we have $\delta_2 = -\delta_1$, where $\delta_1$ is bounded by 
Eq. \eqref{eq:app-proof-1-direct-delta-1-cm-loss-solution}.
Note that the above solution is again not applicable when $h = \frac{d-|\setY|+1}{d |\setY|}$, 
and we always have 
$h_1 - \frac{1}{|\setY|} = -\frac{(|\setY|-1) \left(d_a+d+2\right)}{(d+1) |\setY| d_a} < 0$ when $|\setY| \geq 2$.

\textcircled{\raisebox{-0.9pt}{2}}
\emph{Unit perturbation in 2-hop neighborhood $N_2(v)$}.
We now consider a unit perturbation in the 2-hop neighborhood $N(v)$ of node $v$. In this case we will have $[\bar{\matA}'_\mathrm{s}\matX]_{v, :} = [\bar{\matA}_\mathrm{s}\matX]_{v, :}$.

In the case of $\delta_1 = \pm 1$ and $\delta_2 = 0$, we have 
$[\bar{\matA}'^{2}_\mathrm{s}\matX]_{v, :} -  [\bar{\matA}^2_\mathrm{s}\matX]_{v, :} = 
\frac{\delta _1}{(d_a+1)^2 |\setY|} \columnvec{1}{S_5}{\frac{T_5}{|\setY|-1}}$,
where $$S_5 = \left(d_a (p (h |\setY|-1)+1)+(|\setY|-1) p+1\right)$$ and $T_5 = \left(d_a (-h |\setY| p+|\setY|+p-1)+|\setY| (-p)+|\setY|+p-1\right).$ 
By multiplying $[\bar{\matA}'^{2}_\mathrm{s}\matX]_{v, :}$ with $\matW_*$, we can get the predictions $\mathbf{z}'_v$ after perturbations. Following a similar derivation as before, we can get the change in the CM-type attack loss before and after attack as
\begin{align}
    \Delta\attackLoss^{\mathrm{CM}} 
    = -\frac{(d+1)^2 (|\setY|-1) \left(d_a (h |\setY|-1)+|\setY|-1\right) \delta _1}{\left(d_a+1\right){}^2 (d (h |\setY|-1)+|\setY|-1)^2}.
    \label{eq:app-proof-1-indirect-delta-1-cm-loss}
\end{align}
Solving for $\delta_1$ as in previous cases, we get the valid range of $\delta_1$ as
\begin{equation}
    \label{eq:app-proof-1-indirect-delta-1-cm-loss-solution}
    \begin{cases}
        \delta_1 < 0, 
        \text{ when } d_a < |\setY| -1 \\
        \delta_1 < 0, 
        \text{ when } d_a \geq |\setY| -1
        \text{ and } \frac{d_a-|\setY|+1}{|\setY| d_a} < h \leq 1 \\
        \delta_1 > 0, 
        \text{ when } d_a > |\setY| -1 
        \text{ and } 0 \leq h < \frac{d_a-|\setY|+1}{|\setY| d_a} \\
    \end{cases}.
\end{equation}
Note that the above solution is not applicable when $h = \frac{d-|\setY|+1}{d |\setY|}$, in which case $d (h |\setY|-1)+|\setY|-1=0$ and the solution of optimal weight matrix $\matW_* = \left([\bar{\matA}^2_\mathrm{s}\matX]_{\mathcal{T}_S, :}\right)^{-1} $ is undefined.

\vspace{0.5\baselineskip}

For the case $\delta_1 = 0$ and $\delta_2 = \pm 1$, 
following a similar derivation to that in the previous case, we can compute the change in the CM-type attack loss before and after attack as
\vspace{-0.2cm}
\begin{align}
    \Delta\attackLoss^{\mathrm{CM}} 
    = \frac{(d+1)^2 (|\setY|-1) \left(d_a (h |\setY|-1)+|\setY|-1\right) \delta _2}{\left(d_a+1\right){}^2 (d (h |\setY|-1)+|\setY|-1)^2}
    \label{eq:app-proof-1-indirect-delta-2-cm-loss}
\end{align}
Solving the same system of inequalities as 
in the previous cases for $\delta_2$, we have $\delta_2 = -\delta_1$, where $\delta_1$ is bounded by 
Eq. \eqref{eq:app-proof-1-indirect-delta-1-cm-loss-solution}.
Note that the above solution is again not applicable when $h = \frac{d-|\setY|+1}{d |\setY|}$,
and we always have $\frac{d_a-|\setY|+1}{|\setY| d_a} - \frac{1}{|\setY|} = \frac{1-|\setY|}{|\setY| d_a} < 0$ when $|\setY| \geq 2$. 

\item\paragraph{Summary and validation of theorem statements} Based on our discussions, we validate our statements in the theorem next. 

\textcircled{\raisebox{-0.9pt}{1}} 
From Eq. \eqref{eq:app-proof-1-direct-delta-1-cm-loss-solution}, 
\eqref{eq:app-proof-1-indirect-delta-1-cm-loss-solution}, 
we observe that for both direct attacks in 1-hop neighborhood $N(v)$ and indirect attacks in 2-hop neighborhood $N_2(v)$, when $h \geq \frac{1}{|\setY|}$, the attack loss increases only if $\delta_1 < 0$ (i.e., removal of a homophilous edge or path to node $v$), or if $\delta_2 > 0$ (i.e., addition of a heterophilous edge or path to node $v$). 

\textcircled{\raisebox{-0.9pt}{2}} 
From Eq. \eqref{eq:app-proof-1-direct-delta-1-cm-loss}, \eqref{eq:app-proof-1-direct-delta-2-cm-loss}, \eqref{eq:app-proof-1-indirect-delta-1-cm-loss} and \eqref{eq:app-proof-1-indirect-delta-2-cm-loss}, we can show that $\frac{\Delta\attackLoss^{\mathrm{CM, direct}}}{\Delta\attackLoss^{\mathrm{CM, indirect}}} > 1$

when $h \geq \frac{1}{|\setY|}$,
which means we will always have $\Delta\attackLoss^{\mathrm{CM, direct}} > \Delta\attackLoss^{\mathrm{CM, indirect}} > 0$ for a unit perturbation increasing loss $\attackLoss^{\mathrm{CM}}$.
\end{proof}

\label{app:proof-thm-heterophily}

\begin{proof}[\textbf{Proof for Thm.~\ref{thm:heterophily}}]
\label{prf:thm-heterophily}
The theorem statements can be directly validated from Eq. \eqref{eq:app-proof-1-direct-delta-1-cm-loss-solution} and its discussions in Proof for Thm.~\ref{thm:1}, which solves $\delta_1$ and $\delta_2$ for an effective unit perturbation in direct neighborhood $N(v)$ of node $v$.     
We similarly note that the conclusions are not applicable when $h = \frac{d-|\setY|+1}{d |\setY|}$, in which case the solution of optimal weight matrix $\matW_*$ is undefined. 

\end{proof}

\label{app:proof-thm2}
\begin{proof}[\textbf{Proof for Thm.~\ref{thm:2}}] 
\label{prf:thm2}
Here, we mainly focus on analyzing $\Delta\attackLoss$ for the layer defined as $f(\matA,\matX;\alpha) = \left((1-\alpha) \bar{\matA} + \alpha \matI \right) \matX \matW$, as the layer defined as $f_s(\matA,\matX)=\bar{\matA}_\mathrm{s}\matX\matW$ is a special case when $\alpha = \frac{1}{1+d}$. We follow a similar process as in Proof of Thm.~\ref{thm:1}.

\item\paragraph{Layer $f(\matA,\matX;\alpha) = \left((1-\alpha) \bar{\matA} + \alpha \matI \right) \matX \matW$} 
We first derive the optimal weight matrix $\matW_*$ in a stylized learning setup as in Proof~\ref{prf:thm1}. Following a similar process, for this GNN layer we have
{\small $$\left[\left((1-\alpha) \bar{\matA} + \alpha \matI \right) \matX\right]_{\mathcal{T}_S, :} = 
\frac{1-p}{|\setY|} + \circulantmatrix{p (\alpha+h-\alpha h)}{\frac{(\alpha-1) (h-1) p}{|\setY|-1}}$$}

and
$\matW_* = \frac{|\setY|-1}{p (a (h-1) |\setY|-h |\setY|+1)|\setY|} \cdot \circulantmatrix{1-|\setY|}{1}.$

Now as in Proof for Thm.~\ref{thm:1}, we consider an arbitrary target node $v \in \mathcal{D}_{\vertexSet}$, and a unit structural perturbation leveraging gambit node $u \in \vertexSet$ with degree $d_a$ that affects the predictions $\mathbf{z}_v$ of node $v$ made by layer $f(\matA,\matX;\alpha)$. 
Note that we only discuss the case of direct structural perturbation to the 1-hop neighborhood $N(v)$ of target node $v$, as indirect perturbations do not affect the predictions $\mathbf{z}_v$ for node $v$ produced by a single GNN layer. 
Following a similar derivation as in Proof for Thm.~\ref{thm:1}, 
in the case of $\delta_1 = \pm 1$ and $\delta_2 = 0$, or $\delta_1 = 0$ and $\delta_2 = \pm 1$, 
the change in the CM-type attack loss $\Delta\attackLoss^{\mathrm{CM}}$ for GNN layer $f(\matA,\matX;\alpha)$ considering both $\delta_1$ and $\delta_2$ is
\begin{align}
    \Delta\attackLoss^{\mathrm{CM,f}} 
    = \frac{((1-\alpha)|\setY|+\alpha-1)\delta _1}{d_a  (\alpha (h-1)-h) |\setY|+d_a} + \frac{(\alpha-1) (|\setY|-1) \delta _2}{d_a (\alpha (h-1)-h)|\setY|+d_a}.
\end{align}

\item\paragraph{Layer $f_s(\matA,\matX)=\bar{\matA}_\mathrm{s}\matX\matW$}
This formulation is a special case of the previously discussed $f(\matA,\matX;\alpha)$ formulation when $\alpha = \frac{1}{1+d_a}$. 
We denote the change in the attack loss for this layer as $\Delta\attackLoss^{\mathrm{CM,fs}}$.

\item\paragraph{Comparison of increase in attack loss $\Delta\attackLoss^{\mathrm{CM}}$}
Solving the following system of inequalities for variable $\alpha$
under constraints $\Delta\attackLoss^{\mathrm{CM,fs}}  > \Delta\attackLoss^{\mathrm{CM,f}} > 0, \alpha, h \in [0, 1], |\setY| \geq 2, d_a, |\setY| \in \mathbb{Z}^{+}$ and $\delta_1, \delta_2 \in \{-1, 0, 1\},$
we get the valid range of $\alpha$ as
{\small
\begin{equation}
    \label{eq:app-proof-2-solution}
    \begin{cases}
        \frac{1}{d_a+1}<\alpha<1, 
        \text{ when } 0 \leq h < \frac{1}{|\setY|} 
        \text{ and } 0 < d_a < \frac{1- |\setY|}{h |\setY|-1} 
        \text{ and } \delta _1 < \delta _2 \\
        0\leq \alpha<\frac{1}{d_a+1}, 
        \text{ when } 0 \leq h < \frac{1}{|\setY|} 
        \text{ and } d_a>\frac{1-|\setY|}{h |\setY|-1}
        \text{ and } \delta _1 > \delta _2 \\
        \frac{1}{d_a+1}<\alpha<1, 
        \text{ when } \frac{1}{|\setY|}\leq h\leq 1
        \text{ and } \delta _1 < \delta _2 \\
    \end{cases}.
\end{equation}
}%
From Eq.~\eqref{eq:app-proof-2-solution}, we observe that when $h > \tfrac{1}{|\setY|}$, a unit perturbation increasing $\attackLoss$ as discussed in Thm.~\ref{thm:1} (i.e. $\delta_1 = -1$ and $\delta_2 = 0$, or $\delta_1 = 0$ and $\delta_2 = 1$), which
satisfys the condition $\delta _1 < \delta _2$, will thus lead to a strictly smaller increase  in $\Delta\attackLoss^{\mathrm{CM,f}}$ for layer $f(\matA,\matX;\alpha)$ than the increase $\Delta\attackLoss^{\mathrm{CM,fs}}$ for layer $f_s(\matA,\matX)$ if $\alpha > \tfrac{1}{d_a+1}$. 
\end{proof}

\clearpage
\newpage
\onecolumn\section*{\LARGE \hspace{3.5cm} Additional Details on Empirical Evaluation}
\vspace{0.4cm}

\section{Detailed Experimental Setups and Hyperparameters}
\label{app:exp-details}

\subsection{More Details on the Experimental Setup} \label{app:exp-details-setups}

\paragraph{Attack Implementations} We incorporate the following implementations of attacks from \texttt{DeepRobust}~\citep{li2020deeprobust} to our empirical framework. %

\begin{table}[htb]
    \centering
    \label{table:attack-sources}
    \begin{tabular}{@{}ll@{}}
        \toprule
        \textbf{\nettack}~\citep{zugner2018adversarial} & \url{https://github.com/DSE-MSU/DeepRobust/blob/master/deeprobust/graph/targeted_attack/nettack.py} \\
        
        \noalign{\vskip 0.25ex}
        \cdashline{1-2}[0.8pt/2pt]
        \noalign{\vskip 0.25ex}
        
        \textbf{Metattack}~\citep{zugner2019adversarial} & \url{https://github.com/DSE-MSU/DeepRobust/blob/master/deeprobust/graph/global_attack/mettack.py}\\
        \bottomrule
    \end{tabular}
\end{table}

\paragraph{Benchmark Implementations} \label{appendix-setup-baseline}
Our empirical framework is built on DeepRobust~\citep{li2020deeprobust}, Python Fire\footnote{\url{https://github.com/google/python-fire}} and signac\footnote{\url{https://signac.io}}. 
We incorporated the following implementations of GNN models in our framework. 
For GNNGuard and GCN-SVD, there are some implementation ambiguities, which we discuss in the next paragraph. 
\begin{table}[htb]
    \centering
    \label{table:benchmark-sources}
    \begin{tabular}{@{}ll@{}}
        \toprule
        \textbf{\method}~\citep{zhu2020beyond} & \url{https://github.com/GemsLab/H2GCN}\\
        \textbf{GraphSAGE}~\citep{hamilton2017inductive} & Implemented on top of \url{https://github.com/GemsLab/H2GCN} \\
        \textbf{CPGNN}~\citep{zhu2020graph} & \url{https://github.com/GemsLab/CPGNN} \\
        \textbf{GPR-GNN}~\citep{chien2021adaptive} & \url{https://github.com/jianhao2016/GPRGNN} \\
        \textbf{FAGCN}~\citep{bo2021beyond} & \url{https://github.com/bdy9527/FAGCN} \\
        \textbf{APPNP}~\citep{klicpera2018predict} & Our own implementation \\
        
        \noalign{\vskip 0.25ex}
        \cdashline{1-2}[0.8pt/2pt]
        \noalign{\vskip 0.25ex}
        
        \textbf{GNNGuard}\cellcolor{gray!20}~\citep{zhang2020gnnguard} &\url{https://github.com/mims-harvard/GNNGuard}\\

        \textbf{ProGNN}~\citep{jin2020graph} & \url{https://github.com/ChandlerBang/Pro-GNN}\\

        \textbf{GCN-SVD}\cellcolor{gray!20}~\citep{entezari2020all} &\url{https://github.com/DSE-MSU/DeepRobust/blob/master/examples/graph/test_gcn_svd.py}\\

        \textbf{GCN-SMGDC}~\citep{geisler2020reliable,klicpera_diffusion_2019} & 
        \textbf{Soft Medoid}: \url{https://github.com/sigeisler/reliable_gnn_via_robust_aggregation/blob/main/rgnn/means.py#L358} \\
        & \textbf{GDC}: \url{https://github.com/klicperajo/gdc/blob/master/gdc_demo.ipynb} \\
        & \textbf{GCN}: Implemented on top of \url{https://github.com/GemsLab/H2GCN} \\
        
        \noalign{\vskip 0.25ex}
        \cdashline{1-2}[0.8pt/2pt]
        \noalign{\vskip 0.25ex}
        
        \textbf{GAT}~\citep{velickovic2018graph} & \url{https://github.com/DSE-MSU/DeepRobust/blob/master/deeprobust/graph/defense/gat.py}\\
        \textbf{GCN}~\citep{kipf2016semi} & 
        Implemented on top of \url{https://github.com/GemsLab/H2GCN} \\
        \bottomrule
    \end{tabular}
\end{table}

\paragraph{Notes on the GNNGuard and GCN-SVD Implementations}
We note that there are ambiguities in the implementations of GNNGuard~\citep{zhang2020gnnguard} and GCN-SVD~\citep{entezari2020all}, which can lead to different variants %
with different performance and robustness, as we show in Table~\ref{table:fixation-clean}.

\begin{table}[hbt]

    \centering
    \caption{Performance comparison between variants of GNNGuard and GCN-SVD: mean accuracy $\pm$ stdev over multiple sets of experiments.}
    \label{table:fixation-clean}
    \vspace{0.1cm}
    \resizebox{\columnwidth}{!}{
    \begin{tabular}{rr rr c rr c rr c rr}
        \toprule

        && \multicolumn{2}{c}{\bf Homophilous graphs} && \multicolumn{2}{c}{\bf Heterophilous graphs} &
        & \multicolumn{2}{c}{\bf Homophilous graphs} && \multicolumn{2}{c}{\bf Heterophilous graphs}\\
        \cmidrule{3-4} \cmidrule{6-7}  \cmidrule{9-10} \cmidrule{12-13}
         && \multicolumn{1}{c}{\texttt{\bf Cora}}  & \multicolumn{1}{c}{\texttt{\bf Citeseer}}  && \multicolumn{1}{c}{\texttt{\bf FB100}} & \multicolumn{1}{c}{\texttt{\bf Snap}} & & \multicolumn{1}{c}{\texttt{\bf Cora}}  & \multicolumn{1}{c}{\texttt{\bf Citeseer}}  && \multicolumn{1}{c}{\texttt{\bf FB100}} & \multicolumn{1}{c}{\texttt{\bf Snap}}
        \\
        && \multicolumn{1}{c}{$h$=0.804} & \multicolumn{1}{c}{$h$=0.736} && \multicolumn{1}{c}{$h$=0.531} & \multicolumn{1}{c}{$h$=0.134} && \multicolumn{1}{c}{$h$=0.804} & \multicolumn{1}{c}{$h$=0.736} && \multicolumn{1}{c}{$h$=0.531} & \multicolumn{1}{c}{$h$=0.134}  \\
        
       \cmidrule{3-4} \cmidrule{6-7}  \cmidrule{9-10} \cmidrule{12-13}
        
        &&  \multicolumn{5}{c}{\textbf{Clean Datasets}} 
        && \multicolumn{5}{c}{\textbf{Clean Datasets}}
        \\
        \cmidrule{1-1} \cmidrule{3-7}  \cmidrule{9-13}
        \textbf{GNNGuard (\scaleroman{I})} &       &  75.56\tiny{$\pm$5.15} &  70.00\tiny{$\pm$6.24}& &  68.89\tiny{$\pm$2.08} &             31.67\tiny{$\pm$0.00}& &  79.58\tiny{$\pm$0.97} &  71.68\tiny{$\pm$1.10}& &  65.31\tiny{$\pm$1.48} &                  26.37\tiny{$\pm$0.70} \\
        \textbf{GNNGuard (II)} &       &  77.22\tiny{$\pm$6.29} &  67.78\tiny{$\pm$4.78}& &  67.22\tiny{$\pm$2.08} &             28.33\tiny{$\pm$3.60}& &  80.15\tiny{$\pm$0.55} &  72.61\tiny{$\pm$0.28}& &  65.66\tiny{$\pm$0.60} &                  26.51\tiny{$\pm$0.98} \\

         \noalign{\vskip 0.25ex}
        \cdashline{1-1}[0.8pt/2pt]
        \cdashline{3-7}[0.8pt/2pt]
        \cdashline{9-13}[0.8pt/2pt]
        \noalign{\vskip 0.25ex}

        \textbf{GCN-SVD (\scaleroman{I}) $k=5\ \ $} &  &  66.67\tiny{$\pm$8.16} &  63.89\tiny{$\pm$5.50}& &  51.67\tiny{$\pm$6.24} &             29.44\tiny{$\pm$0.79}& &  69.43\tiny{$\pm$0.99} &  68.31\tiny{$\pm$0.34}& &  52.95\tiny{$\pm$0.13} &                  27.66\tiny{$\pm$0.05} \\
        \textbf{GCN-SVD (II) $k=5\ \ $} &  &  52.78\tiny{$\pm$5.50} &  35.00\tiny{$\pm$1.36}& &  50.56\tiny{$\pm$4.37} &             25.00\tiny{$\pm$5.93}& &  55.05\tiny{$\pm$1.77} &  41.47\tiny{$\pm$0.72}& &  52.40\tiny{$\pm$0.18} &                  25.84\tiny{$\pm$0.07} \\

        \noalign{\vskip 0.25ex}
        \cdashline{1-1}[0.8pt/2pt]
        \cdashline{3-7}[0.8pt/2pt]
        \cdashline{9-13}[0.8pt/2pt]
        \noalign{\vskip 0.25ex}

        \textbf{GCN-SVD (\scaleroman{I}) $k=10$} & &  66.11\tiny{$\pm$6.71} &  65.00\tiny{$\pm$3.60}& &  52.78\tiny{$\pm$4.16} &             30.56\tiny{$\pm$2.08}& &  71.08\tiny{$\pm$0.46} &  69.19\tiny{$\pm$1.13}& &  54.47\tiny{$\pm$0.32} &                  27.57\tiny{$\pm$0.18} \\
        \textbf{GCN-SVD (II) $k=10$} & &  66.11\tiny{$\pm$4.78} &  45.00\tiny{$\pm$3.60}& &  51.11\tiny{$\pm$2.08} &             22.22\tiny{$\pm$4.78}& &  64.79\tiny{$\pm$1.56} &  52.17\tiny{$\pm$0.39}& &  51.19\tiny{$\pm$0.41} &                  25.45\tiny{$\pm$0.21} \\

        \noalign{\vskip 0.25ex}
        \cdashline{1-1}[0.8pt/2pt]
        \cdashline{3-7}[0.8pt/2pt]
        \cdashline{9-13}[0.8pt/2pt]
        \noalign{\vskip 0.25ex}

        \textbf{GCN-SVD (\scaleroman{I}) $k=15$} & &  72.78\tiny{$\pm$6.98} &  63.89\tiny{$\pm$7.74}& &  57.78\tiny{$\pm$2.83} &             28.89\tiny{$\pm$2.08}& &  72.74\tiny{$\pm$0.29} &  66.51\tiny{$\pm$1.53}& &  57.67\tiny{$\pm$0.36} &                  27.61\tiny{$\pm$0.55} \\
        \textbf{GCN-SVD (II) $k=15$} & &  69.44\tiny{$\pm$2.08} &  46.67\tiny{$\pm$6.24}& &  52.78\tiny{$\pm$5.15} &             21.67\tiny{$\pm$1.36}& &  65.61\tiny{$\pm$0.19} &  60.55\tiny{$\pm$0.73}& &  53.24\tiny{$\pm$0.45} &                  26.63\tiny{$\pm$0.25} \\

        \noalign{\vskip 0.25ex}
        \cdashline{1-1}[0.8pt/2pt]
        \cdashline{3-7}[0.8pt/2pt]
        \cdashline{9-13}[0.8pt/2pt]
        \noalign{\vskip 0.25ex}

        \textbf{GCN-SVD (\scaleroman{I}) $k=50$} & &  78.89\tiny{$\pm$6.29} &  66.67\tiny{$\pm$3.60}& &  65.56\tiny{$\pm$2.83} &             31.11\tiny{$\pm$0.79}& &  77.98\tiny{$\pm$0.43} &  68.25\tiny{$\pm$0.86}& &  63.41\tiny{$\pm$0.45} &                  27.81\tiny{$\pm$0.39} \\
        \textbf{GCN-SVD (II) $k=50$} &  \multirow{-10}{*}{\rotatebox[origin=r]{90}{\textsc{Nettack}}} &  75.56\tiny{$\pm$4.16} &  59.44\tiny{$\pm$0.79}& &  55.00\tiny{$\pm$1.36} &             27.78\tiny{$\pm$6.71}&
        \multirow{-10}{*}{\rotatebox[origin=r]{90}{Metattack}}
        &  76.61\tiny{$\pm$0.31} &  66.90\tiny{$\pm$0.16}& &  55.47\tiny{$\pm$0.23} &                  25.62\tiny{$\pm$0.12} \\
        
    \midrule
    
    &&
    \multicolumn{5}{c}{\textbf{Poison Attacks}} 
        && \multicolumn{5}{c}{\textbf{Poison Attacks}} \\
        
        \cmidrule{1-1} \cmidrule{3-7}  \cmidrule{9-13}
         \textbf{GNNGuard (\scaleroman{I})} &       &  57.22\tiny{$\pm$2.08} &  60.00\tiny{$\pm$3.60}& &   0.56\tiny{$\pm$0.79} &             11.11\tiny{$\pm$0.79}& &  73.68\tiny{$\pm$0.99} &  67.89\tiny{$\pm$0.92}& &  60.82\tiny{$\pm$0.45} &                  23.98\tiny{$\pm$0.71} \\
         \textbf{GNNGuard (II)} &       &  58.33\tiny{$\pm$1.36} &  59.44\tiny{$\pm$3.14}& &   0.56\tiny{$\pm$0.79} &              9.44\tiny{$\pm$1.57}& &  74.20\tiny{$\pm$0.55} &  68.13\tiny{$\pm$0.74}& &  60.89\tiny{$\pm$0.48} &                  23.78\tiny{$\pm$0.67} \\

        \noalign{\vskip 0.25ex}
        \cdashline{1-1}[0.8pt/2pt]
        \cdashline{3-7}[0.8pt/2pt]
        \cdashline{9-13}[0.8pt/2pt]
        \noalign{\vskip 0.25ex}

        \textbf{GCN-SVD (\scaleroman{I}) $k=5\ \ $} &  &  64.44\tiny{$\pm$9.06} &  60.00\tiny{$\pm$4.71}& &  41.67\tiny{$\pm$6.24} &             27.78\tiny{$\pm$6.29}& &  64.65\tiny{$\pm$2.57} &  66.35\tiny{$\pm$1.48}& &  53.14\tiny{$\pm$0.43} &                  25.64\tiny{$\pm$0.47} \\
        \textbf{GCN-SVD (II) $k=5\ \ $} &  &  44.44\tiny{$\pm$2.83} &  33.33\tiny{$\pm$2.72}& &  41.67\tiny{$\pm$2.36} &             25.00\tiny{$\pm$6.80}& &  42.19\tiny{$\pm$5.33} &  40.17\tiny{$\pm$1.57}& &  51.87\tiny{$\pm$0.38} &                  24.82\tiny{$\pm$0.43} \\

        \noalign{\vskip 0.25ex}
        \cdashline{1-1}[0.8pt/2pt]
        \cdashline{3-7}[0.8pt/2pt]
        \cdashline{9-13}[0.8pt/2pt]
        \noalign{\vskip 0.25ex}
        
        \textbf{GCN-SVD (\scaleroman{I}) $k=10$} & &  67.78\tiny{$\pm$5.50} &  57.78\tiny{$\pm$1.57}& &  35.56\tiny{$\pm$1.57} &             31.67\tiny{$\pm$5.93}& &  65.54\tiny{$\pm$1.28} &  65.46\tiny{$\pm$0.92}& &  55.68\tiny{$\pm$0.15} &                  25.93\tiny{$\pm$0.75} \\
        \textbf{GCN-SVD (II) $k=10$} & &  48.89\tiny{$\pm$3.14} &  31.67\tiny{$\pm$2.36}& &  34.44\tiny{$\pm$0.79} &             26.11\tiny{$\pm$6.85}& &  49.92\tiny{$\pm$5.88} &  47.16\tiny{$\pm$3.93}& &  53.16\tiny{$\pm$0.45} &                  25.30\tiny{$\pm$0.28} \\

        \noalign{\vskip 0.25ex}
        \cdashline{1-1}[0.8pt/2pt]
        \cdashline{3-7}[0.8pt/2pt]
        \cdashline{9-13}[0.8pt/2pt]
        \noalign{\vskip 0.25ex}

        \textbf{GCN-SVD (\scaleroman{I}) $k=15$} & &  64.44\tiny{$\pm$3.93} &  52.78\tiny{$\pm$4.78}& &  23.89\tiny{$\pm$6.29} &             29.44\tiny{$\pm$3.93}& &  65.46\tiny{$\pm$2.33} &  61.04\tiny{$\pm$1.04}& &  58.06\tiny{$\pm$0.05} &                  25.83\tiny{$\pm$0.69} \\
        \textbf{GCN-SVD (II) $k=15$} & &  51.11\tiny{$\pm$3.42} &  33.89\tiny{$\pm$3.93}& &  30.56\tiny{$\pm$2.83} &             26.11\tiny{$\pm$6.14}& &  50.30\tiny{$\pm$3.80} &  47.87\tiny{$\pm$1.31}& &  54.20\tiny{$\pm$0.36} &                  25.25\tiny{$\pm$0.91} \\

        \noalign{\vskip 0.25ex}
        \cdashline{1-1}[0.8pt/2pt]
        \cdashline{3-7}[0.8pt/2pt]
        \cdashline{9-13}[0.8pt/2pt]
        \noalign{\vskip 0.25ex}

        \textbf{GCN-SVD (\scaleroman{I}) $k=50$} & &  61.67\tiny{$\pm$4.71} &  48.33\tiny{$\pm$7.07}& &  16.67\tiny{$\pm$4.08} &             30.56\tiny{$\pm$7.97}& &  60.06\tiny{$\pm$5.43} &  49.31\tiny{$\pm$4.52}& &  62.07\tiny{$\pm$0.69} &                  26.05\tiny{$\pm$0.63} \\
        \textbf{GCN-SVD (II) $k=50$} & \multirow{-10}{*}{\rotatebox[origin=r]{90}{\textsc{Nettack}}} &  53.33\tiny{$\pm$4.91} &  28.89\tiny{$\pm$2.08}& &  23.33\tiny{$\pm$2.72} &             25.00\tiny{$\pm$5.44}& \multirow{-10}{*}{\rotatebox[origin=r]{90}{Metattack}} &  47.82\tiny{$\pm$7.59} &  51.20\tiny{$\pm$1.78}& &  55.00\tiny{$\pm$2.06} &                  25.18\tiny{$\pm$0.98} \\

    \midrule

    &&  \multicolumn{5}{c}{\textbf{Evasion Attacks}} 
    & &
        \multicolumn{5}{c}{\textbf{Evasion Attacks}} \\
         \cmidrule{1-1} \cmidrule{3-7}  \cmidrule{9-13}
        \textbf{GNNGuard (\scaleroman{I})} &       &     - &     -& &     - &                -& &     - &     -& &     - &                     - \\
        \textbf{GNNGuard (II)} &       &     - &     -& &     - &                -& &     - &     -& &     - &                     - \\

        \noalign{\vskip 0.25ex}
        \cdashline{1-1}[0.8pt/2pt]
        \cdashline{3-7}[0.8pt/2pt]
        \cdashline{9-13}[0.8pt/2pt]
        \noalign{\vskip 0.25ex}

        \textbf{GCN-SVD (\scaleroman{I}) $k=5\ \ $} &  &  64.44\tiny{$\pm$8.20} &  59.44\tiny{$\pm$3.93}& &  41.11\tiny{$\pm$7.97} &             31.67\tiny{$\pm$3.60}& &  68.18\tiny{$\pm$1.13} &  67.54\tiny{$\pm$0.97}& &  52.91\tiny{$\pm$0.28} &                  27.40\tiny{$\pm$0.29} \\
        \textbf{GCN-SVD (II) $k=5\ \ $} &  &  46.67\tiny{$\pm$4.08} &  32.22\tiny{$\pm$4.37}& &  45.56\tiny{$\pm$3.93} &             26.11\tiny{$\pm$6.14}& &  52.01\tiny{$\pm$2.45} &  30.69\tiny{$\pm$1.01}& &  52.32\tiny{$\pm$0.10} &                  25.87\tiny{$\pm$0.27} \\

        \noalign{\vskip 0.25ex}
        \cdashline{1-1}[0.8pt/2pt]
        \cdashline{3-7}[0.8pt/2pt]
        \cdashline{9-13}[0.8pt/2pt]
        \noalign{\vskip 0.25ex}

        \textbf{GCN-SVD (\scaleroman{I}) $k=10$} & &  65.56\tiny{$\pm$7.49} &  57.22\tiny{$\pm$3.42}& &  36.11\tiny{$\pm$1.57} &             31.11\tiny{$\pm$0.79}& &  68.36\tiny{$\pm$1.33} &  67.85\tiny{$\pm$0.72}& &  54.51\tiny{$\pm$0.68} &                  27.30\tiny{$\pm$0.30} \\
        \textbf{GCN-SVD (II) $k=10$} & &  57.22\tiny{$\pm$6.14} &  37.78\tiny{$\pm$3.93}& &  36.67\tiny{$\pm$3.60} &             30.56\tiny{$\pm$8.85}& &  58.70\tiny{$\pm$3.00} &  45.62\tiny{$\pm$2.52}& &  52.58\tiny{$\pm$0.20} &                  24.60\tiny{$\pm$0.26} \\

        \noalign{\vskip 0.25ex}
        \cdashline{1-1}[0.8pt/2pt]
        \cdashline{3-7}[0.8pt/2pt]
        \cdashline{9-13}[0.8pt/2pt]
        \noalign{\vskip 0.25ex}

        \textbf{GCN-SVD (\scaleroman{I}) $k=15$} & &  67.22\tiny{$\pm$6.14} &  54.44\tiny{$\pm$6.14}& &  24.44\tiny{$\pm$5.50} &             30.00\tiny{$\pm$1.36}& &  69.32\tiny{$\pm$1.21} &  65.26\tiny{$\pm$0.97}& &  57.79\tiny{$\pm$0.38} &                  28.06\tiny{$\pm$0.23} \\
        \textbf{GCN-SVD (\scaleroman{II}) $k=15$} & &  65.56\tiny{$\pm$6.14} &  38.33\tiny{$\pm$5.93}& &  23.89\tiny{$\pm$6.98} &             27.22\tiny{$\pm$7.97}& &  64.02\tiny{$\pm$1.30} &  54.09\tiny{$\pm$2.25}& &  53.81\tiny{$\pm$0.35} &                  25.29\tiny{$\pm$0.41} \\

       \noalign{\vskip 0.25ex}
        \cdashline{1-1}[0.8pt/2pt]
        \cdashline{3-7}[0.8pt/2pt]
        \cdashline{9-13}[0.8pt/2pt]
        \noalign{\vskip 0.25ex}

        \textbf{GCN-SVD (\scaleroman{I}) $k=50$} & &  65.00\tiny{$\pm$6.24} &  50.56\tiny{$\pm$6.43}& &  18.33\tiny{$\pm$2.36} &             25.56\tiny{$\pm$1.57}& &  75.30\tiny{$\pm$0.62} &  64.49\tiny{$\pm$1.58}& &  63.53\tiny{$\pm$0.26} &                  27.74\tiny{$\pm$0.61} \\
        \textbf{GCN-SVD (II) $k=50$} & \multirow{-10}{*}{\rotatebox[origin=r]{90}{\textsc{Nettack}}} &  60.00\tiny{$\pm$6.24} &  47.78\tiny{$\pm$4.37}& &  25.00\tiny{$\pm$4.71} &             30.56\tiny{$\pm$9.56}& \multirow{-10}{*}{\rotatebox[origin=r]{90}{Metattack}} &  73.21\tiny{$\pm$1.68} &  59.34\tiny{$\pm$3.42}& &  56.95\tiny{$\pm$0.33} &                  25.80\tiny{$\pm$0.67} \\        
    \bottomrule
    \end{tabular}}
\end{table}

For \textbf{GNNGuard}, the ambiguity comes from different interpretations of Eq. (4) in the original paper~\citep{zhang2020gnnguard}: we consider the authors' original implementation as variant (I), and the model described in the original paper as variant (II), which we implement by building on the authors' implementation. 

Table~\ref{table:fixation-clean} shows that the differences in accuracy between the two variants are in most cases less than 2\%, while in many cases variant (II) shows better accuracy compared to variant (I), especially in experiments against Metattack. 
Thus, we use variant (II) as the default implementation for the empirical evaluations in \S\ref{sec:exp}.

For \textbf{GCN-SVD}, the ambiguity comes from the order of applying the preprocessing and low-rank approximation for the adjacency matrix $\matA$, which is not discussed in the original paper~\citep{entezari2020all}. 
\begin{itemize*}
    \item \textbf{Variant (I)}: 
    Since the original authors' implementation is not publicly available, we consider the implementation provided in DeepRobust~\citep{li2020deeprobust} as variant (I): it first calculates the rank-$k$ approximation $\tilde{\matA}$ of $\matA$, and then generates the preprocessed adjacency matrix 
$\hat{\matA}_\mathrm{s} = \tilde{\matD}_\mathrm{s}^{-1/2}(\tilde{\matA} + \matI)\tilde{\matD}_\mathrm{s}^{-1/2} 
= \tilde{\matD}_\mathrm{s}^{-1/2}\tilde{\matA}_\mathrm{s}\tilde{\matD}_\mathrm{s}^{-1/2}$,
which is then processed by a 
GCN~\citep{kipf2016semi}. %
However, as the identity matrix $\matI$ is added into $\tilde{\matA}$ after the low-rank approximation, the diagonal elements of the resulting $\hat{\matA}_\mathrm{s}$ matrix (i.e., the weights for the self-loop edges in the graph) can become significantly larger than the off-diagonal elements, especially when the rank $k$ is low.
As a result, this order of applying the  preprocessing and low-rank approximation  inadvertently adopts Design 1 which we identified; we have shown in Theorem~\ref{thm:2} that even merely increasing the weights $\alpha$ for the ego-embedding in the linear combination $\texttt{ENC}$ in Eq. \eqref{eq:design1} can lead to reduced attack loss $\attackLoss$ under structural perturbations. 
    \item \textbf{Variant (II)}: 
In variant (II), we consider the opposite order where we first add the identity matrix $\matI$ (self-loops) into the original adjacency matrix $\matA$, then we perform the low-rank approximation, and finally we symmetrically normalize the low-rank matrix $\tilde{\matA}_\mathrm{s}$ to generate the preprocessed $\hat{\matA}_\mathrm{s}$ used by a GCN model. This order allows the diagonal elements to be more on par in magnitude with the off-diagonal elements. As an example, on Citeseer, when using variant (I) with rank $k=5$, the average magnitude of the diagonal elements of the resulting $\hat{\matA}_\mathrm{s}$ can be 22.3 times the average magnitude of the off-diagonal elements; when using variant (II) instead, the average magnitude of the diagonal elements is only 9.0 times that of the off-diagonal elements.
\end{itemize*}

In Table~\ref{table:fixation-clean}, we report the performance of the two variants of GCN-SVD under the  experimental settings considered in \S\ref{sec:exp} with rank $k \in \{5, 10, 15, 50\}$: Variant (I), with our first design implicitly built-in, has in most cases significantly higher performance than variant (II), especially on homophilous datasets and when rank $k$ is low. 
These results further demonstrate the effectiveness of Design 1 that we identified.  
To enable a clear perspective of the performance and robustness improvement brought by Design 1, in our empirical analysis in \S\ref{sec:exp}, on top of the low-rank approximation vaccination we adopt variant (II) as the default implementation.

\paragraph{More Details on Randomized Smoothing Setup} Following \citet{bojchevski2020efficient}, we similarly set the significance level $\alpha=0.01$ (i.e., the certificates hold with probability $1 - \alpha = 0.99$), using $10^3$ samples to estimate 
the predictions of the smoothed classifier $f(\phi(\mathbf{s}))$ for input $\mathbf{s}$, and another $10^6$ samples to obtain multi-class certificates.
For the randomization scheme $\phi$, we only consider structural perturbations where with probability $p_+$ an new edge is added, and with probability $p_-$ an existing edge is removed. 
We consider multiple sets of $(p_+, p_-)$ in our experiments for a finer-grained evaluation: (1) $p_+ = 0.001, p_- = 0.4$, where both addition and deletion are allowed; (2) $p_+ = 0.001, p_- = 0$, where only addition is allowed; and (3) $p_+ = 0, p_- = 0.4$, where only deletion is allowed.

\subsection{Combining Heterophilous Design with Explicit Robustness-Enhancing Mechanisms}
\label{app:heterophily-svd}
In this section, we provide more details on how we incorporate the low-rank approximation vaccination into the formulations of \method~\citep{zhu2020beyond} and GraphSAGE~\citep{hamilton2017inductive} in order to form the hybrid methods, H$_2$GCN-SVD and GraphSAGE-SVD. 

\paragraph{H$_2$GCN-SVD and \method-SMGDC} From \citep{zhu2020beyond}, each layer in the neighborhood aggregation stage of \method is algebraically formulated as 
\begin{equation}
\mathbf{R}^{(k)} = 
    \texttt{CONCAT}\left(
        \hat{\matA}_2\mathbf{R}^{(k-1)}, \hat{\matA}\mathbf{R}^{(k-1)}, \mathbf{R}^{(k-1)}
    \right),
\label{eq:app-h2gcn-formulation}
\end{equation}
where $\hat{\matA}=\matD^{-1/2}\matA\matD^{-1/2}$ is the symmetrically normalized  adjacency matrix without self-loops; 
$\hat{\matA}_2=\matD_2^{-1/2}\matA_2\matD_2^{-1/2}$ is the symmetrically normalized 2-hop graph adjacency matrix $\matA_2\in \{0, 1\}^{|\vertexSet|\times |\vertexSet|}$, with $[\matA_2]_{u,v} = 1$ if $v$ is in the 2-hop neighborhood $N_2(u)$ of node $u$;
$\mathbf{R}^{(k)}$ are the node representations after the $k$-th layer, 
and $\texttt{CONCAT}$ is the column-wise concatenation function. 
For H$_2$GCN-SVD, we replace $\hat{\matA}$ and $\hat{\matA}_2$ in Eq. \eqref{eq:app-h2gcn-formulation} respectively with the corresponding low-rank approximated versions $\tilde{\matA}$ and $\tilde{\matA}_2$, which are both postprocessed to be symmetrically normalized;
For \method-SMGDC, we replace $\hat{\matA}$ and $\hat{\matA}_2$ respectively with the corresponding GDC-preprocessed versions, and the aggregations $\hat{\matA}_2\mathbf{R}^{(k-1)}$ and $\hat{\matA}\mathbf{R}^{(k-1)}$ in the 1-hop and 2-hop neighborhoods are calculated with the Soft Medoid aggregators instead.
We give links to the GDC~\cite{klicpera_diffusion_2019} and Soft Medoid~\cite{geisler2020reliable} implementations that we used in our empirical evaluations in \S\ref{app:exp-details-setups}. 

\paragraph{GraphSAGE-SVD and GraphSAGE-SMGDC} From \citep{hamilton2017inductive}, each layer in GraphSAGE can be algebraically formulated as 
\begin{equation}
\mathbf{R}^{(k)} = \sigma\left(
    \texttt{CONCAT}\left(
        \bar{\matA}\mathbf{R}^{(k-1)}, \mathbf{R}^{(k-1)}
    \right) \cdot \matW^{(k-1)}
\right),
\label{eq:app-graphsage-formulation}
\end{equation}
where $\bar{\matA}$ is the row-stochastic graph adjacency matrix without self-loops; $\mathbf{R}^{(k)}$ are the node representations after the $k$-th layer; $\texttt{CONCAT}$ is the column-wise concatenation function; $\matW^{(k)}$ is the learnable weight matrix for the $k$-th layer, and $\sigma$ is the non-linear activation function (ReLU). 
For GraphSAGE-SVD, we replace $\bar{\matA}$ in Eq. \eqref{eq:app-graphsage-formulation} with the low-rank approximation of the adjacency matrix 
$\tilde{\matA}$, 
postprocessed by row-stochastic normalization;
For GraphSAGE-SMGDC, similar to \method-SMGDC, we replace $\bar{\matA}$ with the corresponding GDC-preprocessed versions, and the aggregation $\bar{\matA}\mathbf{R}^{(k-1)}$ is derived with the Soft Medoid aggregator. 
Note that we do not enable the neighborhood sampling function for the GraphSAGE and GraphSAGE-SVD models tested in this work, as noted in Appendix \ref{app:exp-details-params}.

\subsection{Hyperparameters} \label{app:exp-details-params}
\begin{itemize}
    \item \textbf{H$_2$GCN-SVD}
    \begin{multicols}{2}
    Initialization Parameters:
    \begin{itemize}
        \item \texttt{adj\_svd\_rank}: \newline best \texttt{k} chosen from \{5, 50\} for each dataset
        \item[\vspace{\fill}]
    \end{itemize}
    \columnbreak
    Training Parameters:
    \begin{itemize}
        \item \texttt{early\_stopping}: Yes
        \item \texttt{train\_iters}: 200
        \item \texttt{patience}: 100
    \end{itemize}
    \end{multicols}
    
    \item \textbf{GraphSAGE-SVD}
    \begin{multicols}{2}
    Initialization Parameters:
    \begin{itemize}
        \item \texttt{adj\_nhood}: \texttt{['1']}
        \item \texttt{network\_setup}: \newline \texttt{ I-T1-G-V-C1-M64-R-T2-G-V-C2-MO-R}
        \item \texttt{adj\_norm\_type}: \texttt{rw}
        \item \texttt{adj\_svd\_rank}: \newline best \texttt{k} chosen from \{5, 50\} for each dataset
    \end{itemize}
    \columnbreak
    Training Parameters:
    \begin{itemize}
        \item \texttt{early\_stopping}: Yes
        \item \texttt{train\_iters}: 200
        \item \texttt{patience}: 100
        \item[\vspace{\fill}]
        \item[\vspace{\fill}]
        \item[\vspace{\fill}]
    \end{itemize}
    \end{multicols}

    \item \textbf{\textbf{H$_2$GCN-SMGDC}}
    \begin{multicols}{2}
    Initialization Parameters:
    \begin{itemize}
        \item \texttt{network\_setup}: \newline \texttt{ M64-R-T1-GS-V-T2-GS-V-C1-C2-D0.5-MO}
        \item \texttt{adj\_norm\_type}: \texttt{gdc}
    \end{itemize}

    \columnbreak
    Training Parameters:
    \begin{itemize}
        \item \texttt{early\_stopping}: Yes
        \item \texttt{train\_iters}: 200
        \item \texttt{patience}: 100
    \end{itemize}
    \end{multicols}

    \item \textbf{\textbf{GraphSAGE-SMGDC}}
    \begin{multicols}{2}
    Initialization Parameters:
    \begin{itemize}
        \item \texttt{adj\_nhood}: \texttt{['1']}
        \item \texttt{network\_setup}: \newline \texttt{I-T1-GS-V-C1-M64-R-T2-GS-V-C2-MO-R}
        \item \texttt{adj\_norm\_type}: \texttt{gdc}
    \end{itemize}

    \columnbreak
    Training Parameters:
    \begin{itemize}
        \item \texttt{early\_stopping}: Yes
        \item \texttt{train\_iters}: 200
        \item \texttt{patience}: 100
    \end{itemize}
    \end{multicols}

    \newpage
    \item \textbf{\textbf{H$_2$GCN}}
    \begin{multicols}{2}
    Initialization Parameters:\\
    (default parameters)

    \columnbreak
    Training Parameters:
    \begin{itemize}
        \item \texttt{early\_stopping}: Yes
        \item \texttt{train\_iters}: 200
        \item \texttt{patience}: 100
        \item \texttt{lr}: 0.01
    \end{itemize}
    \end{multicols}

    \item \textbf{GraphSAGE}
    \begin{multicols}{2}
    Initialization Parameters:
    \begin{itemize}
        \item \texttt{adj\_nhood}: \texttt{['1']}
        \item \texttt{network\_setup}: \newline \texttt{\small I-T1-G-V-C1-M64-R-T2-G-V-C2-MO-R}
        \item \texttt{adj\_norm\_type}: \texttt{rw}
    \end{itemize}
    \columnbreak
    Training Parameters:
    \begin{itemize}
        \item \texttt{early\_stopping}: Yes
        \item \texttt{train\_iters}: 200
        \item \texttt{patience}: 100
        \item \texttt{lr}: 0.01
    \end{itemize}
    \end{multicols}

    \item \textbf{CPGNN}
    \begin{multicols}{2}
    Initialization Parameters:
    \begin{itemize}
        \item \texttt{network\_setup}: \newline \texttt{\small M64-R-MO-E-BP2}
        \item[\vspace{\fill}]
        \item[\vspace{\fill}]
    \end{itemize}
    \columnbreak
    Training Parameters:
    \begin{itemize}
        \item \texttt{early\_stopping}: Yes
        \item \texttt{train\_iters}: 400
        \item \texttt{patience}: 100
        \item \texttt{lr}: 0.01
    \end{itemize}
    \end{multicols}

    \item \textbf{GPR-GNN}
    \begin{multicols}{2}
    Initialization Parameters:
    \begin{itemize}
        \item \texttt{nhid}: 64
        \item \texttt{alpha}: 0.9, which is chosen from the best\\ $\alpha \in \{0.1, 0.2, 0.5, 0.9\}$ on all datasets
    \end{itemize}
    \columnbreak
    Training Parameters:
    \begin{itemize}
        \item \texttt{train\_iters}: 200
        \item \texttt{lr}: 0.01
        \item[\vspace{\fill}]
    \end{itemize}
    \end{multicols}
    
    \item \textbf{FAGCN}
    \begin{multicols}{2}
    Initialization Parameters:
    \begin{itemize}
        \item \texttt{nhid}: 64
        \item \texttt{alpha}: 0.9, which is chosen from the best\\ $\alpha \in \{0.1, 0.2, 0.5, 0.9\}$ on all datasets
        \item \texttt{dropout}: 0.5
    \end{itemize}
    \columnbreak
    Training Parameters:
    \begin{itemize}
        \item \texttt{early\_stopping}: Yes
        \item \texttt{lr}: 0.01
        \item[\vspace{\fill}]
        \item[\vspace{\fill}]
    \end{itemize}
    \end{multicols}
    
    \item \textbf{APPNP}
    \begin{multicols}{2}
    Initialization Parameters:
    \begin{itemize}
        \item \texttt{nhid}: 64
        \item \texttt{alpha}: 0.9, which is chosen from the best\\ $\alpha \in \{0.1, 0.2, 0.9\}$ on all datasets
        \item \texttt{dropout}: 0.5
    \end{itemize}
    \columnbreak
    Training Parameters:
    \begin{itemize}
        \item \texttt{train\_iters}: 200
        \item \texttt{lr}: 0.01
    \end{itemize}
    \end{multicols}

    \item \textbf{GNNGuard}
    \begin{multicols}{2}
    Initialization Parameters:
    \begin{itemize}
        \item \texttt{nhid}: 64
        \item \texttt{dropout}: 0.5
        \item \texttt{base\_model}: \texttt{GCN} for variant (I);\newline \texttt{GCN-fixed} for variant (II) (default).
    \end{itemize}
    \columnbreak
    Training Parameters:
    \begin{itemize}
        \item \texttt{train\_iters}: 81
        \item \texttt{lr}: 0.01
        \item[\vspace{\fill}]
        \item[\vspace{\fill}]
    \end{itemize}
    \end{multicols}
    
    \newpage
    \item \textbf{ProGNN}
    \begin{multicols}{2}
    Initialization Parameters:
    \begin{itemize}
        \item \texttt{nhid}: 64
        \item \texttt{dropout}: 0.5
        \item[\vspace{\fill}]
        \item[\vspace{\fill}]
        \item[\vspace{\fill}]
        \item[\vspace{\fill}]
        \item[\vspace{\fill}]
        \item[\vspace{\fill}]
        \item[\vspace{\fill}]
        \item[\vspace{\fill}]
        \item[\vspace{\fill}]
    \end{itemize}
    \columnbreak
    Training Parameters:
    \begin{itemize}
        \item \texttt{epochs}: 400
        \item \texttt{lr}: 0.01
        \item \texttt{lr\_adj}: 0.01
        \item \texttt{weight\_decay}: 5e-4
        \item \texttt{alpha}: 5e-4
        \item \texttt{beta}: 1.5
        \item \texttt{gamma}: 1
        \item \texttt{lambda\_}: 0
        \item \texttt{phi}: 0
        \item \texttt{outer\_steps}: 1
        \item \texttt{innter\_steps}: 2
    \end{itemize}
    \end{multicols}
    
    \item \textbf{GCN-SVD}
    \begin{multicols}{2}
    Initialization Parameters:
    \begin{itemize}
        \item \texttt{nhid}: 64
        \item \texttt{k}: best \texttt{k} chosen from \{5, 10, 15, 50\} for \\ each dataset
        \item \texttt{dropout}: 0.5
        \item \texttt{svd\_solver}: \newline \texttt{eye-svd} (for variant (II) only)
    \end{itemize}
    \columnbreak
    Training Parameters:
    \begin{itemize}
        \item \texttt{train\_iters}: 200
        \item \texttt{weight\_decay}: 5e-4
        \item \texttt{lr}: 0.01
        \item[\vspace{\fill}]
        \item[\vspace{\fill}]
        \item[\vspace{\fill}]
    \end{itemize}
    \end{multicols}
    
    \item \textbf{GCN-SMGDC}
    \begin{multicols}{2}
    Initialization Parameters:
    \begin{itemize}
        \item \texttt{adj\_nhood}: \texttt{['0,1']}
        \item \texttt{network\_setup}: \newline \texttt{M64-GS-V-R-D0.5-MO-GS-V}
        \item \texttt{adj\_norm\_type}: \texttt{gdc}
    \end{itemize}

    \columnbreak
    Training Parameters:
    \begin{itemize}
        \item \texttt{early\_stopping}: Yes
        \item \texttt{train\_iters}: 200
        \item \texttt{patience}: 100
    \end{itemize}
    \end{multicols}

    \item \textbf{GCN} %
    \begin{multicols}{2}
    Initialization Parameters \\
    (in \texttt{class MultiLayerGCN}):
    \begin{itemize}
        \item \texttt{nhid}: 64
        \item \texttt{nlayer}: 2
    \end{itemize}
    \columnbreak
    Training Parameters:
    \begin{itemize}
        \item \texttt{train\_iters}: 200
        \item \texttt{lr}: 0.01
        \item \texttt{weight\_decay}: 5e-4
    \end{itemize}
    \end{multicols}
    
    \item \textbf{GAT}
    \begin{multicols}{2}
    Initialization Parameters
    \begin{itemize}
        \item \texttt{nhid}: 8
        \item \texttt{heads}: 8
        \item \texttt{dropout}: 0.5
        \item[\vspace{\fill}]
        \item[\vspace{\fill}]
    \end{itemize}
    \columnbreak
    Training Parameters:
    \begin{itemize}
        \item \texttt{early\_stopping}: Yes
        \item \texttt{train\_iters}: 1000
        \item \texttt{patience}: 100
        \item \texttt{lr}: 0.01
        \item \texttt{weight\_decay}: 5e-4
    \end{itemize}
    \end{multicols}
    
    \item \textbf{\textbf{MLP}}
    \begin{multicols}{2}
    Initialization Parameters:\\
    (in \texttt{class H2GCN}):
    \begin{itemize}
        \item \texttt{network\_setup}: \newline \texttt{M64-R-D0.5-MO}
        \item[\vspace{\fill}]
        \item[\vspace{\fill}]
    \end{itemize}

    \columnbreak
    Training Parameters:
    \begin{itemize}
        \item \texttt{early\_stopping}: Yes
        \item \texttt{train\_iters}: 200
        \item \texttt{patience}: 100
        \item \texttt{lr}: 0.01
    \end{itemize}
    \end{multicols}
    
\end{itemize}

\newpage
\subsection{Datasets}
\label{appendix-experiment-datasets}
\paragraph{Dataset and Unidentifiability}
\begin{itemize}
    \item \textbf{Heterophilous Datasets:} FB100~\citep{FB100Source} is a set of 100  university friendship network snapshots from Facebook in 2005~\citep{lim2021new}, from which we use one network. Each node is labeled with the reported gender, and the features encode education and accommodation. 
    Data is sent to the original authors~\citep{lim2021new} in an anonymized form. Though the dataset contains limited demographic (categorical) information volunteered by users on their individual Facebook pages, we manually inspect the dataset and confirm that the anonymized dataset is not recoverable and thus not identifiable. Also, no offensive content is found within the data.

    Snap Patents~\citep{SnapSource1, snapnets} is a utility patent citation network. Node labels reflect the time the patent was granted, and the features are derived from the patent's metadata. The dataset is maintained by the National Bureau of Economic Research, and is freely available for download\footnote{\url{https://www.nber.org/research/data/us-patents}}. Neither personally identifiable information nor offensive content is identified when we manually inspect the dataset.
    
    \item \textbf{Homophilous Datasets:} Cora~\citep{cora_dataset,citeseer_dataset,namata2012query}, Citeseer~\citep{citeseer_dataset,namata2012query} and Pubmed~\cite{citeseer_dataset,namata2012query} are scientific publication citation networks, whose labels categorize the research field, and features indicate the absence or presence of the corresponding word from the dictionary. No personally identifiable information or offensive content is identified when we manually inspect both datasets.
\end{itemize}

\paragraph{Downsampling}
For better computational tractability, we sample a subset of the Snap Patents data using a snowball sampling approach~\citep{Leo1961Snowball}, where a random 20\% of the neighbors for each traversed node are kept. We provide the pseudocode for the downsampling process in Algorithm~\ref{pseudo-snowball}.

\begin{center}
\scalebox{1}{
\begin{minipage}{\linewidth}
\begin{algorithm}[H]
\DontPrintSemicolon 
\SetAlgoLined
\SetKwInput{KwInput}{Input}                %
\SetKwInput{KwOutput}{Output}  
\SetAlgoLined
\caption{Downsampling Algorithm For Snap Patents}
\label{pseudo-snowball}
\KwInput{Graph to sample $G$\newline
Number of nodes to sample $N$ \newline
Sampling ratio $p$}
\KwOutput{Downsampled graph $G'$}
initialization\;
\tcc{Initialize a queue bfs$_{\text{quene}}$ for Breadth First Search, \newline and a list nodes$_\text{sampled}$ for storing  sampled nodes}
bfs$_{\text{quene}}$ $\leftarrow$ \textsc{Queue}() \;
nodes$_\text{sampled}$ $\leftarrow$ \textsc{List}()\;
\tcc{Start BFS with a random node from the largest connected component in G;\newline
\textsc{Random}(array, n) returns n elements from an array with equal probability without replacement}
node$_\text{starting}$ $\leftarrow$  \textsc{Random}(\textsc{LargestConnectedComponent}($G$), 1) \;
push node$_\text{starting}$ into bfs$_{\text{quene}}$\;
\While{\textsc{Length}(nodes$_\text{sampled}$) < $N$}{
node $\leftarrow$ bfs$_{\text{quene}}$.pop() \;
neighbors $\leftarrow$ one hop neighbors of node\;
neighbors$_{\text{drawn}}$ $\leftarrow$ \textsc{Random}(neighbors, $p\times$ \textsc{Length}(neighbors)) \;
\For{neighbor $\in$ neighbors$_{\text{drawn}}$}{
\If{neighbor $\notin$ nodes$_\text{sampled}$}{
append neighbor to nodes$_\text{sampled}$ \;
push neighbor into bfs$_{\text{quene}}$\;
}
}
}
$G'$ $\leftarrow$ subgraph induced by nodes$_\text{sampled}$ \;
\Return{$G'$}\;
\end{algorithm}
\end{minipage}
}
\end{center}

\section{Detailed Experiment Results}
\label{app:real}

\subsection{Detailed Results for Evaluation on Empirical Robustness}
\label{app:real-benchmark-results}
\begin{table}[H]
    \centering
    \caption{
        Detailed classification accuracy (and standard deviation) of each method for the target nodes attacked by \textsc{Nettack}, calculated across different sets of perturbation. 
        Best GNN performance against attacks is highlighted in blue per dataset, and in gray per model group. OOM denotes experiment settings that run out of memory. MLP is immune to structural attacks and not considered as a GNN model. 
    }
    \label{table:real-results-detailed-netattack}
    \resizebox*{0.95\columnwidth}{!}{%
    \begin{tabular}{l ccc cc>{\color{dark-gray}}c c cc>{\color{dark-gray}}c c cc>{\color{dark-gray}}c c cc>{\color{dark-gray}}c c cc>{\color{dark-gray}}c}
        \toprule
        
        & \multirow{4}{*}{\rotatebox[origin=r]{90}{\textbf{Hete.}}} & \multirow{3}{*}{\rotatebox[origin=r]{90}{\textbf{Vaccin.}}} &
        
        & \multicolumn{3}{c}{\texttt{\bf Cora}} && \multicolumn{3}{c}{\texttt{\bf Pubmed}} && \multicolumn{3}{c}{\texttt{\bf Citeseer}}\  && \multicolumn{3}{c}{\texttt{\bf FB100}} && \multicolumn{3}{c}{\texttt{\bf Snap}} \\
        &&& & \multicolumn{3}{c}{$h$=0.804} && \multicolumn{3}{c}{$h$=0.802} && \multicolumn{3}{c}{$h$=0.736} && \multicolumn{3}{c}{$h$=0.531} && \multicolumn{3}{c}{$h$=0.134}  \\
        
        \cmidrule{5-7} \cmidrule{9-11}  \cmidrule{13-15} \cmidrule{17-19} \cmidrule{21-23}
        
        &&& & Poison & Evasion & Clean & & Poison & Evasion & Clean && Poison & Evasion & Clean && Poison & Evasion & Clean && Poison & Evasion & Clean\\
        
        \cmidrule{1-3} \cmidrule{5-7} \cmidrule{9-11}  \cmidrule{13-15} \cmidrule{17-19} \cmidrule{21-23}

        \textbf{H$_2$GCN-SVD}       & \checkmark & \checkmark & & $\underset{\scriptscriptstyle{\pm 2.72}}{70.00}$ &   $\cellcolor{blue!20}\underset{\scriptscriptstyle{\pm 3.42}}{70.56}$ &  $\underset{\scriptscriptstyle{\pm 3.42}}{74.44}$&&\cellcolor{blue!20}
        $\underset{\scriptscriptstyle{\pm 3.93}}{86.11}$&\cellcolor{blue!20}
        $\underset{\scriptscriptstyle{\pm 3.93}}{86.11}$&
        $\underset{\scriptscriptstyle{\pm 4.37}}{87.22}$  &&
        $\underset{\scriptscriptstyle{\pm 3.60}}{65.00}$ &     $\underset{\scriptscriptstyle{\pm 4.78}}{66.11}$ &  $\underset{\scriptscriptstyle{\pm 2.72}}{70.00}$& &   $\underset{\scriptscriptstyle{\pm 3.42}}{59.44}$ & \cellcolor{blue!20}$\underset{\scriptscriptstyle{\pm 2.72}}{60.00}$ &  $\underset{\scriptscriptstyle{\pm 2.36}}{61.67}$& &  \cellcolor{blue!20}$\underset{\scriptscriptstyle{\pm 3.42}}{28.89}$ &                     \cellcolor{gray!20}$\underset{\scriptscriptstyle{\pm 3.14}}{28.89}$ &                   $\underset{\scriptscriptstyle{\pm 2.08}}{30.56}$ \\
        
        \textbf{GraphSAGE-SVD}   & \checkmark &\checkmark & & \cellcolor{blue!20}$\underset{\scriptscriptstyle{\pm 2.36}}{71.67}$ & \cellcolor{blue!20}$\underset{\scriptscriptstyle{\pm 2.08}}{70.56}$ &  $\underset{\scriptscriptstyle{\pm 4.78}}{77.22}$& &$\underset{\scriptscriptstyle{\pm 4.16}}{81.11}$&
        $\underset{\scriptscriptstyle{\pm 3.42}}{81.11}$
        &$\underset{\scriptscriptstyle{\pm 2.08}}{84.44}$ &
        & \cellcolor{blue!20}$\underset{\scriptscriptstyle{\pm 3.42}}{67.78}$ & \cellcolor{blue!20}$\underset{\scriptscriptstyle{\pm 3.60}}{68.33}$ &  $\underset{\scriptscriptstyle{\pm 1.36}}{70.00}$& &   \cellcolor{blue!20}$\underset{\scriptscriptstyle{\pm 1.36}}{60.00}$ &  $\underset{\scriptscriptstyle{\pm 2.83}}{59.44}$ &  $\underset{\scriptscriptstyle{\pm 4.08}}{60.00}$& &                    $\underset{\scriptscriptstyle{\pm 6.80}}{26.67}$ &                     $\underset{\scriptscriptstyle{\pm 6.85}}{26.11}$ &                   $\underset{\scriptscriptstyle{\pm 5.50}}{27.22}$ \\
        \textbf{H$_2$GCN-SMGDC}     & \checkmark  & \checkmark & & $\underset{\scriptscriptstyle{\pm 4.37}}{59.44}$ &   $\underset{\scriptscriptstyle{\pm 6.24}}{60.00}$ &  $\underset{\scriptscriptstyle{\pm 4.78}}{77.22}$& &OOM &OOM &OOM &&  $\underset{\scriptscriptstyle{\pm 3.60}}{43.33}$ &     $\underset{\scriptscriptstyle{\pm 4.91}}{46.67}$ &  $\underset{\scriptscriptstyle{\pm 1.57}}{67.22}$& &   $\underset{\scriptscriptstyle{\pm 1.57}}{47.22}$ &  $\underset{\scriptscriptstyle{\pm 6.14}}{52.22}$ &  $\underset{\scriptscriptstyle{\pm 0.00}}{61.67}$& &                    $\underset{\scriptscriptstyle{\pm 1.57}}{22.22}$ &                     $\underset{\scriptscriptstyle{\pm 2.72}}{23.33}$ &                   $\underset{\scriptscriptstyle{\pm 0.79}}{30.56}$ \\
        
        \textbf{GraphSAGE-SMGDC} & \checkmark  & \checkmark & & $\underset{\scriptscriptstyle{\pm 8.28}}{56.67}$ &  $\underset{\scriptscriptstyle{\pm 10.30}}{57.78}$ &  $\underset{\scriptscriptstyle{\pm 5.44}}{78.33}$& &OOM &OOM &OOM &&   $\underset{\scriptscriptstyle{\pm 3.60}}{46.67}$ &     $\underset{\scriptscriptstyle{\pm 3.42}}{46.11}$ &  $\underset{\scriptscriptstyle{\pm 2.83}}{67.78}$& &   $\underset{\scriptscriptstyle{\pm 4.16}}{47.22}$ &  $\underset{\scriptscriptstyle{\pm 2.83}}{46.11}$ &  $\underset{\scriptscriptstyle{\pm 1.57}}{59.44}$& &                    $\underset{\scriptscriptstyle{\pm 3.14}}{20.56}$ &                     $\underset{\scriptscriptstyle{\pm 2.08}}{22.22}$ &                   $\underset{\scriptscriptstyle{\pm 4.16}}{29.44}$ \\
        
        \noalign{\vskip 0.25ex}
        \cdashline{1-3}[0.8pt/2pt]\cdashline{5-23}[0.8pt/2pt]
        \noalign{\vskip 0.25ex}
        
        \textbf{H$_2$GCN}           & \checkmark & & & $\underset{\scriptscriptstyle{\pm 5.50}}{38.89}$ &   $\underset{\scriptscriptstyle{\pm 3.42}}{45.56}$ &  $\underset{\scriptscriptstyle{\pm 8.31}}{82.78}$& &$\underset{\scriptscriptstyle{\pm 5.67}}{44.44}$&
        $\underset{\scriptscriptstyle{\pm 8.16}}{46.67}$&
        $\underset{\scriptscriptstyle{\pm 3.14}}{87.78}$&&   $\underset{\scriptscriptstyle{\pm 1.57}}{27.22}$ &     $\underset{\scriptscriptstyle{\pm 1.57}}{33.89}$ &  $\underset{\scriptscriptstyle{\pm 6.98}}{69.44}$& &   $\underset{\scriptscriptstyle{\pm 3.42}}{27.78}$ &  $\underset{\scriptscriptstyle{\pm 0.79}}{32.78}$ &  $\underset{\scriptscriptstyle{\pm 1.57}}{60.56}$& &                    $\underset{\scriptscriptstyle{\pm 2.83}}{12.78}$ &                     $\underset{\scriptscriptstyle{\pm 2.83}}{12.78}$ &                   $\underset{\scriptscriptstyle{\pm 2.72}}{30.00}$ \\
        
        \textbf{GraphSAGE}      & \checkmark & & & $\underset{\scriptscriptstyle{\pm 2.72}}{36.67}$ &   $\underset{\scriptscriptstyle{\pm 3.14}}{44.44}$ &  $\underset{\scriptscriptstyle{\pm 9.56}}{82.22}$& &$\underset{\scriptscriptstyle{\pm 8.92}}{33.33}$&$\underset{\scriptscriptstyle{\pm 9.06}}{34.44}$&$\underset{\scriptscriptstyle{\pm 3.93}}{84.44}$&&  $\underset{\scriptscriptstyle{\pm 10.89}}{31.67}$ &     $\underset{\scriptscriptstyle{\pm 8.92}}{35.00}$ &  $\underset{\scriptscriptstyle{\pm 6.85}}{70.56}$& &   $\underset{\scriptscriptstyle{\pm 3.42}}{33.89}$ & \cellcolor{gray!20}$\underset{\scriptscriptstyle{\pm 2.83}}{42.22}$ &  $\underset{\scriptscriptstyle{\pm 2.72}}{60.00}$& &                    $\underset{\scriptscriptstyle{\pm 7.07}}{16.67}$ &                     $\underset{\scriptscriptstyle{\pm 6.71}}{15.56}$ &                   $\underset{\scriptscriptstyle{\pm 4.16}}{24.44}$ \\
        
        \textbf{CPGNN}       & \checkmark & & & $\underset{\scriptscriptstyle{\pm 6.14}}{47.22}$ &   $\underset{\scriptscriptstyle{\pm 6.98}}{52.22}$ &  $\underset{\scriptscriptstyle{\pm 8.28}}{81.67}$&& $\underset{\scriptscriptstyle{\pm 7.20}}{60.00}$ & $\underset{\scriptscriptstyle{\pm 5.93}}{60.00}$ & $\underset{\scriptscriptstyle{\pm 5.67}}{82.78}$& &  $\underset{\scriptscriptstyle{\pm 9.65}}{40.56}$ &     $\underset{\scriptscriptstyle{\pm 6.80}}{46.67}$ &  $\underset{\scriptscriptstyle{\pm 1.36}}{73.33}$& & \cellcolor{gray!20}$\underset{\scriptscriptstyle{\pm 10.30}}{49.44}$ &  $\underset{\scriptscriptstyle{\pm 2.83}}{15.56}$ &  $\underset{\scriptscriptstyle{\pm 4.16}}{66.11}$& &                    $\underset{\scriptscriptstyle{\pm 2.72}}{21.67}$ &                     $\underset{\scriptscriptstyle{\pm 4.78}}{22.78}$ &                   $\underset{\scriptscriptstyle{\pm 5.50}}{28.89}$ \\

        \textbf{GPR-GNN}         & \checkmark & & & $\underset{\scriptscriptstyle{\pm 2.72}}{21.67}$ &   $\underset{\scriptscriptstyle{\pm 3.14}}{29.44}$ &  $\underset{\scriptscriptstyle{\pm 7.49}}{82.22}$& &$\underset{\scriptscriptstyle{\pm 4.78}}{13.89}$ &$\underset{\scriptscriptstyle{\pm 6.14}}{15.56}$&$\underset{\scriptscriptstyle{\pm 1.57}}{85.56}$&&   $\underset{\scriptscriptstyle{\pm 2.08}}{24.44}$ &     $\underset{\scriptscriptstyle{\pm 0.79}}{32.22}$ &  $\underset{\scriptscriptstyle{\pm 2.08}}{67.78}$& &    $\underset{\scriptscriptstyle{\pm 0.79}}{2.78}$ &   $\underset{\scriptscriptstyle{\pm 2.83}}{9.44}$ &  $\underset{\scriptscriptstyle{\pm 4.91}}{56.67}$& &                     $\underset{\scriptscriptstyle{\pm 2.08}}{4.44}$ &                      $\underset{\scriptscriptstyle{\pm 1.36}}{3.33}$ &                   $\underset{\scriptscriptstyle{\pm 3.42}}{27.78}$ \\
        
        \textbf{FAGCN}           & \checkmark & & & $\underset{\scriptscriptstyle{\pm 6.14}}{26.11}$ &   $\underset{\scriptscriptstyle{\pm 6.71}}{38.89}$ &  $\underset{\scriptscriptstyle{\pm 8.16}}{83.33}$& & $\underset{\scriptscriptstyle{\pm 11.00}}{27.78}$ &$\underset{\scriptscriptstyle{\pm 13.40}}{31.67}$&$\underset{\scriptscriptstyle{\pm 2.72}}{86.67}$&&   $\underset{\scriptscriptstyle{\pm 6.43}}{25.56}$ &     $\underset{\scriptscriptstyle{\pm 4.78}}{37.78}$ &  $\underset{\scriptscriptstyle{\pm 5.15}}{70.56}$& &    $\underset{\scriptscriptstyle{\pm 2.83}}{6.11}$ &  $\underset{\scriptscriptstyle{\pm 2.83}}{12.78}$ &  $\underset{\scriptscriptstyle{\pm 5.93}}{58.33}$& &                     $\underset{\scriptscriptstyle{\pm 3.60}}{8.33}$ &                     $\underset{\scriptscriptstyle{\pm 2.36}}{10.00}$ &                   $\underset{\scriptscriptstyle{\pm 0.79}}{29.44}$ \\
        
        \textbf{APPNP}           & \checkmark & & & \cellcolor{gray!20}$\underset{\scriptscriptstyle{\pm 3.60}}{58.33}$ & \cellcolor{gray!20}$\underset{\scriptscriptstyle{\pm 7.86}}{67.78}$ &  $\underset{\scriptscriptstyle{\pm 5.50}}{72.22}$& &\cellcolor{gray!20}$\underset{\scriptscriptstyle{\pm 2.83}}{79.44}$&\cellcolor{gray!20}$\underset{\scriptscriptstyle{\pm 2.72}}{81.67}$&$\underset{\scriptscriptstyle{\pm 2.36}}{86.67}$& & \cellcolor{gray!20}$\underset{\scriptscriptstyle{\pm 3.14}}{56.11}$ & \cellcolor{gray!20}$\underset{\scriptscriptstyle{\pm 4.16}}{65.56}$ &  $\underset{\scriptscriptstyle{\pm 4.71}}{68.33}$& &   $\underset{\scriptscriptstyle{\pm 2.36}}{36.67}$ &  $\underset{\scriptscriptstyle{\pm 4.78}}{49.44}$ &  $\underset{\scriptscriptstyle{\pm 3.93}}{58.89}$& & \cellcolor{gray!20}$\underset{\scriptscriptstyle{\pm 1.36}}{25.00}$ & \cellcolor{gray!20}$\underset{\scriptscriptstyle{\pm 3.42}}{25.56}$ &                   $\underset{\scriptscriptstyle{\pm 2.36}}{28.33}$ \\
        
        \noalign{\vskip 0.25ex}
        \cdashline{1-3}[0.8pt/2pt]\cdashline{5-23}[0.8pt/2pt]
        \noalign{\vskip 0.25ex}
        
        \textbf{GNNGuard}        & & \checkmark & & \cellcolor{gray!20}$\underset{\scriptscriptstyle{\pm 1.36}}{58.33}$ &      - &  $\underset{\scriptscriptstyle{\pm 6.29}}{77.22}$& &\cellcolor{gray!20}$\underset{\scriptscriptstyle{\pm 6.71}}{73.89}$&-&$\underset{\scriptscriptstyle{\pm 2.83}}{82.78}$& & \cellcolor{gray!20}$\underset{\scriptscriptstyle{\pm 3.14}}{59.44}$ &        - &  $\underset{\scriptscriptstyle{\pm 4.78}}{67.78}$& &    $\underset{\scriptscriptstyle{\pm 0.79}}{0.56}$ &     - &  $\underset{\scriptscriptstyle{\pm 2.08}}{67.22}$& &                     $\underset{\scriptscriptstyle{\pm 1.57}}{9.44}$ &                        - &                   $\underset{\scriptscriptstyle{\pm 3.60}}{28.33}$ \\
        
        \textbf{ProGNN}          & & \checkmark & & $\underset{\scriptscriptstyle{\pm 7.97}}{48.89}$ &      - &  $\underset{\scriptscriptstyle{\pm 3.42}}{79.44}$& &OOM &- &OOM &&   $\underset{\scriptscriptstyle{\pm 7.49}}{32.78}$ &        - &  $\underset{\scriptscriptstyle{\pm 4.78}}{67.22}$& &   $\underset{\scriptscriptstyle{\pm 4.78}}{33.89}$ &     - &  $\underset{\scriptscriptstyle{\pm 3.93}}{51.11}$& &                    $\underset{\scriptscriptstyle{\pm 9.26}}{17.78}$ &                        - &                   $\underset{\scriptscriptstyle{\pm 5.50}}{27.22}$ \\
        
        \textbf{GCN-SVD}         & & \checkmark & & $\underset{\scriptscriptstyle{\pm 4.91}}{53.33}$ & \cellcolor{gray!20}$\underset{\scriptscriptstyle{\pm 6.24}}{60.00}$ &  $\underset{\scriptscriptstyle{\pm 4.16}}{75.56}$& &OOM&OOM&OOM&&   $\underset{\scriptscriptstyle{\pm 2.08}}{28.89}$ & \cellcolor{gray!20}$\underset{\scriptscriptstyle{\pm 4.37}}{47.78}$ &  $\underset{\scriptscriptstyle{\pm 0.79}}{59.44}$& & \cellcolor{gray!20}$\underset{\scriptscriptstyle{\pm 2.36}}{41.67}$ & \cellcolor{gray!20}$\underset{\scriptscriptstyle{\pm 3.93}}{45.56}$ &  $\underset{\scriptscriptstyle{\pm 4.37}}{50.56}$& & \cellcolor{gray!20}$\underset{\scriptscriptstyle{\pm 5.44}}{25.00}$ & \cellcolor{blue!20}$\underset{\scriptscriptstyle{\pm 9.56}}{30.56}$ &                   $\underset{\scriptscriptstyle{\pm 6.71}}{27.78}$ \\
        
        \textbf{GCN-SMGDC}       & &\checkmark & & $\underset{\scriptscriptstyle{\pm 4.91}}{40.00}$ &   $\underset{\scriptscriptstyle{\pm 6.14}}{40.56}$ &  $\underset{\scriptscriptstyle{\pm 3.93}}{77.78}$& &OOM &OOM &OOM &&   $\underset{\scriptscriptstyle{\pm 2.83}}{33.89}$ &     $\underset{\scriptscriptstyle{\pm 4.08}}{35.00}$ &  $\underset{\scriptscriptstyle{\pm 0.79}}{62.22}$& &   $\underset{\scriptscriptstyle{\pm 4.08}}{16.67}$ &  $\underset{\scriptscriptstyle{\pm 3.42}}{36.11}$ &  $\underset{\scriptscriptstyle{\pm 5.67}}{51.11}$& &                    $\underset{\scriptscriptstyle{\pm 5.15}}{20.56}$ &                     $\underset{\scriptscriptstyle{\pm 6.85}}{22.22}$ &                   $\underset{\scriptscriptstyle{\pm 2.36}}{28.33}$ \\
        
        \noalign{\vskip 0.25ex}
        \cdashline{1-3}[0.8pt/2pt]\cdashline{5-23}[0.8pt/2pt]
        \noalign{\vskip 0.25ex}
        
        \textbf{GAT}           &&&  &  $\underset{\scriptscriptstyle{\pm 0.79}}{13.89}$ &   $\underset{\scriptscriptstyle{\pm 4.16}}{12.22}$ &  $\underset{\scriptscriptstyle{\pm 3.42}}{84.44}$& &\cellcolor{gray!20}$\underset{\scriptscriptstyle{\pm 4.16}}{7.22}$&\cellcolor{gray!20}$\underset{\scriptscriptstyle{\pm 4.08}}{6.67}$&$\underset{\scriptscriptstyle{\pm 1.36}}{83.33}$&&    $\underset{\scriptscriptstyle{\pm 3.42}}{8.89}$ &     $\underset{\scriptscriptstyle{\pm 5.44}}{23.33}$ &  $\underset{\scriptscriptstyle{\pm 7.20}}{70.00}$& &  \cellcolor{gray!20}$\underset{\scriptscriptstyle{\pm 0.79}}{0.56}$ & \cellcolor{gray!20}$\underset{\scriptscriptstyle{\pm 1.36}}{1.67}$ &  $\underset{\scriptscriptstyle{\pm 0.79}}{60.56}$& & \cellcolor{gray!20}$\underset{\scriptscriptstyle{\pm 4.37}}{3.89}$ &                      \cellcolor{gray!20}$\underset{\scriptscriptstyle{\pm 3.14}}{2.22}$ &                   $\underset{\scriptscriptstyle{\pm 2.83}}{30.56}$ \\
        \textbf{GCN}          &&&   & \cellcolor{gray!20}$\underset{\scriptscriptstyle{\pm 3.60}}{18.33}$ & \cellcolor{gray!20}$\underset{\scriptscriptstyle{\pm 1.57}}{23.89}$ &  $\underset{\scriptscriptstyle{\pm 5.50}}{82.78}$& &$\underset{\scriptscriptstyle{\pm 0.79}}{5.56}$&$\underset{\scriptscriptstyle{\pm 0.79}}{5.56}$&$\underset{\scriptscriptstyle{\pm 2.72}}{85.00}$& & \cellcolor{gray!20}$\underset{\scriptscriptstyle{\pm 5.50}}{20.56}$ & \cellcolor{gray!20}$\underset{\scriptscriptstyle{\pm 4.08}}{26.67}$ &  $\underset{\scriptscriptstyle{\pm 8.20}}{72.78}$& &    $\underset{\scriptscriptstyle{\pm 0.00}}{0.00}$ &   $\underset{\scriptscriptstyle{\pm 0.00}}{0.00}$ &  $\underset{\scriptscriptstyle{\pm 7.97}}{56.11}$& &                     $\underset{\scriptscriptstyle{\pm 3.14}}{2.22}$ &                      \cellcolor{gray!20}$\underset{\scriptscriptstyle{\pm 3.14}}{2.22}$ &                   $\underset{\scriptscriptstyle{\pm 2.08}}{30.56}$ \\
        
        \noalign{\vskip 0.25ex}
        \cdashline{1-3}[0.8pt/2pt]\cdashline{5-23}[0.8pt/2pt]
        \noalign{\vskip 0.25ex}
        
        \textbf{MLP*}        &&&     &  $\underset{\scriptscriptstyle{\pm 3.42}}{64.44}$ &   $\underset{\scriptscriptstyle{\pm 3.42}}{64.44}$ &  $\underset{\scriptscriptstyle{\pm 3.42}}{64.44}$& &$\underset{\scriptscriptstyle{\pm 4.37}}{86.11}$&$\underset{\scriptscriptstyle{\pm 4.37}}{86.11}$&$\underset{\scriptscriptstyle{\pm 4.37}}{86.11}$&&   $\underset{\scriptscriptstyle{\pm 3.42}}{70.56}$ &     $\underset{\scriptscriptstyle{\pm 3.42}}{70.56}$ &  $\underset{\scriptscriptstyle{\pm 3.42}}{70.56}$& &   $\underset{\scriptscriptstyle{\pm 2.83}}{57.78}$ &  $\underset{\scriptscriptstyle{\pm 2.83}}{57.78}$ &  $\underset{\scriptscriptstyle{\pm 2.83}}{57.78}$& &                    $\underset{\scriptscriptstyle{\pm 2.72}}{30.00}$ &                     $\underset{\scriptscriptstyle{\pm 2.72}}{30.00}$ &                   $\underset{\scriptscriptstyle{\pm 2.72}}{30.00}$ \\
        
        \bottomrule
    \end{tabular}
    }
    \vspace{-0.5cm}
\end{table}

\begin{table}[H]
    \centering
    \caption{
        Detailed classification accuracy (and standard deviation) for the unlabeled nodes of each method attacked by Metattack with budget as 20\% of the total number of edges of each graph, calculated across different sets of perturbation. Best GNN performance against attacks is highlighted in blue per dataset, and in gray per model group. Metattack runs out of memory on Pubmed; MLP is immune to structural attacks and not considered as a GNN model. 
    }
    \label{table:real-results-detailed-metattack-20p}
    \resizebox*{0.95\columnwidth}{!}{%
        \begin{tabular}{l ccc cc>{\color{dark-gray}}c c cc>{\color{dark-gray}}c c cc>{\color{dark-gray}}c c cc>{\color{dark-gray}}c}
        \toprule
        
        & \multirow{4}{*}{\rotatebox[origin=r]{90}{\textbf{Hete.}}} & \multirow{3}{*}{\rotatebox[origin=r]{90}{\textbf{Vaccin.}}} &
        
        & \multicolumn{3}{c}{\texttt{\bf Cora}}  && \multicolumn{3}{c}{\texttt{\bf Citeseer}}\  && \multicolumn{3}{c}{\texttt{\bf FB100}} && \multicolumn{3}{c}{\texttt{\bf Snap}} \\
        &&& & \multicolumn{3}{c}{$h$=0.804} && \multicolumn{3}{c}{$h$=0.736} && \multicolumn{3}{c}{$h$=0.531} && \multicolumn{3}{c}{$h$=0.134}  \\
        
        \cmidrule{5-7} \cmidrule{9-11}  \cmidrule{13-15} \cmidrule{17-19}
        
        &&& & Poison & Evasion & Clean & & Poison & Evasion & Clean && Poison & Evasion & Clean && Poison & Evasion & Clean\\
        
        \cmidrule{1-3} \cmidrule{5-7} \cmidrule{9-11}  \cmidrule{13-15} \cmidrule{17-19}
        
        \textbf{H$_2$GCN-SVD}  & \checkmark & \checkmark   &  &    $\underset{\scriptscriptstyle{\pm 0.47}}{67.87}$ &     $\underset{\scriptscriptstyle{\pm 0.35}}{74.01}$ &  $\underset{\scriptscriptstyle{\pm 0.37}}{76.89}$& & \cellcolor{blue!20}$\underset{\scriptscriptstyle{\pm 0.46}}{70.42}$ &         $\underset{\scriptscriptstyle{\pm 1.92}}{71.54}$ &      $\underset{\scriptscriptstyle{\pm 1.03}}{73.42}$& & \cellcolor{gray!20}$\underset{\scriptscriptstyle{\pm 0.08}}{56.72}$ &      $\underset{\scriptscriptstyle{\pm 0.63}}{56.58}$ &   $\underset{\scriptscriptstyle{\pm 0.77}}{56.81}$& &                        $\underset{\scriptscriptstyle{\pm 0.14}}{25.60}$ &                         $\underset{\scriptscriptstyle{\pm 0.17}}{27.26}$ &                       $\underset{\scriptscriptstyle{\pm 0.26}}{27.63}$ \\
        \textbf{GraphSAGE-SVD} & \checkmark & \checkmark  & & \cellcolor{gray!20}$\underset{\scriptscriptstyle{\pm 1.32}}{68.86}$ &     $\underset{\scriptscriptstyle{\pm 0.40}}{74.31}$ &  $\underset{\scriptscriptstyle{\pm 0.29}}{77.52}$& &        $\underset{\scriptscriptstyle{\pm 0.52}}{69.10}$ &         $\underset{\scriptscriptstyle{\pm 0.90}}{70.22}$ &      $\underset{\scriptscriptstyle{\pm 0.17}}{72.16}$& &     $\underset{\scriptscriptstyle{\pm 0.33}}{55.76}$ & \cellcolor{gray!20}$\underset{\scriptscriptstyle{\pm 0.88}}{57.38}$ &   $\underset{\scriptscriptstyle{\pm 0.86}}{57.38}$& & \cellcolor{gray!20}$\underset{\scriptscriptstyle{\pm 0.30}}{26.58}$ &                         $\underset{\scriptscriptstyle{\pm 1.01}}{26.77}$ &                       $\underset{\scriptscriptstyle{\pm 0.70}}{26.72}$ \\
        \textbf{H$_2$GCN-SMGDC} & \checkmark & \checkmark   & &    $\underset{\scriptscriptstyle{\pm 1.65}}{66.50}$ & \cellcolor{gray!20}$\underset{\scriptscriptstyle{\pm 0.56}}{77.65}$ &  $\underset{\scriptscriptstyle{\pm 0.33}}{80.60}$& &        $\underset{\scriptscriptstyle{\pm 1.24}}{69.04}$ &         $\underset{\scriptscriptstyle{\pm 1.35}}{72.33}$ &      $\underset{\scriptscriptstyle{\pm 0.92}}{74.31}$& &     $\underset{\scriptscriptstyle{\pm 1.51}}{54.63}$ &      $\underset{\scriptscriptstyle{\pm 0.09}}{56.95}$ &   $\underset{\scriptscriptstyle{\pm 0.10}}{56.52}$& &                        $\underset{\scriptscriptstyle{\pm 1.09}}{24.41}$ & \cellcolor{gray!20}$\underset{\scriptscriptstyle{\pm 0.42}}{27.40}$ &                       $\underset{\scriptscriptstyle{\pm 0.62}}{27.50}$ \\
        \textbf{GraphSAGE-SMGDC} & \checkmark & \checkmark & &    $\underset{\scriptscriptstyle{\pm 2.07}}{66.95}$ &     $\underset{\scriptscriptstyle{\pm 0.16}}{75.99}$ &  $\underset{\scriptscriptstyle{\pm 0.26}}{79.39}$& &        $\underset{\scriptscriptstyle{\pm 0.97}}{68.68}$ & \cellcolor{gray!20}$\underset{\scriptscriptstyle{\pm 0.35}}{72.79}$ &      $\underset{\scriptscriptstyle{\pm 0.38}}{74.31}$& &     $\underset{\scriptscriptstyle{\pm 0.29}}{55.39}$ &      $\underset{\scriptscriptstyle{\pm 0.30}}{55.08}$ &   $\underset{\scriptscriptstyle{\pm 0.19}}{55.19}$& &                        $\underset{\scriptscriptstyle{\pm 0.76}}{25.21}$ &                         $\underset{\scriptscriptstyle{\pm 0.14}}{26.27}$ &                       $\underset{\scriptscriptstyle{\pm 0.29}}{26.38}$ \\
        
        \noalign{\vskip 0.25ex}
        \cdashline{1-3}[0.8pt/2pt]\cdashline{5-19}[0.8pt/2pt]
        \noalign{\vskip 0.25ex}
        
        \textbf{H$_2$GCN}  & \checkmark & &         &    $\underset{\scriptscriptstyle{\pm 6.61}}{57.75}$ & \cellcolor{blue!20}$\underset{\scriptscriptstyle{\pm 1.01}}{82.86}$ &  $\underset{\scriptscriptstyle{\pm 0.97}}{83.94}$& &        $\underset{\scriptscriptstyle{\pm 0.82}}{54.34}$ &         $\underset{\scriptscriptstyle{\pm 2.04}}{73.20}$ &      $\underset{\scriptscriptstyle{\pm 0.90}}{75.34}$& &     $\underset{\scriptscriptstyle{\pm 0.76}}{54.84}$ &      $\underset{\scriptscriptstyle{\pm 0.20}}{57.05}$ &   $\underset{\scriptscriptstyle{\pm 0.13}}{56.95}$& &                        $\underset{\scriptscriptstyle{\pm 0.59}}{25.34}$ &                         $\underset{\scriptscriptstyle{\pm 0.06}}{27.10}$ &                       $\underset{\scriptscriptstyle{\pm 0.05}}{27.49}$ \\
        \textbf{GraphSAGE}  & \checkmark & &     &    $\underset{\scriptscriptstyle{\pm 2.56}}{54.68}$ &     $\underset{\scriptscriptstyle{\pm 0.55}}{80.57}$ &  $\underset{\scriptscriptstyle{\pm 0.63}}{82.21}$& &        $\underset{\scriptscriptstyle{\pm 1.74}}{59.74}$ &         $\underset{\scriptscriptstyle{\pm 1.72}}{72.89}$ &      $\underset{\scriptscriptstyle{\pm 0.93}}{74.64}$& &     $\underset{\scriptscriptstyle{\pm 0.83}}{54.72}$ &      $\underset{\scriptscriptstyle{\pm 1.61}}{56.91}$ &   $\underset{\scriptscriptstyle{\pm 1.40}}{56.60}$& &                        $\underset{\scriptscriptstyle{\pm 0.76}}{24.14}$ &                         $\underset{\scriptscriptstyle{\pm 0.78}}{27.16}$ &                       $\underset{\scriptscriptstyle{\pm 0.84}}{27.18}$ \\
        \textbf{CPGNN} & \checkmark & &      & \cellcolor{blue!20}$\underset{\scriptscriptstyle{\pm 1.23}}{74.55}$ &     $\underset{\scriptscriptstyle{\pm 1.18}}{79.06}$ &  $\underset{\scriptscriptstyle{\pm 0.51}}{80.67}$& & \cellcolor{gray!20}$\underset{\scriptscriptstyle{\pm 1.93}}{68.07}$ & \cellcolor{blue!20}$\underset{\scriptscriptstyle{\pm 0.98}}{73.44}$ &      $\underset{\scriptscriptstyle{\pm 0.62}}{74.92}$& & \cellcolor{blue!20}$\underset{\scriptscriptstyle{\pm 1.50}}{61.58}$ &      $\underset{\scriptscriptstyle{\pm 7.20}}{60.19}$ &   $\underset{\scriptscriptstyle{\pm 7.09}}{60.17}$& & \cellcolor{blue!20}$\underset{\scriptscriptstyle{\pm 0.41}}{26.76}$ &                         $\underset{\scriptscriptstyle{\pm 0.75}}{27.02}$ &                       $\underset{\scriptscriptstyle{\pm 0.63}}{27.13}$ \\
        \textbf{GPR-GNN}  & \checkmark & &       &    $\underset{\scriptscriptstyle{\pm 5.23}}{48.29}$ &     $\underset{\scriptscriptstyle{\pm 1.67}}{80.80}$ &  $\underset{\scriptscriptstyle{\pm 1.75}}{81.84}$& &        $\underset{\scriptscriptstyle{\pm 2.77}}{35.25}$ &         $\underset{\scriptscriptstyle{\pm 0.42}}{69.77}$ &      $\underset{\scriptscriptstyle{\pm 0.46}}{70.71}$& &     $\underset{\scriptscriptstyle{\pm 0.60}}{59.94}$ & \cellcolor{blue!20}$\underset{\scriptscriptstyle{\pm 0.74}}{61.91}$ &   $\underset{\scriptscriptstyle{\pm 0.83}}{62.40}$& &                        $\underset{\scriptscriptstyle{\pm 1.29}}{21.06}$ &                         $\underset{\scriptscriptstyle{\pm 0.25}}{26.16}$ &                       $\underset{\scriptscriptstyle{\pm 0.31}}{26.08}$ \\
        \textbf{FAGCN}  & \checkmark & &         &    $\underset{\scriptscriptstyle{\pm 4.82}}{60.11}$ &     $\underset{\scriptscriptstyle{\pm 0.99}}{80.70}$ &  $\underset{\scriptscriptstyle{\pm 0.82}}{81.59}$& &        $\underset{\scriptscriptstyle{\pm 6.00}}{53.18}$ &         $\underset{\scriptscriptstyle{\pm 1.02}}{73.14}$ &      $\underset{\scriptscriptstyle{\pm 0.63}}{73.99}$& &     $\underset{\scriptscriptstyle{\pm 1.81}}{55.97}$ &      $\underset{\scriptscriptstyle{\pm 1.36}}{59.39}$ &   $\underset{\scriptscriptstyle{\pm 1.38}}{59.64}$& &                        $\underset{\scriptscriptstyle{\pm 0.62}}{24.04}$ &                         $\underset{\scriptscriptstyle{\pm 0.30}}{27.25}$ &                       $\underset{\scriptscriptstyle{\pm 0.23}}{27.15}$ \\
        \textbf{APPNP} & \checkmark & &          &    $\underset{\scriptscriptstyle{\pm 0.91}}{62.56}$ &     $\underset{\scriptscriptstyle{\pm 0.33}}{72.80}$ &  $\underset{\scriptscriptstyle{\pm 0.38}}{72.87}$& &        $\underset{\scriptscriptstyle{\pm 1.73}}{49.70}$ &         $\underset{\scriptscriptstyle{\pm 0.24}}{69.51}$ &      $\underset{\scriptscriptstyle{\pm 0.23}}{69.59}$& &     $\underset{\scriptscriptstyle{\pm 0.35}}{57.81}$ &      $\underset{\scriptscriptstyle{\pm 0.55}}{57.89}$ &   $\underset{\scriptscriptstyle{\pm 0.59}}{57.89}$& &                        $\underset{\scriptscriptstyle{\pm 0.29}}{27.76}$ & \cellcolor{blue!20}$\underset{\scriptscriptstyle{\pm 0.12}}{27.45}$ &                       $\underset{\scriptscriptstyle{\pm 0.11}}{27.41}$ \\
        
        \noalign{\vskip 0.25ex}
        \cdashline{1-3}[0.8pt/2pt]\cdashline{5-19}[0.8pt/2pt]
        \noalign{\vskip 0.25ex}
        
        \textbf{GNNGuard}  & &\checkmark &      & \cellcolor{gray!20}$\underset{\scriptscriptstyle{\pm 0.55}}{74.20}$ &        - &  $\underset{\scriptscriptstyle{\pm 0.55}}{80.15}$& & \cellcolor{gray!20}$\underset{\scriptscriptstyle{\pm 0.74}}{68.13}$ &            - &      $\underset{\scriptscriptstyle{\pm 0.28}}{72.61}$& & \cellcolor{gray!20}$\underset{\scriptscriptstyle{\pm 0.48}}{60.89}$ &         - &   $\underset{\scriptscriptstyle{\pm 0.60}}{65.66}$& &                        $\underset{\scriptscriptstyle{\pm 0.67}}{23.78}$ &                            - &                       $\underset{\scriptscriptstyle{\pm 0.98}}{26.51}$ \\
        \textbf{ProGNN}   & &\checkmark &       &    $\underset{\scriptscriptstyle{\pm 6.20}}{45.10}$ &        - &  $\underset{\scriptscriptstyle{\pm 0.43}}{81.32}$& &        $\underset{\scriptscriptstyle{\pm 1.02}}{46.58}$ &            - &      $\underset{\scriptscriptstyle{\pm 1.12}}{71.82}$& &     $\underset{\scriptscriptstyle{\pm 1.19}}{53.40}$ &         - &   $\underset{\scriptscriptstyle{\pm 0.03}}{49.84}$& &                        $\underset{\scriptscriptstyle{\pm 1.09}}{24.80}$ &                            - &                       $\underset{\scriptscriptstyle{\pm 0.66}}{27.49}$ \\
        \textbf{GCN-SVD} & &\checkmark &           & \cellcolor{gray!20}$\underset{\scriptscriptstyle{\pm 7.59}}{47.82}$ &     $\cellcolor{gray!20}\underset{\scriptscriptstyle{\pm 1.68}}{73.21}$ &  $\underset{\scriptscriptstyle{\pm 0.31}}{76.61}$& &        $\underset{\scriptscriptstyle{\pm 1.78}}{51.20}$ &         $\underset{\scriptscriptstyle{\pm 3.42}}{59.34}$ &      $\underset{\scriptscriptstyle{\pm 0.16}}{66.90}$& &     $\underset{\scriptscriptstyle{\pm 2.06}}{55.00}$ &    \cellcolor{gray!20}$\underset{\scriptscriptstyle{\pm 0.33}}{56.95}$ &   $\underset{\scriptscriptstyle{\pm 0.23}}{55.47}$& & \cellcolor{gray!20}$\underset{\scriptscriptstyle{\pm 0.91}}{25.25}$ &                         $\underset{\scriptscriptstyle{\pm 0.41}}{25.29}$ &                       $\underset{\scriptscriptstyle{\pm 0.25}}{26.63}$ \\
        \textbf{GCN-SMGDC}  & &\checkmark &        &    $\underset{\scriptscriptstyle{\pm 1.18}}{29.66}$ &     $\underset{\scriptscriptstyle{\pm 1.00}}{70.32}$ &  $\underset{\scriptscriptstyle{\pm 0.52}}{77.26}$& &        $\underset{\scriptscriptstyle{\pm 2.36}}{55.04}$ &         $\cellcolor{gray!20}\underset{\scriptscriptstyle{\pm 1.89}}{63.27}$ &      $\underset{\scriptscriptstyle{\pm 0.59}}{72.33}$& &     $\underset{\scriptscriptstyle{\pm 1.19}}{50.76}$ &      $\underset{\scriptscriptstyle{\pm 0.25}}{51.76}$ &   $\underset{\scriptscriptstyle{\pm 0.30}}{51.99}$& &                        $\underset{\scriptscriptstyle{\pm 1.21}}{24.71}$ & \cellcolor{gray!20}$\underset{\scriptscriptstyle{\pm 0.39}}{25.50}$ &                       $\underset{\scriptscriptstyle{\pm 0.60}}{26.06}$ \\
        
        \noalign{\vskip 0.25ex}
        \cdashline{1-3}[0.8pt/2pt]\cdashline{5-19}[0.8pt/2pt]
        \noalign{\vskip 0.25ex}
        
        \textbf{GAT}    &&&         & \cellcolor{gray!20}$\underset{\scriptscriptstyle{\pm 3.60}}{41.70}$ & \cellcolor{gray!20}$\underset{\scriptscriptstyle{\pm 0.31}}{81.96}$ &  $\underset{\scriptscriptstyle{\pm 0.24}}{83.72}$& &        $\underset{\scriptscriptstyle{\pm 2.17}}{48.40}$ &         $\underset{\scriptscriptstyle{\pm 0.69}}{70.70}$ &      $\underset{\scriptscriptstyle{\pm 1.00}}{73.40}$& &     $\underset{\scriptscriptstyle{\pm 0.66}}{50.37}$ & \cellcolor{gray!20}$\underset{\scriptscriptstyle{\pm 0.94}}{61.44}$ &   $\underset{\scriptscriptstyle{\pm 0.92}}{61.69}$& & \cellcolor{gray!20}$\underset{\scriptscriptstyle{\pm 0.73}}{25.00}$ & \cellcolor{blue!20}$\underset{\scriptscriptstyle{\pm 0.13}}{27.45}$ &                       $\underset{\scriptscriptstyle{\pm 0.03}}{27.30}$ \\
        \textbf{GCN}    &&&           &    $\underset{\scriptscriptstyle{\pm 4.83}}{31.98}$ &     $\underset{\scriptscriptstyle{\pm 1.00}}{81.30}$ &  $\underset{\scriptscriptstyle{\pm 0.96}}{83.12}$& & \cellcolor{gray!20}$\underset{\scriptscriptstyle{\pm 2.52}}{49.43}$ &         $\cellcolor{gray!20}\underset{\scriptscriptstyle{\pm 1.73}}{73.30}$ &      $\underset{\scriptscriptstyle{\pm 1.05}}{75.30}$& & \cellcolor{gray!20}$\underset{\scriptscriptstyle{\pm 0.25}}{52.62}$ &      $\underset{\scriptscriptstyle{\pm 0.25}}{54.02}$ &   $\underset{\scriptscriptstyle{\pm 0.13}}{54.20}$& &                        $\underset{\scriptscriptstyle{\pm 0.63}}{24.36}$ &                         $\underset{\scriptscriptstyle{\pm 0.25}}{26.87}$ &                       $\underset{\scriptscriptstyle{\pm 0.13}}{26.68}$ \\
        
        \noalign{\vskip 0.25ex}
        \cdashline{1-3}[0.8pt/2pt]\cdashline{5-19}[0.8pt/2pt]
        \noalign{\vskip 0.25ex}
        
        \textbf{MLP*}    &&&           &    $\underset{\scriptscriptstyle{\pm 1.58}}{64.55}$ &     $\underset{\scriptscriptstyle{\pm 1.58}}{64.55}$ &  $\underset{\scriptscriptstyle{\pm 1.58}}{64.55}$& &        $\underset{\scriptscriptstyle{\pm 0.11}}{67.67}$ &         $\underset{\scriptscriptstyle{\pm 0.11}}{67.67}$ &      $\underset{\scriptscriptstyle{\pm 0.11}}{67.67}$& &     $\underset{\scriptscriptstyle{\pm 0.58}}{56.56}$ &      $\underset{\scriptscriptstyle{\pm 0.58}}{56.56}$ &   $\underset{\scriptscriptstyle{\pm 0.58}}{56.56}$& &                        $\underset{\scriptscriptstyle{\pm 1.05}}{26.25}$ &                         $\underset{\scriptscriptstyle{\pm 1.05}}{26.25}$ &                       $\underset{\scriptscriptstyle{\pm 1.05}}{26.25}$ \\
       
        \bottomrule
    \end{tabular}
    }
\end{table}

\subsection{Detailed Results for Evaluation on Certifiable Robustness}
\label{app:real-cert-results}

\begin{table}[H]

    \centering
    \caption{Accumulated certifications (AC), average certifiable radii ($\bar{r}_a$ and $\bar{r}_d$) and accuracy of GNNs with randomized smoothing enabled (i.e., $f(\phi(\mathbf{s}))$, shown in gray for reference) on all nodes of the clean datasets, with a ramdomization scheme $\phi$ allowing both addition and deletion (i.e., $p_+ = 0.001, p_- = 0.4$), addition only (i.e., $p_+ = 0.001, p_- = 0$) or deletion only (i.e., $p_+ = 0, p_- = 0.4$). For each statistic, we report the mean and stdev across 3 runs. Best results are highlighted in blue per dataset.
    }
    \label{table:real-results-cert-add-or-del}
    \vspace{0.1cm}
    \resizebox{\columnwidth}{!}{
    \begin{tabular}{l lH c rrr>{\color{dark-gray}}r c rrc>{\color{dark-gray}}r c rcr>{\color{dark-gray}}r}
        \toprule
        & \multirow{2}{*}{\rotatebox[origin=r]{90}{\textbf{Hete.}}} & \multirow{3}{*}{\rotatebox[origin=r]{90}{\textbf{Vaccin.}}} &
        
        & \multicolumn{4}{c}{\bf Addition \& Deletion} & 
        & \multicolumn{4}{c}{\bf Addition Only} & 
        & \multicolumn{4}{c}{\bf Deletion Only}
        \\
        \cmidrule{5-8} \cmidrule{10-13} \cmidrule{15-18}
        
        & & & & 
        \multicolumn{1}{c}{AC}  & \multicolumn{1}{c}{$\bar{r}_a$} & \multicolumn{1}{c}{$\bar{r}_d$} & \multicolumn{1}{c}{Acc. \%} & & 
        \multicolumn{1}{c}{AC}  & \multicolumn{1}{c}{$\bar{r}_a$} & \multicolumn{1}{c}{$\bar{r}_d$} & \multicolumn{1}{c}{Acc. \%} & & 
        \multicolumn{1}{c}{AC}  & \multicolumn{1}{c}{$\bar{r}_a$} & \multicolumn{1}{c}{$\bar{r}_d$} & \multicolumn{1}{c}{Acc. \%}
        \\

        \cmidrule{1-3} \cmidrule{5-18}
        
        \textbf{H$_2$GCN} & \checkmark & &
        \multirow{9}{*}{\rotatebox[origin=c]{90}{\textbf{Cora}}}
         & 3.96\tiny{$\pm$0.33} & 0.46\tiny{$\pm$0.08} & 3.90\tiny{$\pm$0.30} & 79.34\tiny{$\pm$1.93} & 
         & 0.42\tiny{$\pm$0.02} & 0.53\tiny{$\pm$0.03} & - & 80.97\tiny{$\pm$1.95} & 
         & 5.44\tiny{$\pm$0.26} & - & 6.46\tiny{$\pm$0.22} & 84.15\tiny{$\pm$1.70} \\
        
        \textbf{GraphSAGE} & \checkmark & &
         & 2.16\tiny{$\pm$0.06} & 0.13\tiny{$\pm$0.00} & 2.43\tiny{$\pm$0.03} & 79.61\tiny{$\pm$1.48} & 
         & 0.28\tiny{$\pm$0.03} & 0.34\tiny{$\pm$0.04} & - & 81.07\tiny{$\pm$1.11} & 
         & 5.07\tiny{$\pm$0.14} & - & 6.12\tiny{$\pm$0.06} & 82.71\tiny{$\pm$1.36} \\
        
        \textbf{CPGNN} & \checkmark & &
         & 1.87\tiny{$\pm$0.27} & 0.14\tiny{$\pm$0.05} & 2.24\tiny{$\pm$0.30} & 75.37\tiny{$\pm$1.65} & 
         & 0.17\tiny{$\pm$0.02} & 0.21\tiny{$\pm$0.03} & - & 78.34\tiny{$\pm$1.26} & 
         & 4.92\tiny{$\pm$0.33} & - & 6.17\tiny{$\pm$0.42} & 79.69\tiny{$\pm$0.81} \\
        
        \textbf{GPR-GNN} & \checkmark & &
         & 4.42\tiny{$\pm$0.43} & 0.63\tiny{$\pm$0.06} & 4.35\tiny{$\pm$0.22} & 74.90\tiny{$\pm$2.34} & 
         & 0.43\tiny{$\pm$0.03} & 0.55\tiny{$\pm$0.03} & - & 76.96\tiny{$\pm$2.18} & 
         & 5.37\tiny{$\pm$0.14} & - & 6.58\tiny{$\pm$0.05} & 81.56\tiny{$\pm$1.59} \\
        
        \textbf{FAGCN} & \checkmark & &
         & 4.30\tiny{$\pm$0.07} & 0.57\tiny{$\pm$0.02} & 4.25\tiny{$\pm$0.04} & 76.49\tiny{$\pm$1.73} & 
         & 0.43\tiny{$\pm$0.01} & 0.54\tiny{$\pm$0.01} & - & 79.04\tiny{$\pm$0.68} & 
         & 5.74\tiny{$\pm$0.06} & - & 7.03\tiny{$\pm$0.05} & 81.56\tiny{$\pm$0.80} \\
        
        \textbf{APPNP} & \checkmark & &
         & 10.11\tiny{$\pm$0.04}\cellcolor{blue!20} & 1.86\tiny{$\pm$0.01}\cellcolor{blue!20} & 8.52\tiny{$\pm$0.06}\cellcolor{blue!20} & 71.97\tiny{$\pm$0.25} & 
         & 0.69\tiny{$\pm$0.00}\cellcolor{blue!20} & 0.95\tiny{$\pm$0.00}\cellcolor{blue!20} & - & 72.27\tiny{$\pm$0.31} & 
         & 6.38\tiny{$\pm$0.02}\cellcolor{blue!20} & - & 8.78\tiny{$\pm$0.02}\cellcolor{blue!20} & 72.75\tiny{$\pm$0.41} \\
        
        \noalign{\vskip 0.25ex}
        \cdashline{1-3}[0.8pt/2pt]
        \cdashline{5-8}[0.8pt/2pt]
        \cdashline{10-13}[0.8pt/2pt]
        \cdashline{15-18}[0.8pt/2pt]
        \noalign{\vskip 0.25ex}
        
        \textbf{GAT} & & &
         & 1.61\tiny{$\pm$0.10} & 0.08\tiny{$\pm$0.01} & 1.85\tiny{$\pm$0.06} & 79.83\tiny{$\pm$2.36} & 
         & 0.19\tiny{$\pm$0.04} & 0.23\tiny{$\pm$0.04} & - & 81.99\tiny{$\pm$1.94} & 
         & 5.58\tiny{$\pm$0.13} & - & 6.60\tiny{$\pm$0.07} & 84.57\tiny{$\pm$1.12} \\
        
        \textbf{GCN} & & &
         & 1.40\tiny{$\pm$0.02} & 0.06\tiny{$\pm$0.01} & 1.75\tiny{$\pm$0.08} & 74.36\tiny{$\pm$3.46} & 
         & 0.13\tiny{$\pm$0.00} & 0.17\tiny{$\pm$0.01} & - & 78.17\tiny{$\pm$2.89} & 
         & 5.39\tiny{$\pm$0.07} & - & 6.50\tiny{$\pm$0.13} & 82.93\tiny{$\pm$0.80} \\

    \midrule

        \textbf{H$_2$GCN} & \checkmark & &
        \multirow{9}{*}{\rotatebox[origin=c]{90}{\textbf{Citeseer}}}
        & 2.96\tiny{$\pm$0.88} & 0.33\tiny{$\pm$0.13} & 3.27\tiny{$\pm$0.67} & 71.76\tiny{$\pm$4.05} & 
        & 0.29\tiny{$\pm$0.05} & 0.40\tiny{$\pm$0.06} & - & 72.99\tiny{$\pm$2.22} & 
        & 5.60\tiny{$\pm$0.09} & - & 7.35\tiny{$\pm$0.15} & 76.28\tiny{$\pm$0.31} \\
        
        \textbf{GraphSAGE} & \checkmark & &
        & 2.21\tiny{$\pm$0.15} & 0.19\tiny{$\pm$0.01} & 2.56\tiny{$\pm$0.09} & 73.48\tiny{$\pm$2.90} & 
        & 0.33\tiny{$\pm$0.01} & 0.44\tiny{$\pm$0.01} & - & 74.70\tiny{$\pm$1.37} & 
        & 5.48\tiny{$\pm$0.09} & - & 7.20\tiny{$\pm$0.13} & 76.05\tiny{$\pm$0.58} \\
        
        \textbf{CPGNN} & \checkmark & &
        & 2.03\tiny{$\pm$0.17} & 0.11\tiny{$\pm$0.01} & 2.52\tiny{$\pm$0.20} & 73.48\tiny{$\pm$0.61} & 
        & 0.15\tiny{$\pm$0.02} & 0.20\tiny{$\pm$0.02} & - & 74.62\tiny{$\pm$0.30} & 
        & 5.59\tiny{$\pm$0.22} & - & 7.54\tiny{$\pm$0.22} & 74.05\tiny{$\pm$0.84} \\
        
        \textbf{GPR-GNN} & \checkmark & &
        & 4.63\tiny{$\pm$0.27} & 0.81\tiny{$\pm$0.07} & 4.92\tiny{$\pm$0.24} & 66.33\tiny{$\pm$0.20} & 
        & 0.40\tiny{$\pm$0.01} & 0.59\tiny{$\pm$0.02} & - & 67.52\tiny{$\pm$0.49} & 
        & 5.05\tiny{$\pm$0.05} & - & 7.21\tiny{$\pm$0.06} & 70.10\tiny{$\pm$0.15} \\
        
        \textbf{FAGCN} & \checkmark & &
        & 4.07\tiny{$\pm$0.15} & 0.58\tiny{$\pm$0.02} & 4.23\tiny{$\pm$0.09} & 71.82\tiny{$\pm$0.73} & 
        & 0.38\tiny{$\pm$0.02} & 0.53\tiny{$\pm$0.02} & - & 72.41\tiny{$\pm$1.03} & 
        & 5.46\tiny{$\pm$0.09} & - & 7.42\tiny{$\pm$0.12} & 73.56\tiny{$\pm$0.18} \\
        
        \textbf{APPNP} & \checkmark & &
        & 9.87\tiny{$\pm$0.02}\cellcolor{blue!20} & 1.88\tiny{$\pm$0.00}\cellcolor{blue!20} & 8.61\tiny{$\pm$0.01}\cellcolor{blue!20} & 69.39\tiny{$\pm$0.23} & 
        & 0.66\tiny{$\pm$0.00}\cellcolor{blue!20} & 0.95\tiny{$\pm$0.00}\cellcolor{blue!20} & - & 69.41\tiny{$\pm$0.22} & 
        & 6.16\tiny{$\pm$0.02}\cellcolor{blue!20} & - & 8.86\tiny{$\pm$0.01}\cellcolor{blue!20} & 69.55\tiny{$\pm$0.22} \\
        
        \noalign{\vskip 0.25ex}
        \cdashline{1-3}[0.8pt/2pt]
        \cdashline{5-8}[0.8pt/2pt]
        \cdashline{10-13}[0.8pt/2pt]
        \cdashline{15-18}[0.8pt/2pt]
        \noalign{\vskip 0.25ex}
        
        \textbf{GAT} & & &
        & 1.29\tiny{$\pm$0.07} & 0.07\tiny{$\pm$0.02} & 1.60\tiny{$\pm$0.06} & 73.62\tiny{$\pm$1.06} & 
        & 0.09\tiny{$\pm$0.01} & 0.12\tiny{$\pm$0.02} & - & 74.47\tiny{$\pm$0.26} & 
        & 5.40\tiny{$\pm$0.17} & - & 7.25\tiny{$\pm$0.12} & 74.49\tiny{$\pm$1.62} \\
        
        \textbf{GCN} & & &
        & 1.79\tiny{$\pm$0.04} & 0.17\tiny{$\pm$0.02} & 2.15\tiny{$\pm$0.11} & 70.38\tiny{$\pm$4.17} & 
        & 0.17\tiny{$\pm$0.01} & 0.24\tiny{$\pm$0.02} & - & 72.04\tiny{$\pm$3.64} & 
        & 5.67\tiny{$\pm$0.09} & - & 7.50\tiny{$\pm$0.12} & 75.63\tiny{$\pm$1.04} \\

    \midrule

        \textbf{H$_2$GCN} & \checkmark & &
        \multirow{9}{*}{\rotatebox[origin=c]{90}{\textbf{FB100}}}
        & 8.12\tiny{$\pm$0.10} & 1.76\tiny{$\pm$0.02} & 8.14\tiny{$\pm$0.06} & 57.38\tiny{$\pm$0.17} & 
        & 0.54\tiny{$\pm$0.00} & 0.94\tiny{$\pm$0.00} & - & 57.11\tiny{$\pm$0.10} & 
        & 4.75\tiny{$\pm$0.03} & - & 8.32\tiny{$\pm$0.03} & 57.15\tiny{$\pm$0.23} \\
        
        \textbf{GraphSAGE} & \checkmark & &
        & 6.98\tiny{$\pm$0.06} & 1.50\tiny{$\pm$0.04} & 7.32\tiny{$\pm$0.13} & 56.72\tiny{$\pm$1.56} & 
        & 0.52\tiny{$\pm$0.01} & 0.92\tiny{$\pm$0.01} & - & 56.70\tiny{$\pm$1.41} & 
        & 4.28\tiny{$\pm$0.05} & - & 7.56\tiny{$\pm$0.10} & 56.58\tiny{$\pm$1.32} \\
        
        \textbf{CPGNN} & \checkmark & &
        & 6.80\tiny{$\pm$0.19} & 1.41\tiny{$\pm$0.21} & 7.05\tiny{$\pm$0.70} & 59.00\tiny{$\pm$5.71} & 
        & 0.54\tiny{$\pm$0.04} & 0.90\tiny{$\pm$0.04} & - & 60.39\tiny{$\pm$7.26} & 
        & 4.22\tiny{$\pm$0.05} & - & 6.66\tiny{$\pm$0.11} & 63.30\tiny{$\pm$0.29} \\
        
        \textbf{GPR-GNN} & \checkmark & &
        & 5.81\tiny{$\pm$0.16} & 1.11\tiny{$\pm$0.02} & 5.95\tiny{$\pm$0.10} & 61.99\tiny{$\pm$0.44} & 
        & 0.46\tiny{$\pm$0.01} & 0.73\tiny{$\pm$0.02} & - & 62.26\tiny{$\pm$0.26} & 
        & 4.04\tiny{$\pm$0.08} & - & 6.51\tiny{$\pm$0.10} & 62.05\tiny{$\pm$0.31} \\
        
        \textbf{FAGCN} & \checkmark & &
        & 7.45\tiny{$\pm$0.21} & 1.53\tiny{$\pm$0.02} & 7.40\tiny{$\pm$0.06} & 59.76\tiny{$\pm$1.47} & 
        & 0.55\tiny{$\pm$0.00} & 0.90\tiny{$\pm$0.01} & - & 60.60\tiny{$\pm$0.36} & 
        & 4.58\tiny{$\pm$0.10} & - & 7.71\tiny{$\pm$0.03} & 59.45\tiny{$\pm$1.26} \\
        
        \textbf{APPNP} & \checkmark & &
        & 8.90\tiny{$\pm$0.03}\cellcolor{blue!20} & 1.92\tiny{$\pm$0.02}\cellcolor{blue!20} & 8.73\tiny{$\pm$0.05}\cellcolor{blue!20} & 57.87\tiny{$\pm$0.57} & 
        & 0.57\tiny{$\pm$0.00}\cellcolor{blue!20} & 0.98\tiny{$\pm$0.01}\cellcolor{blue!20} & - & 57.89\tiny{$\pm$0.59} & 
        & 5.09\tiny{$\pm$0.04}\cellcolor{blue!20} & - & 8.79\tiny{$\pm$0.03}\cellcolor{blue!20} & 57.89\tiny{$\pm$0.59} \\
        
        \noalign{\vskip 0.25ex}
        \cdashline{1-3}[0.8pt/2pt]
        \cdashline{5-8}[0.8pt/2pt]
        \cdashline{10-13}[0.8pt/2pt]
        \cdashline{15-18}[0.8pt/2pt]
        \noalign{\vskip 0.25ex}
        
        \textbf{GAT} & & &
        & 4.30\tiny{$\pm$0.26} & 0.77\tiny{$\pm$0.04} & 4.72\tiny{$\pm$0.19} & 61.56\tiny{$\pm$0.78} & 
        & 0.46\tiny{$\pm$0.03} & 0.74\tiny{$\pm$0.04} & - & 61.97\tiny{$\pm$1.41} & 
        & 3.33\tiny{$\pm$0.09} & - & 5.37\tiny{$\pm$0.11} & 62.01\tiny{$\pm$1.01} \\
        
        \textbf{GCN} & & &
        & 5.19\tiny{$\pm$0.03} & 1.14\tiny{$\pm$0.00} & 6.05\tiny{$\pm$0.01} & 54.16\tiny{$\pm$0.08} & 
        & 0.43\tiny{$\pm$0.00} & 0.79\tiny{$\pm$0.01} & - & 54.39\tiny{$\pm$0.14} & 
        & 3.63\tiny{$\pm$0.01} & - & 6.73\tiny{$\pm$0.02} & 53.96\tiny{$\pm$0.08} \\

    \midrule

        \textbf{H$_2$GCN} & \checkmark & &
        \multirow{9}{*}{\rotatebox[origin=c]{90}{\textbf{Snap}}}
        & 1.44\tiny{$\pm$0.18} & 0.59\tiny{$\pm$0.10} & 3.79\tiny{$\pm$0.40} & 26.97\tiny{$\pm$0.10} & 
        & 0.11\tiny{$\pm$0.01} & 0.42\tiny{$\pm$0.05} & - & 26.74\tiny{$\pm$0.18} & 
        & 1.41\tiny{$\pm$0.04} & - & 5.16\tiny{$\pm$0.19} & 27.28\tiny{$\pm$0.21} \\
        
        \textbf{GraphSAGE} & \checkmark & &
        & 0.70\tiny{$\pm$0.21} & 0.19\tiny{$\pm$0.11} & 2.16\tiny{$\pm$0.54} & 26.84\tiny{$\pm$0.47} & 
        & 0.06\tiny{$\pm$0.02} & 0.24\tiny{$\pm$0.08} & - & 27.00\tiny{$\pm$0.63} & 
        & 1.10\tiny{$\pm$0.11} & - & 4.04\tiny{$\pm$0.36} & 27.21\tiny{$\pm$0.99} \\
        
        \textbf{CPGNN} & \checkmark & &
        & 1.45\tiny{$\pm$0.23} & 0.61\tiny{$\pm$0.14} & 3.89\tiny{$\pm$0.51} & 26.71\tiny{$\pm$0.25} & 
        & 0.12\tiny{$\pm$0.02} & 0.43\tiny{$\pm$0.08} & - & 27.00\tiny{$\pm$0.41} & 
        & 1.69\tiny{$\pm$0.06} & - & 6.39\tiny{$\pm$0.13} & 26.45\tiny{$\pm$0.50} \\
        
        \textbf{GPR-GNN} & \checkmark & &
        & 0.52\tiny{$\pm$0.06} & 0.11\tiny{$\pm$0.01} & 1.70\tiny{$\pm$0.14} & 26.31\tiny{$\pm$1.03} & 
        & 0.03\tiny{$\pm$0.01} & 0.11\tiny{$\pm$0.02} & - & 26.14\tiny{$\pm$0.73} & 
        & 1.10\tiny{$\pm$0.04} & - & 4.19\tiny{$\pm$0.17} & 26.24\tiny{$\pm$0.43} \\
        
        \textbf{FAGCN} & \checkmark & &
        & 1.41\tiny{$\pm$0.10} & 0.56\tiny{$\pm$0.06} & 3.81\tiny{$\pm$0.22} & 27.07\tiny{$\pm$0.16} & 
        & 0.10\tiny{$\pm$0.01} & 0.36\tiny{$\pm$0.03} & - & 27.13\tiny{$\pm$0.16} & 
        & 1.77\tiny{$\pm$0.05} & - & 6.48\tiny{$\pm$0.20} & 27.25\tiny{$\pm$0.18} \\
        
        \textbf{APPNP} & \checkmark & &
        & 3.54\tiny{$\pm$0.03}\cellcolor{blue!20} & 1.68\tiny{$\pm$0.01}\cellcolor{blue!20} & 7.95\tiny{$\pm$0.04}\cellcolor{blue!20} & 27.45\tiny{$\pm$0.14} & 
        & 0.24\tiny{$\pm$0.00}\cellcolor{blue!20} & 0.86\tiny{$\pm$0.00}\cellcolor{blue!20} & - & 27.46\tiny{$\pm$0.17} & 
        & 2.38\tiny{$\pm$0.01}\cellcolor{blue!20} & - & 8.69\tiny{$\pm$0.05}\cellcolor{blue!20} & 27.41\tiny{$\pm$0.09} \\
        
        \noalign{\vskip 0.25ex}
        \cdashline{1-3}[0.8pt/2pt]
        \cdashline{5-8}[0.8pt/2pt]
        \cdashline{10-13}[0.8pt/2pt]
        \cdashline{15-18}[0.8pt/2pt]
        \noalign{\vskip 0.25ex}
        
        \textbf{GAT} & & &
        & 0.28\tiny{$\pm$0.09} & 0.04\tiny{$\pm$0.01} & 0.95\tiny{$\pm$0.33} & 27.12\tiny{$\pm$0.52} & 
        & 0.02\tiny{$\pm$0.00} & 0.08\tiny{$\pm$0.02} & - & 27.00\tiny{$\pm$0.59} & 
        & 1.10\tiny{$\pm$0.03} & - & 4.06\tiny{$\pm$0.11} & 27.18\tiny{$\pm$0.04} \\
        
        \textbf{GCN} & & &
        & 0.32\tiny{$\pm$0.08} & 0.06\tiny{$\pm$0.03} & 1.08\tiny{$\pm$0.24} & 26.17\tiny{$\pm$0.34} & 
        & 0.02\tiny{$\pm$0.01} & 0.08\tiny{$\pm$0.03} & - & 26.38\tiny{$\pm$0.49} & 
        & 1.29\tiny{$\pm$0.06} & - & 4.83\tiny{$\pm$0.27} & 26.69\tiny{$\pm$0.36} \\

    \bottomrule
    \end{tabular}}
\end{table}

\subsection{Comparison Between Certifiable and Empirical Robustness} 
\label{app:real-empirical-cert-diff}
While the evaluations on both empirical and certifiable robustness have shown that methods featuring the design have shown largely improved robustness compared to the best-performing unvaccinated method, we find that the robustness rankings for methods under certifiable and empirical robustness are different; previous results from \cite{geisler2020reliable} have also shown discrepancy in certifiable and empirical robustness rankings.

We think this discrepancy may be attributed to multiple factors.  
Firstly, the radius on which certificates can be issued with randomized smoothing may not cover the radius of perturbations we allowed for \nettack{} and Metattack in \S\ref{sec:exp-benchmark-study}, which are more than tens of edges for Metattack and high-degree nodes in \nettack{} (where we use an attack budget equal to the degree of the target node). Furthermore, even for low-degree nodes, attacks are much more inclined to introduce new edges instead of removing existing ones, as we have shown in \S\ref{sec:exp-perturb-observations}, which can make it much harder to obtain certificates \citep{bojchevski2020efficient}. In our case, the methods we evaluated generally have $\bar{r}_a < 1$, meaning that for most nodes there are no certificates to cover even the addition of a single edge. Thus, methods which display higher certifiable robustness under smaller perturbations may not keep their robustness under larger perturbations by empirical attacks. 
Moreover, while we are evaluating on randomized smoothed models $f(\phi(\mathbf{s}))$ to measure certifiable robustness, in empirical robustness we are evaluating on the robustness of the base models $f(s)$, which may differ from the robustness of the randomized smoothed models.
Lastly, it is worth noting that the lack of certification for a model within certain radius does not imply a vulnerability to all adversarial attacks within that radius; the model may still be robust against many attacks within that radius. Similarly, it is also possible for existing certification approaches to miss some certifiable cases. 
Taking all these factors into account, we believe that while evaluations on certifiable robustness provide complementary perspectives to evaluations empirical robustness, at the current stage it cannot replace the evaluations on empirical robustness, and the relation between certifiable and empirical robustness remains as a question for future works. 

\subsection{Complexity and Runtime of Heterophilous Vaccination}
\label{app:real-runtime-complexity}

Another benefit of adopting heterophily-inspired design for boosting robustness of GNNs is their smaller computational overhead compared to existing vaccination mechanisms, especially vaccinations based on low-rank approximation. 
As our identified design can be applied as simple architectural changes on top of an existing GNN, they usually maintain the same order of computational complexity as the base model. For example, adding the heterophilous design to GCN~\citep{kipf2016semi} results in an architecture similar to GraphSAGE~\citep{hamilton2017inductive}; both have the same order of computational complexity as $O(|\vertexSet| + |\edgeSet|)$ 
by leveraging the sparse connectivity of most real-world graphs.
Low-rank approximation-based vaccination, on the other hand, approximates the adjacency matrix of a graph by an SVD, resulting in an adjusted low-rank adjacency matrix $\tilde{\matA}$ based on which the GNN runs.
However, not only is computing an SVD potentially costly ($O(|\vertexSet|^3)$ in general), but in most cases it also results in a dense $\tilde{\matA}$ (in contrast to the sparse original adjacency matrix), thus increasing the complexity of each iteration of the GNN. 

\begin{table}[h]
	\centering
	\caption{Runtime (in seconds) of 200 training iterations on Cora. See App. \ref{app:exp-details} for the implementation used for each method. %
	} %
	\label{table:runtime}
	\resizebox{1\columnwidth}{!}{
	\begin{tabular}{cccccccc}
		\toprule 
		 \multicolumn{1}{c}{\textbf{GCN}} 
		 & \multicolumn{1}{c}{\textbf{GAT}} 
		 & \multicolumn{1}{c}{\textbf{GNNGuard}} 
		 & \multicolumn{1}{c}{\textbf{ProGNN}} 
		 & \multicolumn{1}{c}{\textbf{GCN-SVD}} 
		 & \multicolumn{1}{c}{\textbf{\method}} 
         & \multicolumn{1}{c}{\textbf{GraphSAGE}} 
         & \multicolumn{1}{c}{\textbf{FAGCN}}
		 \\ \midrule 
	
	11.11 &2.98 &39.63 &220.30 &134.81 &16.54 & 17.24 & 1.91 \\

    \midrule
    \multicolumn{1}{c}{\textbf{GPR-GNN}} 
    & \multicolumn{1}{c}{\textbf{CPGNN}}
    & \multicolumn{1}{c}{\textbf{\textsc{H$_{2}$GCN}-SVD}} 
    & \multicolumn{1}{c}{\textbf{GraphSAGE-SVD}} 
    & \multicolumn{1}{c}{\textbf{GCN-SMGDC}} 
    & \multicolumn{1}{c}{\textbf{H2GCN-SMGDC}} 
    & \multicolumn{1}{c}{\textbf{GraphSAGE-SMGDC}} 
    & \multicolumn{1}{c}{\textbf{APPNP}} \\
    \midrule
    2.66 & 24.08 &62.33 & 55.45 & 23.56 & 33.56 & 22.78 & 5.23 \\
    \bottomrule
	\end{tabular}
	}
\end{table}

Table~\ref{table:runtime} shows the runtime of 200 training iterations of each model.
We observe that models with the heterophilous design have the smallest runtime among all vaccinated models. 
Even for methods based on the same implementation, \method and GraphSAGE are still 3-4 times faster than the corresponding H$_2$GCN-SVD and GraphSAGE-SVD methods. For fair runtime measurements, we measure the runtime of each model on an Amazon EC2 instance with instance type as \texttt{p3.2xlarge}, which features an 8-core CPU, 61 GB Memory, and a Tesla V100 GPU with 16 GB GPU Memory.

\end{document}